\documentclass[journal, onecolumn, pagenumbers]{IEEEtran}
\usepackage{listings}
\usepackage{amsmath}
\usepackage{amsthm}
\usepackage{tikz}
\usepackage{caption}
\usepackage{array}
\usepackage{mdwmath}
\usepackage{multirow}
\usepackage{mdwtab}
\usepackage{eqparbox}
\usepackage{amsfonts}
\usepackage{tikz}
\usepackage{multirow,bigstrut,threeparttable}
\usepackage{amsthm}
\usepackage{array}
\usepackage{bbm}
\usepackage{subfigure}
\usepackage{epstopdf}
\usepackage{mdwmath}
\usepackage{mdwtab}
\usepackage{eqparbox}
\usepackage{tikz}
\usepackage{latexsym}
\usepackage{cite}
\usepackage{amssymb}
\usepackage{bm}
\usepackage{amssymb}
\usepackage{graphicx}
\usepackage{mathrsfs}
\usepackage{epsfig}
\usepackage{psfrag}
\usepackage{setspace}
\usepackage{hyperref}
\usepackage{algorithm}
\usepackage{algpseudocode}
\usepackage{stfloats}
\usepackage{mathtools}
\usepackage{algorithm}
\usepackage{algorithmicx}
\usepackage{algpseudocode}
\usepackage{float}

\usepackage{ytableau}
\usepackage{hypcap}

\newtheorem{remark}{Remark}

\newtheorem{theorem}{Theorem}

\newtheorem{definition}{Definition}

\newcommand{\beq}{\begin{equation}}
\newcommand{\eeq}{\end{equation}}
\newcommand{\mc}{\mathcal}
\newcommand{\mb}{\mathbb}
\newcommand{\mbf}{\mathbf}
\newcommand{\eqdef}{\triangleq}
\newcommand{\ra}{\rightarrow}
\DeclareMathOperator{\perm}{perm}
\DeclareMathOperator{\supp}{Supp}
\DeclareMathOperator*{\argmax}{argmax}
\DeclareMathOperator*{\argmin}{argmin}

\begin{document}

\title{Approximate Profile Maximum Likelihood}

\author{Dmitri S. Pavlichin,~Jiantao~Jiao,~\IEEEmembership{Student Member,~IEEE}, and Tsachy~Weissman,~\IEEEmembership{Fellow,~IEEE}
\thanks{Dmitri S. Pavlichin is with the Department of Applied Physics, and Jiantao Jiao and Tsachy Weissman are with the Department of Electrical Engineering, Stanford University, CA, USA. Email: \{dmitrip, jiantao, tsachy\}@stanford.edu}. 
}
%
%

\date{\today}

\vspace{-10pt}

\maketitle

\begin{abstract}
We propose an efficient algorithm for approximate computation of the profile maximum likelihood (PML), a variant of maximum likelihood maximizing the probability of observing a sufficient statistic rather than the empirical sample.  The PML has appealing theoretical properties, but is difficult to compute exactly.  Inspired by observations gleaned from exactly solvable cases, we look for an approximate PML solution, which, intuitively, clumps 
comparably frequent
symbols into one symbol.  This amounts to lower-bounding a certain matrix permanent by summing over a subgroup of the symmetric group rather than the whole group during the computation.  We extensively experiment with the approximate solution, and find the empirical performance of our approach is competitive and sometimes significantly better than state-of-the-art performance for various estimation problems. 
\end{abstract}


\section{Introduction}

The maximum likelihood principle, proposed by Ronald Fisher, has proved to be a powerful and versatile method used throughout nearly all scientific fields. However, it is still the source of considerable controversy in the statistics and machine learning community. Indeed, quoting Efron~\cite{Efron1982maximum}:
\vspace{.5ex}

\begin{center}
\parbox{.85\textwidth}{~~\emph{ \small ``The controversy centers on the relationship between decision theory and maximum likelihood. Beginning with the Neyman--Pearson lemma, decision theory has reshaped the theory and practice of hypothesis testing..."}}
\end{center}
\vspace{.5ex}

Indeed, Wald's statistical decision theory~\cite{Wald1950statistical} provided a comprehensive framework for evaluating and proposing statistical procedures. The maximum likelihood approach, which can be proved to be asymptotically efficient under mild conditions in the H\'ajek--Le Cam theory~\cite[Chap. 9]{Vandervaart2000}, lacks a finite-sample justification. Indeed, as Le Cam~\cite{LeCam1990maximum} argued, even in the asymptotic regime the maximum likelihood approach may have weird behavior and is not always consistent. 

The recent work~\cite{acharya2016unified} provided an elegant justification of the maximum likelihood approach which is intimately connected to decision theory. It was shown in~\cite{acharya2016unified} that plugging-in the \emph{profile maximum likelihood} (PML) distribution estimators into a variety of functionals achieves the optimal sample complexity in estimating those functionals, which the generic \emph{sequence maximum likelihood} (SML) fails to achieve. 

To compute the profile maximum likelihood estimator we must solve the following optimization problem: given $n$ samples with empirical distribution $\hat{p} = (\hat{p}_1, \hat{p}_2, \ldots)$, maximize the probability to observe $\hat{p}$ up to \textit{relabeling} $\sigma \hat{p} \eqdef (\hat{p}_{\sigma(1)}, \hat{p}_{\sigma(2)}, \ldots)$ of the empirical distribution, where $\sigma$ is some permutation.  This amounts to computing the PML distribution $p^*$:
\begin{equation}
    p^* = \argmax_p \sum_{\sigma} e^{-n D(\sigma \hat{p} || p)} \label{eq:PML_def_in_introduction}
\end{equation}
where the max is over all discrete distributions $p$ with a known support set size 
(we treat the unknown case later; it is similar),
the sum is over all permutations $\sigma$ of the support set of distribution $p$, and where $D(\cdot||\cdot)$ is the Kullback-Leibler divergence.

The profile maximum likelihood estimator is computationally challenging to find -- as can shown to be equivalent to optimizing a certain matrix permanent -- and the best known algorithm has running time exponential in the support size of the unknown discrete distribution. Our main contributions are the following: 
\begin{enumerate}
    \item We present an efficiently computable approximation for the PML distribution.
        The approximation idea is motivated by observations for the exact PML for small alphabets, where the PML distribution tends to put symbols
        whose empirical counts are close (within $O(\sqrt{n})$) into the same level set\footnote{A level set of distribution $p$ is $\{x: p_{x} = u\}$ for some value $u$.}.  The idea is to lower bound the sum in (\ref{eq:PML_def_in_introduction}) by summing over only the permutations that contribute ``a lot'' -- namely the subgroup of permutations that only mix symbols within the level sets of $p$.  This leads to the objective function in the argmax below (a lower bound to the objective function in (\ref{eq:PML_def_in_introduction})) to define the approximate PML (APML) distribution $\bar{p}^*$: 
        \begin{align}
            \bar{p}^* &= \argmax_p \bigg(e^{- n D(\hat{p}||p)} \prod_{\alpha \in \mc{A}(p)} |\alpha|!\bigg) 
        \end{align}
        where $\mc{A}(p)$ is the partition of the support set of $p$ into the level sets $\alpha$ of $p$.  The second factor of the objective function rewards clumping many symbols together into a few large level sets, while the first factor encourages similarity to the empirical distribution and dominates as the sample size $n \ra \infty$.  

        We present a dynamic programming approach to compute the approximate PML distribution $\bar{p}^*$.
        Compared with existing approximations to the PML~\cite{orlitsky2004modeling,Vontobel12}, our algorithm has no tuning parameters, is deterministic, runs in at most linear time in the sample size (usually computing the empirical histogram is the slowest part), and achieves state-of-the-art empirical performance in various estimation tasks that we detail below. 

    For the case of unknown support set size, we state the appropriate generalization of the PML distribution and its approximation.
        We modify our dynamic programming algorithm slightly to handle this case, preserving a linear worst-case runtime.
        It may happen that the PML and our approximate PML distributions have a discrete part with finite support and a ``continuous part'' in the terminology of \cite{orlitsky2004modeling}; we are able to detect this case and find the discrete part of the approximate PML distribution and the probability mass of the continuous part.

\item Given the result in~\cite{acharya2016unified} that plugging in the PML distribution into various functionals achieves the optimal sample complexity, we extensively experiment with the plug-in estimator using our APML distribution.  We estimate entropy, R\'enyi entropy, support set size, $L_1$ distance to uniformity, and the sorted probability vector of a discrete distribution and compare with the state-of-the-art approaches with available code for estimating those functionals.  We find that the performance of plugging in our APML distribution into those functionals is consistently competitive, and sometimes much better than state-of-the-art packages.

\item We extend our approximation scheme to the multi-dimensional PML problem, which is intimately connected to estimating divergence functions of discrete distributions. Utilizing results on partitions of bipartite numbers, we show that solving the two-dimensional PML problem leads to estimators that achieve the optimal sample complexity for estimating a variety of functionals such as the KL divergence, the $L_1$ distance, the squared Hellinger distance, and the $\chi^2$ divergence.  The multi-dimensional APML distribution turns out to be harder to compute than the one-dimensional APML, 
        so we settle for a greedy heuristic to approximate the multi-dimensional APML distribution.
    
    
    

\item We provide extensive experimental results on the plug-in approach for estimating divergence functions using our approximation to the two-dimensional PML. This achieves competitive and sometimes much better results than state-of-the-art approaches in KL divergence and $L_1$ distance estimation. 


\item We generalize the PML idea to general group actions, leading to the generalized PML approach for mutual information estimation.  We analyze several candidates for the PML solution and show that one candidate makes the key arguments in~\cite{acharya2016unified} fail, while the other succeeds. 
\end{enumerate}

This work is organized as follows.  Section \ref{sec:PML_and_sufficiency} reviews some results on profile maximum likelihood and functional estimation.  Section \ref{sec.discredistributions} specializes the discussion of Section \ref{sec:PML_and_sufficiency} to the setting of estimating functionals of discrete distributions, and defines the PML and multi-dimensional PML distributions.  Section \ref{sec.approximatepml} presents our APML approach and observations about the exact PML that inspired it.  Section \ref{sec.standardpmlplugin} shows the results of numerical experiments using the APML to estimate symmetric functionals of a distribution: the sorted probability vector, entropy, R\'enyi entropy, $L_1$ distance to uniformity, and support set size.  Section \ref{sec.divergencepmlplugin} shows the results of numerical experiments using the APML to estimate symmetric functionals of multiple distributions: the KL divergence and $L_1$ distance.  Appendices contain proofs, examples, most of the notation and algorithms related to the multi-dimensional APML (Appendix \ref{app:dd_PML}), and application of the PML to estimation of mutual information (Appendix \ref{sec.mutuallautum}).

We release code for computing the approximate PML distributions at \cite{PMLsoftware}.

\section{The principle of profile maximum likelihood and sufficiency} \label{sec:PML_and_sufficiency}

We rephrase the key result in~\cite{acharya2016unified} regarding the maximum likelihood principle in Theorem \ref{theorem.ml}; similar ideas appeared in~\cite{acharya2011competitive}. We define a statistical model 
\begin{align}
\mathcal{E} & = \{\mathcal{X}, P_\theta, \theta \in \Theta\},
\end{align}
where the observation $X\sim P_\theta$ for some $\theta \in \Theta$, $X\in \mathcal{X}$. Let $F(\theta): \Theta \mapsto \mathcal{Y}$ be a measurable function. Let $d(F,\hat{F}): \mathcal{Y} \times \mathcal{Y} \mapsto \mathbb{R}_{\geq 0}$ be the loss function, which is also assumed to be a metric.

Given observation $x$, upon asserting that a statistic $T$ is \emph{sufficient} for estimating $F(\theta)$, the general profile maximum likelihood approach aims at maximizing the probability that $t = T(x)$ appears rather than the probability of $x$, which is the aim of traditional maximum likelihood. Then, we simply plug-in the PML estimator into the function $F(\cdot)$ to obtain an estimate of the functional $F(\theta)$. The profile maximum likelihood estimator of $\theta$ is defined as follows. 
\begin{definition}[Profile maximum likelihood]\label{def.pmlalgorithm}
The profile maximum likelihood estimator of $\theta$ is defined as
\begin{align}
\hat{\theta}_{\mathrm{T}}(t) \triangleq \argmax_\theta P_\theta( T(X) = t ). 
\end{align}
\end{definition}

\begin{remark}
The notion of \emph{profile likelihood} has appeared in the statistics literature before~\cite{murphy2000profile}, bearing a different meaning from the one we adopted. 
\end{remark}
 
The following theorem provides a general performance guarantee for the PML algorithm introduced in Definition~\ref{def.pmlalgorithm} in terms of estimating any functional $F(\theta)$. 
\begin{theorem}\label{theorem.ml} \cite{acharya2016unified}
We fix a statistical model $\mathcal{E} = \{\mathcal{X}, P_\theta: \theta \in \Theta\}$. Let $T = T(X)$ be a statistic such that $T: \mathcal{X} \mapsto \mathcal{T}$ and $|\mathcal{T}|<\infty$. Let $F(\theta): \Theta \mapsto \mathcal{Y}$ be a measurable function. Suppose there exists an estimator $\hat{F}: \mathcal{T} \mapsto \mathcal{Y}$ such that
\begin{align}
\sup_{\theta\in \Theta} P_\theta\left( d(F(\theta), \hat{F}(T)) >\epsilon \right) <\delta.
\end{align}
Then, 
\begin{align}
\sup_{\theta \in \Theta} P_\theta \left( d( F(\theta) , F(\hat{\theta}_{\mathrm{T}})) > 2\epsilon \right) \leq \delta \cdot |\mathcal{T}|. 
\end{align}
\end{theorem}

The usefulness of Theorem~\ref{theorem.ml} would rely on two factors: the existence of a good estimator $\hat{F}(T)$ in terms of \emph{worst case risk}, and the small cardinality of the set $\mathcal{T}$ that the statistic $T$ lies in. As we argued above, it would be sensible to employ the generalized PML approach if we can assert that the statistic $T$ is \emph{sufficient} for estimating $F(\theta)$. However, what is the definition of sufficiency for a subparameter (or sufficiency in the presence of a nuisance parameter)?

The definition of sufficiency in this context is one of the most basic problems in decision theory, which turns out to be a highly non-trivial question. It was considered by Kolmogorov~\cite{kolmogoroff1942estimation}, who proposed the following definition:
\begin{definition}[Kolmogorov's definition of sufficiency]\cite{kolmogoroff1942estimation}
A statistic $T = T(X)$ is called sufficient for $F(\theta)$, if the posterior distribution of $F(\theta)$ given $X = x$, depends only on $T = t$ and on the prior distribution of $\theta$. 
\end{definition}

It was later shown by H\'ajek~\cite{hajek1967basic} that the Kolmogorov definition is void in the sense that if $F(\theta)$ is not a constant, and $T$ is sufficient for $F(\theta)$ in the sense of Kolmogorov, then $T$ is sufficient for $\theta$ as well. 

In this context, H\'ajek proposed another definition of sufficiency. \footnote{For other definitions of sufficiency, see~\cite{le1964sufficiency, birnbaum1962foundations, blackwell1979theory}.  }
\begin{definition}\cite[
Def. 2.2]{hajek1967basic}\label{def.hajeksufficiency}
A statistic $T = T(X)$ is called sufficient for $F(\theta)$ if there exists some other functional $R(\theta)$ such that $F(\theta) = F_1(R(\theta))$, and the following is satisfied: 
\begin{enumerate}
\item The distribution of $T$ will depend on $R(\theta)$ only, that is
\begin{align}
P_\theta(dT) = P_{R(\theta)}(dT);
\end{align}
\item There exists a distribution $Q_R \in \mathcal{P}_R$ such that $T$ is sufficient for the family $\{Q_R\}$, where $\mathcal{P}_R$ is the convex hull of the distributions $\{P_\theta \colon R(\theta) = R\}$. 
\end{enumerate}
\end{definition}

It was shown in~\cite{hajek1967basic} that under the sufficiency in Def.~\ref{def.hajeksufficiency}, an analogue of the Rao--Blackwell theorem can be proved. 


It begs the question: in general how could one prove a certain statistic $T$ is sufficient for $F(\theta)$? The following equivalence relation induced by group transformations is common in practice. Let $G = \{g\}$ be a group of one-to-one transformations of the $\mathcal{X}$ space on itself. We shall say that an event is $G$-invariant, if $gA = A$ for all $g\in G$. The set of $G$-invariant events is a sub-$\sigma$-algebra $\mathcal{B}$, and a measurable function $f$ is $\mathcal{B}$-measurable if and only if $f(gx) = f(x)$ for all $g\in G$. 

\begin{theorem}\cite{hajek1967basic}\label{theorem.groupaction}
Consider a family of probability distributions $\{P_\tau\}$ and define $P_{\tau,g}$ for each $\tau$ as
\begin{align}
P_{\tau,g}(X\in A)  = P_{\tau}(gX \in A). 
\end{align}
Letting $\theta = (\tau,g)$, the statistic corresponding to the sub-$\sigma$-algebra of $G$-invariant events is sufficient for $\tau$ in the sense of Def.~\ref{def.hajeksufficiency}. In other words, it is sufficient for $\tau$ to look at the set of equivalence classes $\mc{X}/G$, while $Gx \in \mc{X}/G$ denotes the equivalence class (orbit) of $x$:
\begin{align}
Gx \eqdef \{gx : g \in G\}. 
\end{align}
\end{theorem}





\section{Discrete distributions up to relabeling} 
\label{sec.discredistributions}

In this section, we specialize Theorem~\ref{theorem.groupaction} to several different settings, centered around the problem of estimating functionals of discrete distributions.

\subsection{Permutation group and single sorted probability vector}\label{sec.standardpml}

The standard PML problem can be viewed as a special case of 
Theorem~\ref{theorem.groupaction} with the group being the permutation group. 
Concretely, let $G$ be the permutation group $\mc{S}_{\mathcal{X}}$ on 
$\mathcal{X}$, let $p$ be a distribution supported on set $\mc{X}$ -- that is, 
$p_x > 0 \ \forall x \in \mc{X}$ --
and let $\tau$ be the sorted non-increasing probability vector 
$(p_{(1)},p_{(2)},\ldots,p_{(|\mathcal{X}|)})$ of $p$.  It is clear that the 
label-invariant properties of a distribution, such as the entropy, R\'enyi 
entropy, and support set size, depend on $p$ only through $\tau$.


Suppose we observe $n$ i.i.d. samples $x_1^n = (x_1,\ldots,x_n)$ with 
distribution $p$. Denote by $\hat{p} = \hat{p}(x_1^n)$ the empirical 
distribution:
\begin{align}
    \hat{p} = ( \hat{p}_x)_{x\in \mathcal{X}} = \left(\frac{1}{n} \sum_{i = 1}^n \mathbbm{1}(x_i = x)\right)_{x \in \mc{X}}
\end{align}

It is well known that the empirical distribution is the minimum complete 
sufficient statistic for $p$~\cite{Lehmann--Casella1998theory}.
The probability of observing a specific empirical distribution is given 
by
\begin{align}
    \mb{P}_p(\hat{p}) = \left(\begin{array}{c} n \\ n \hat{p} 
        \end{array}\right) \prod_{x \in \mc{X}} p_x^{n \hat{p}_x} 
            \label{eq:prob_empirical_distribution}
\end{align}
where the prefactor in (\ref{eq:prob_empirical_distribution}) is a multinomial 
coefficient\footnote{$\left(\begin{array}{c} n \\ n \hat{p} \end{array}\right) \eqdef n! \prod_{x 
    \in \mc{X}} \frac{1}{(n \hat{p}_x)!}$.}.



Applying Theorem~\ref{theorem.groupaction} with $\theta = p = (\tau,g)$, we obtain 
that the \emph{fingerprint} 
statistic~\cite{ValiantValiant11,Valiant--Valiant2013estimating}\footnote{The 
fingerprint is also called the profile~\cite{orlitsky2004modeling}, histogram 
order statistics~\cite{Paninski2003}, and histogram of 
histograms~\cite{batu2000testing}. } is sufficient for $\tau$. Concretely, the 
fingerprint $\mc{F} = \left(\mc{F}_i\right)_{i \geq 0}$ is defined so that 
$\mc{F}_i$ is the number of symbols observed exactly $i$ times in $x_1^n$:
\beq
    \mc{F} = \mc{F}(\hat{p}(x_1^n)) = \left(\mc{F}_i\right)_{i \geq 0} \eqdef 
    \left(|\{x \in \mc{X} : n \hat{p}_x = i\}|\right)_{i \geq 0} 
    \label{eq:def_fingerprint}
\eeq
Below, we compute the probability of a specific fingerprint for the case that 
$p$ is supported on finite alphabet $\mc{X} = \supp(p)$ with
\begin{equation}
    K \eqdef |\mc{X}|
\end{equation}
and the empirical distribution $\hat{p}$ is supported on empirical 
alphabet $\hat{\mc{X}} \eqdef \supp(\hat{p}) = \{x \in \mc{X} : \hat{p}_x > 0\} 
\subset \mc{X}$ with 
\begin{equation}
    \hat{K} \eqdef |\hat{\mc{X}}|.
\end{equation}
Then $\mc{F}_0 = |\mc{X} \setminus \hat{\mc{X}}| = K - \hat{K}$ counts the 
number of ``unseen'' symbols, and is thus unknown if the support set size $K$ 
is unknown.  The probability of a specific fingerprint is given by:
\begin{align}
    \mb{P}_p(\mc{F}) &= \left(\prod_{i \geq 0} \frac{1}{\mc{F}_i!}\right) 
    \left(\begin{array}{c} n \\ n \hat{p} \end{array}\right) 
        \perm\bigg(\underbrace{\left(\begin{array}{c} p_x^{n \hat{p}_{x'}} 
        \end{array}\right)_{x,x'\in \mc{X}}}_{Q}\bigg) 
            \label{eq:prob_unlabeled_empirical_distribution}
\end{align}
where $\perm(A)$ denotes the matrix permanent of the $K \times K$ matrix $A$:
\beq
    \perm(A) = \sum_{\sigma \in \mc{S}_\mc{X}} \prod_{x \in \mc{X}} 
    A_{x,\sigma(x)} \label{eq:def_permanent}
\eeq
where the sum is over the symmetric group $\mc{S}_\mc{X}$ on $\mc{X}$.  
To simplify notation, let $Q$ denote the $K \times K$ matrix in 
(\ref{eq:prob_unlabeled_empirical_distribution}).
Note that to evaluate expression 
(\ref{eq:prob_unlabeled_empirical_distribution}) we need to know the support 
set $\mc{X}$, both to evaluate $\perm(Q)$ and $\mc{F}_0 = |\mc{X} 
\setminus \hat{\mc{X}}|$. 
Appendix \ref{app:prob_unlabeled_empirical_distribution} proves expression 
(\ref{eq:prob_unlabeled_empirical_distribution}) for $\mb{P}_p(\mc{F})$.
Appendix \ref{app.illustration} shows examples of the computation of 
$\mb{P}_p(\mc{F})$ over small alphabets.


For a given collection of distributions $\mathcal{P}$, the \emph{profile 
maximum likelihood} distribution is defined as \begin{align}
    p^* &\eqdef \argmax_{p \in \mathcal{P}} \mb{P}_p(\mc{F}) = \argmax_{p \in 
    \mathcal{P}} \frac{\perm\left(Q\right)}{(K - \hat{K})!},
    \label{eq:def_unlabeled_ML_alphabet_size_known}
\end{align}
where in the second equality we discarded all $p$-independent factors of 
$\mb{P}_p(\mc{F})$ (\ref{eq:prob_unlabeled_empirical_distribution}).  $\mc{F}_0 
= K - \hat{K} =
|\mc{X}\setminus\hat{\mc{X}}|$ depends on $p$ through its support set size.  
Note that $\mb{P}_p(\mc{F})$ is invariant under relabeling of the components of 
$p$, so we can choose $p^*$ to be non-increasing in the same ordering as we 
choose for the support set $\mc{X}$.
Note that the set $\mathcal{P}$ is not necessarily the same as the set of all 
discrete distributions.

If the collection of distributions $\mc{P}$ includes distributions with 
different support set sizes (for example, all finite support set sizes at least 
as large as $\hat{K}$), then we can estimate the support set size by breaking 
up the optimization in (\ref{eq:def_unlabeled_ML_alphabet_size_known}) into two 
steps:
\begin{equation}
    K^* \eqdef \argmax_K \left( \frac{1}{(K - \hat{K})!} \max_{p \in \mc{P}_K} 
    \perm\left(Q\right)\right)
    \label{eq:def_unlabeled_ML_alphabet_size_unknown}
\end{equation}
whenever the max over $K$ exists, where $\mc{P}_K \eqdef \{p \in \mc{P}: 
|\supp(p)| = K\}$.
The maximizer $K^*$ usually exists because 
increasing $K$ makes the first factor in 
(\ref{eq:def_unlabeled_ML_alphabet_size_unknown}) smaller, but makes the $K 
\times K$ matrix $Q$ 
bigger, boosting the number $K!$ of permutations to sum over in computing the 
permanent.

It may happen that $K^*$ does not exist\footnote{For example, if $n \geq 2$ and 
each symbol occurs exactly once in the sample, so $\hat{p}_x = \frac{1}{n}$ for 
all $x \in \mc{X}(\hat{p})$.}, in which case we are still able to define a PML 
distribution in terms of a discrete part and continuous part in the terminology 
of \cite{orlitsky2004modeling}.  \cite{orlitsky2004modeling} showed that for any 
fingerprint $\mc{F}$, there exists a distribution $p^*$ maximizing 
$\mb{P}_{p}(\mc{F})$, but this distribution may assign some symbols 
to the continuous part, and others to the discrete part with finite support.  
The intuition is that if there are sufficiently many symbols that appear 
exactly once in the sample, then the PML distribution $p^*$ assigns discrete 
probability 0 to infinitely many symbols to maximize the probability that each 
of them is seen only once.
Then we can define $K^*_\text{d}$ to be the optimal support set size of the 
discrete part of $p^*$.  \cite{orlitsky2004modeling} further derived conditions 
lower- and upper-bounding $K_\text{d}^*$ in terms of $\max_x \hat{p}_x$ and 
$\min_x \hat{p}_x$ and thus showing $K_\text{d}^*$ exists, and upper-bounding 
the size of the continuous part in terms of $\mc{F}_1$ (the numbers of symbols 
occurring once).  \cite{orlitsky2009PML} computed $p^*$ for all sample sizes up 
to $n = 7$.  




\subsection{Permutation group and multiple probability vectors} 
\label{sec.divergenceestimationpml}

Suppose we have two distributions $p$, $q$ on the same alphabet $\mc{X}$, such 
that $p_x + q_x > 0 \ \forall x \in \mc{X}$.
This condition ensures that there are no symbols $x$ such that $p_x = q_x = 0$, 
which simplifies the expressions below.
Draw $n$ samples i.i.d. from $p$ with empirical distribution $\hat{p}$ and draw 
$m$ samples i.i.d. from $q$ with empirical distribution $\hat{q}$. It is clear 
that the label-invariant properties of two distributions, such as the 
Kullback--Leibler divergence, $L_1$ distance, and the general family of 
divergences $\sum_{x\in \mathcal{X}} f(p_x,q_x)$, are invariant to the 
permutation group $\mc{S}_{\mathcal{X}}$ acting on pairs of distributions as 
$\sigma(p,q) \eqdef (\sigma p, \sigma q)$ for all $\sigma \in \mc{S}_\mc{X}$.  
Denote the 2-D fingerprint \cite{Raghunathan--Valiant--Zou2017estimating} by 
$\mc{F}(\hat{p},\hat{q})$, analogous to (\ref{eq:def_fingerprint}):
\beq
    \mc{F} = \mc{F}(\hat{p}, \hat{q}) = (\mc{F}_{i,j})_{i,j \geq 0} \eqdef 
    (|\{x \in \mc{X} : n \hat{p}_x = i, m \hat{q}_x = j|)_{i,j \geq 0}  
    \label{eq:def_fingerprint2d}
\eeq
The probability to draw the 2-D fingerprint under $p$, $q$ is:
\begin{align}
    \mb{P}_{p,q}(\mc{F}) &= \left(\prod_{i,j \geq 0} 
    \frac{1}{\mc{F}_{i,j}!}\right) \left(\begin{array}{c}n \\ n 
        \hat{p}\end{array}\right) \left(\begin{array}{c}m \\ m 
    \hat{q}\end{array}\right) \perm\bigg(\left(p_x^{n \hat{p}_{x'}} q_x^{m 
    \hat{q}_{x'}}\right)_{x,x' \in \mc{X}}\bigg) 
    \label{eq:prob_unlabeled_empirical_distribution_2_distributions}
\end{align}

Following the general profile maximum likelihood methodology, for a collection 
$\mc{P}$ of pairs of distributions on alphabet $\mc{X}$, the PML distributions 
$p^*$, $q^*$ are:
\begin{align}
    (p^*, q^*) &\eqdef \argmax_{(p,q)\in \mathcal{P}} \mb{P}_{p,q}(\mc{F}) 
    \label{eq:def_unlabeled_ML_alphabet_size_known_2D}
\end{align}
Note that finding the PML pair $(p^*, q^*)$ is not equivalent to solving two 
independent optimization problems since the matrix permanent in 
(\ref{eq:prob_unlabeled_empirical_distribution_2_distributions}) does not 
factor into two terms.
As in the 1D PML case, if the support set size $K = |\mc{X}|$ is unknown, then 
we can attempt to estimate it analogously to 
(\ref{eq:def_unlabeled_ML_alphabet_size_unknown}).  See Appendix 
\ref{app:dd_PML} for a precise statement.

Appendix \ref{app:dd_PML} generalizes the argument above in a straightforward 
manner to $D$-dimensional fingerprints 
\cite{Raghunathan--Valiant--Zou2017estimating}.  This allows estimation of 
functionals of more than two distributions.




%

\section{An approximation to the profile maximum likelihood distribution} 
\label{sec.approximatepml}

\subsection{Our approximate PML algorithm: an overview}

It's computationally challenging to compute the profile maximum likelihood 
distribution in all the cases discussed in 
Section~\ref{sec.discredistributions}. Indeed, it involves computing the matrix 
permanent in (\ref{eq:def_unlabeled_ML_alphabet_size_known}), and computing 
permanents is \#$P$-hard~\cite{Valiant79}. Since the computation of permanents 
is exponentially slow in the support set size $|\mc{X}|$ with the best-known 
algorithms, one may lower the target and hope to find an approximate efficient 
algorithm to all the profile maximum likelihood problems in 
Section~\ref{sec.discredistributions}. Some algorithms on approximate PML have 
been developed~\cite{Vontobel12, orlitsky2004modeling} in the past.

Here we develop a PML distribution approximation inspired by some empirical 
observations in the small examples where this distribution is computable 
exactly in reasonable time.  We provide a fast algorithm for computing this 
approximation.  The algorithm has no tunable parameters and runs in time at 
most linear in the sample size 
in the standard PML problem~(Section~\ref{sec.standardpml}), its run time 
usually dominated by the computation of the empirical distribution $\hat{p}$.
It is also efficient in solving divergence estimation problems as described in 
Section~\ref{sec.divergenceestimationpml}. 

Observations from solving the PML exactly in small cases (see Section 
\ref{sec:solving_PML_exactly})
yield the intuition that the PML distribution $p^*$ assigns equal mass to 
(``clumps together'') symbols whose empirical counts are close, differing on 
the order of $\sqrt{n}$, where $n$ is the sample size.  
For such a ``clumped'' distribution $p^*$, we further guess that the value of 
the matrix permanent in $\mb{P}_{p^*}(\mc{F}) \sim \perm(Q)/\mc{F}_0!$ 
(\ref{eq:prob_unlabeled_empirical_distribution})
is dominated by a subset of terms in the summation over the symmetric group 
$\mc{S}_\mc{X}$ -- namely the subgroup of permutations that only mix symbols 
within the same ``clump'' -- a level set of $p^*$ -- but not between different 
clumps.
Armed with this intuition, we replace maximization of the matrix permanent, 
which is hard, with maximization of a lower bound involving a sum over only 
this subgroup of permutations, which is much easier.

Let $\mc{A}(p) = \{\alpha\}$ be the partition of $\mc{X}$ into level sets of 
distribution $p$.  Then our lower bound $\bar{V}$ to the log permanent is 
defined by summing over a subgroup of the symmetric group, and satisfies (see 
derivation in Section \ref{sec:approx_PML_single_distribution}):
\begin{align}
    \log(\perm(Q)) &\geq \bar{V}(p) = \sum_{\alpha \in \mc{A}(p)} 
    \log(|\alpha|!) - n (D(\hat{p}||p) + H(\hat{p})) 
    \label{eq:permanent_lower_bound_preview}
\end{align}
where $D(\hat{p}||p) = \sum_{x \in \mc{X}} \hat{p}_x \log(\hat{p}_x / p_x)$ is 
the Kullback-Leibler divergence and $H(\hat{p}) = -\sum_{x \in \mc{X}} p_x 
\log(p_x)$ is the entropy of the empirical distribution.  The approximate PML 
(APML)
distribution is then the maximizer of the lower bound $\bar{V}$:
\begin{equation}
    \bar{p}^* \eqdef \argmax_{p \in \mc{P}} \bar{V}(p) 
    \label{eq:approx_PML_distribution_in_overview}
\end{equation}

The right hand side of (\ref{eq:permanent_lower_bound_preview}) yields some 
intuition for the approximate PML distribution.  Suppose $\mc{P}$ contains only 
distributions with support set size $K = |\mc{X}|$, so $\mc{F}_0 = K - 
|\hat{\mc{X}}|$ is fixed in (\ref{eq:def_unlabeled_ML_alphabet_size_known}), 
and the first term on the right hand side of 
(\ref{eq:permanent_lower_bound_preview}) encourages $\bar{p}^*$ to clump many 
symbols together into a few large clumps (since $\log(|\alpha|!) \sim |\alpha| 
\log(|\alpha|)$ is superlinear in $|\alpha|$), while the second term encourages 
$\bar{p}^*$ to be similar to the empirical distribution $\hat{p}$.  As the 
sample size $n \ra \infty$, the second term dominates, and we have $\bar{p}^* 
\ra \hat{p}$, consistent with our intuition for the large sample size limit and 
consistent with the result of~\cite{orlitsky2004modeling} (Theorem 16) for the 
exact PML distribution.

It turns out that computation of the approximate PML distribution $\bar{p}^*$ 
is equivalent to optimization over all partitions of $\mc{X}$.
This is because, as we show, any distribution $\bar{p}^*$ maximizing $\bar{V}$ 
satisfies an averaging property with respect to the empirical distribution 
$\hat{p}$ for all $x \in \mc{X}$: \begin{equation}
    \bar{p}^*_x = \frac{1}{|\alpha(x)|} \sum_{x' \in \alpha(x)} \hat{p}_{x'} 
    \label{eq:averaging_property}
\end{equation}
where $\alpha(x)$ is the level set of $p$ containing $x$.  Therefore 
$\bar{p}^*$ is determined by its partition of the alphabet $\mc{X}$ into level 
sets, so we can replace maximization of $\bar{V}$ over $p$ with maximization 
over all partitions $\mc{A}$.  Let $\bar{\mc{A}}^*$ denote the optimal 
partition:
\begin{equation}
        \bar{\mc{A}}^* \eqdef \argmax_{\mc{A}: \text{ partition of } \mc{X}} 
        \bar{V}(\mc{A})
\end{equation}
where $\bar{V}(\mc{A}) \eqdef \bar{V}(p_\mc{A}(\hat{p}))$ and 
$p_\mc{A}(\hat{p})$ is the distribution obtained by averaging the empirical 
distribution over the partition elements -- that is, satisfying property 
(\ref{eq:averaging_property}) for all $\alpha \in \mc{A}$ and $x \in \mc{X}$.  
The approximate PML distribution is then $\bar{p}^* = 
p_{\bar{\mc{A}}^*}(\hat{p})$.

%

We derive constraints on the optimal partition $\bar{\mc{A}}^*$ that enable an 
efficient
\footnote{Running in time $O(|\supp(\mc{F})|^2)$.  Any empirical distribution 
$\hat{p}$ satisfies $|\supp(\mc{F}(\hat{p}))| \leq (\sqrt{8 n + 1} + 1)/2$ (see 
discussion at end of Section 
\ref{sec:dynamic_programming_ML_unlabeled_approximation}), where $n$ is the 
sample size, so the running time is $O(n)$.  The run time of our algorithm is 
usually dominated by computation of $\hat{p}$.
} dynamic programming algorithm to compute $\bar{\mc{A}}^*$ and the approximate 
PML distribution $\bar{p}^*$.  This algorithm is presented in Section 
\ref{sec:dynamic_programming_ML_unlabeled_approximation}.


For the case of $D \geq 2$ probability vectors and $D$-dimensional fingerprints 
-- for example, when estimating the KL divergence, corresponding to $D = 2$ as 
in Section \ref{sec.divergenceestimationpml} -- we are unable to give an 
efficient algorithm for maximizing our lower bound $\bar{V}$ suitably 
generalized to the $D$-dimensional case.  The difficulty is that there is no 
natural ordering on $\mb{N}^D$ for $D \geq 2$, so the dynamic programming 
approach for $D=1$ does not work here.  In this case we settle for another 
layer of approximation, using a greedy heuristic to iteratively merge clumps of 
symbols:
we enlarge the first term in (\ref{eq:permanent_lower_bound_preview}) until the 
second term becomes too large.  This algorithm is presented in Appendix 
\ref{app:dd_PML}.


Figures \ref{fig:ML_unlabeled} and \ref{fig:fingerprint2d} show the approximate 
PML distributions for the cases $D=1$ and $D=2$, respectively.  For $D=1$, the 
approximate PML distribution is as defined in 
(\ref{eq:approx_PML_distribution_in_overview}).  For $D \geq 2$, we 
approximately maximize our lower bound $\bar{V}$ to the log matrix permanent as 
described in Appendix \ref{app:dd_PML}.  Figure \ref{fig:fingerprint3d} in 
Appendix \ref{app:dd_PML} shows the case $D=3$.


\begin{figure}[!h]
    \capstart
    \begin{center}
    \includegraphics[width=7in]{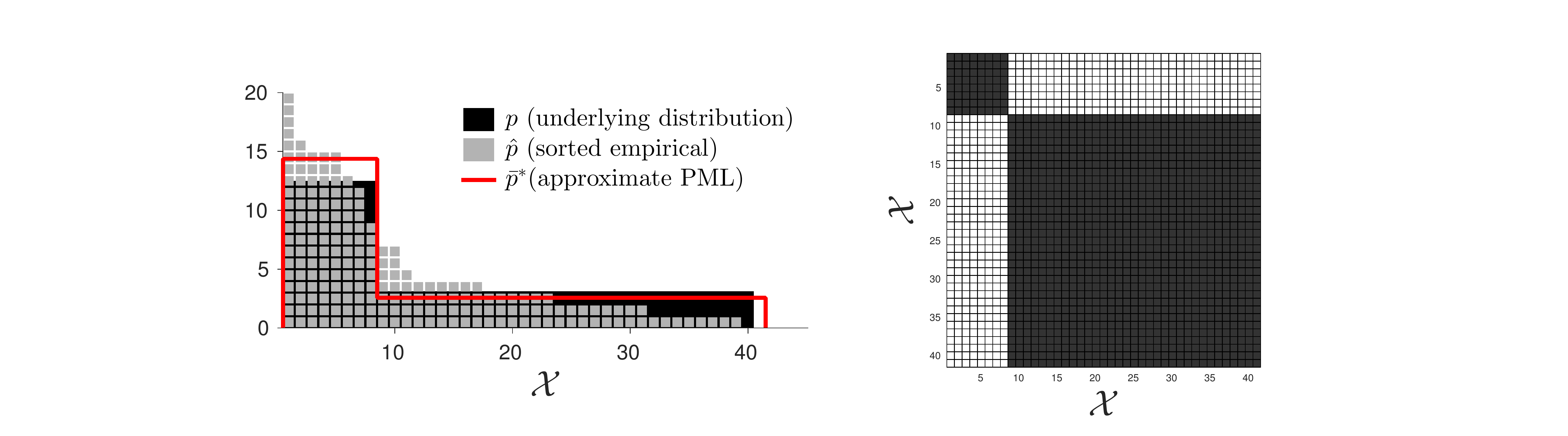}
        \caption{(left) An empirical histogram $\hat{p}$ sorted in 
        non-increasing order (gray squares) drawn from underlying distribution 
        $p$ (black area) and the approximate PML (APML) distribution 
        $\bar{p}^*$ (\ref{eq:approx_PML_distribution_in_overview}) (red line), 
        scaled by the sample size $n = 200$.  The alphabet is $\mc{X} = 
        \{1,\ldots,40\}$, assumed unknown in computing $\bar{p}^*$.  (right) 
        Computation of the log permanent lower bound $\bar{V}(\bar{p}^*)$ 
        (\ref{eq:permanent_lower_bound}) involves summing over all permutations 
        that mix symbols only within the level sets of $\bar{p}^*$, 
        corresponding to all permutation matrices with nonzero entries within 
        the black blocks.}
    \label{fig:ML_unlabeled}
    \end{center}
\end{figure}

\begin{figure}[!h]
    \capstart
    \begin{center}
    \includegraphics[width=8.4in]{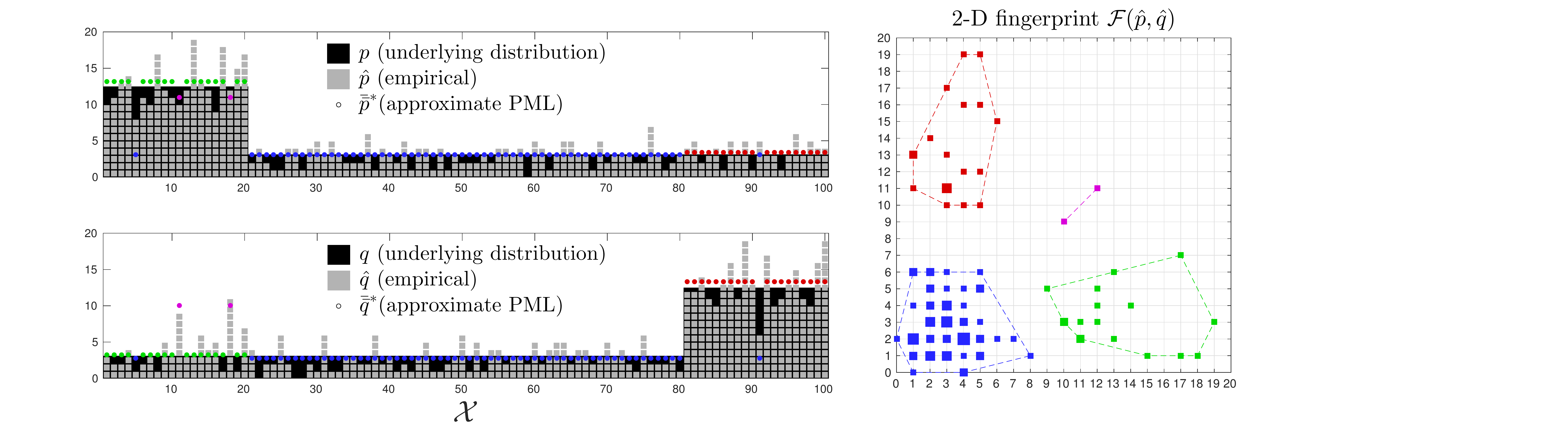}
        \caption{(Left) empirical distributions $\hat{p}$ and $\hat{q}$ (gray 
        squares) drawn from underlying distributions $p$ and $q$ (black area) 
        with $n = m = 400$ samples.  The approximate PML distributions 
        $(\bar{\bar{p}}^*, \bar{\bar{q}}^*)$ are computed by clumping entries 
        of a 2-D fingerprint according to the greedy heuristic described in 
        Appendix \ref{app:dd_PML}.  The approximate PML distributions are 
        ordered in the same way as the empirical distributions (with a distinct 
        color for each level set corresponding to the subplot on the right).  
        All distributions are plotted scaled by the sample size.  (Right) the 
        2-D fingerprint $\mathcal{F}_{i,j}$ at coordinates $(i,j)$.  Marker 
        areas are roughly proportional to $\mathcal{F}_{i,j}$.  All points 
        within the colored convex hulls correspond to a level set of the 
        approximate PML distributions.  We can see three large level sets 
        corresponding to the three distinct values of $(p_x, q_x)$ for $x \in 
        \mc{X}$.
        }
    \label{fig:fingerprint2d}
    \end{center}
\end{figure}

\subsection{Intuitions from solving the PML exactly} 
\label{sec:solving_PML_exactly}

For small alphabets, we can numerically and sometimes analytically optimize the 
permanent in (\ref{eq:def_unlabeled_ML_alphabet_size_known})
and (\ref{eq:def_unlabeled_ML_alphabet_size_known_2D}).   We observe that the 
PML distribution tends to assign equal mass to symbols whose empirical counts 
are close, differing on the order of $\sqrt{n}$, where $n$ is the sample size. This observation motivates our approximate PML scheme in Section 
\ref{sec:approx_PML_single_distribution}.


\subsubsection{Exact solution for size $2$ alphabet}

Suppose the alphabet size is $|\mc{X}| = 2$. Given sample size $n$ and 
empirical distribution $\hat{p} = (\hat{p}_1,\hat{p}_2) = 
(\hat{p}_1,1-\hat{p}_1)$, without loss of generality one may assume $\hat{p}_1 
\geq \frac{1}{n}$ (otherwise the PML is given by $\mathsf{Bern}(1)$) and 
$\hat{p}_1 \leq \frac{1}{2}$.  The exact PML distribution $\hat{p}^*$ was 
computed in~\cite{orlitsky2004modeling}:
\begin{theorem}\label{theorem.prasad} (\cite{orlitsky2004modeling})
    For all $\frac{1}{n} \leq \hat{p}_1 \leq \frac{1}{2}$,
\begin{align}
\hat{p}^* = \begin{cases}  \left( \frac{1}{2},\frac{1}{2} \right) & |\hat{p}_1 
-\hat{p}_2| \leq \frac{1}{\sqrt{n}} \\ \left(\frac{1}{1+p}, \frac{p}{1+p} 
\right) & |\hat{p}_1 - \hat{p}_2| > \frac{1}{\sqrt{n}} \end{cases}, 
    \label{eq:uniformity_1_distribution_alphabet_size_2}
\end{align}
where $p$ is the unique root in $(0,1)$ of the polynomial
\begin{align}
    \hat{p}_1 p^{n(1-2\hat{p}_1)+1} - \hat{p}_2 p^{n(1-2\hat{p}_1)} + \hat{p}_2 
    p -\hat{p}_1.
\end{align}
\end{theorem}

\begin{figure}[!h]
    \capstart
    \begin{center}
    \includegraphics[width=2.5in]{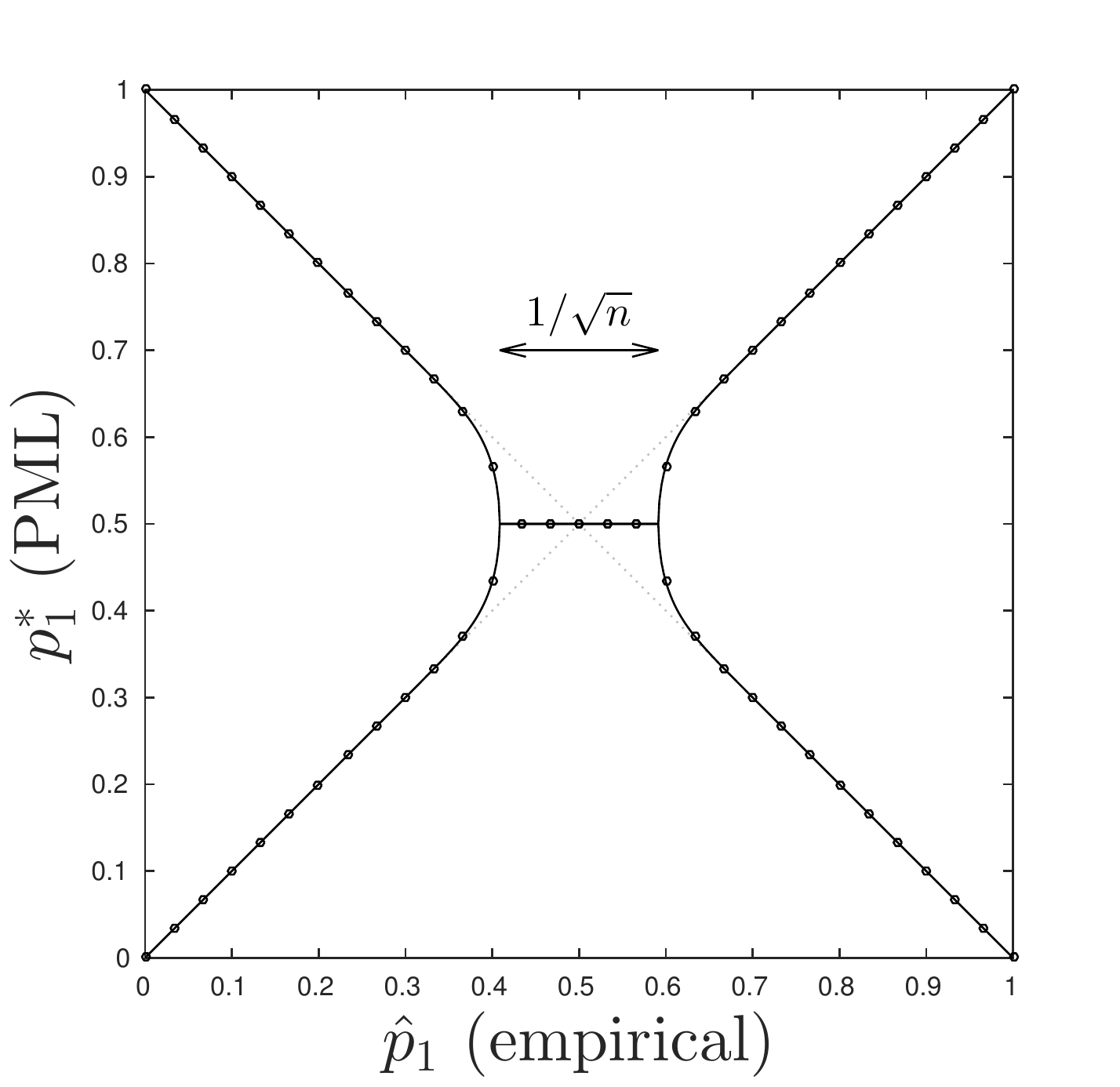}
        \caption{``Phase diagram'' for the PML distribution $p^* = (p^*_1, 
        1-p^*_1)$ on binary alphabet $\mc{X} = \{1,2\}$ and empirical 
        distribution $(\hat{p}_1, 1-\hat{p}_1)$ on $n = 30$ samples (plotting 
        the first component).
    Circles correspond to the possible histograms ($\hat{p}_1 \in 
        \{0/30,1/30,\ldots,30/30\}$).  Solid black lines correspond to the 
        solution for arbitrary $\hat{p}_1 \in [0,1]$.  The PML distribution 
        $p^*$ is uniform when $|\hat{p}_1 - \hat{p}_2| \leq 1/\sqrt{n}$.  
        Outside this middle region there are two PML branches, since any 
        permutation of the PML distribution is another PML distribution.  The 
        shapes of the branches were obtained numerically.  Diagonal dashed 
        lines correspond to the lines $p^*_1 = \hat{p}_1$ and $p^*_1 = 
        1-\hat{p}_1$.}
    \label{fig:UnlabeledML2bins}
    \end{center}
\end{figure}

Theorem~\ref{theorem.prasad} shows that $|\hat{p}_1 - \hat{p}_2| \leq 
\frac{1}{\sqrt{n}}
    \ \Leftrightarrow \ p^* \text{ is uniform} $, which confirms our intuition that the relative ranking of two bins is 
``resolvable'' if their empirical counts are more than about $\sqrt{n}$ apart, 
since the empirical counts are marginally binomially-distributed with standard 
deviation proportional to $\sqrt{n}$. Figure \ref{fig:UnlabeledML2bins} 
summarizes these observations for the size-$2$ alphabet.

\begin{figure}[!h]
    \capstart  
    \begin{center}
    \includegraphics[width=3.2in]{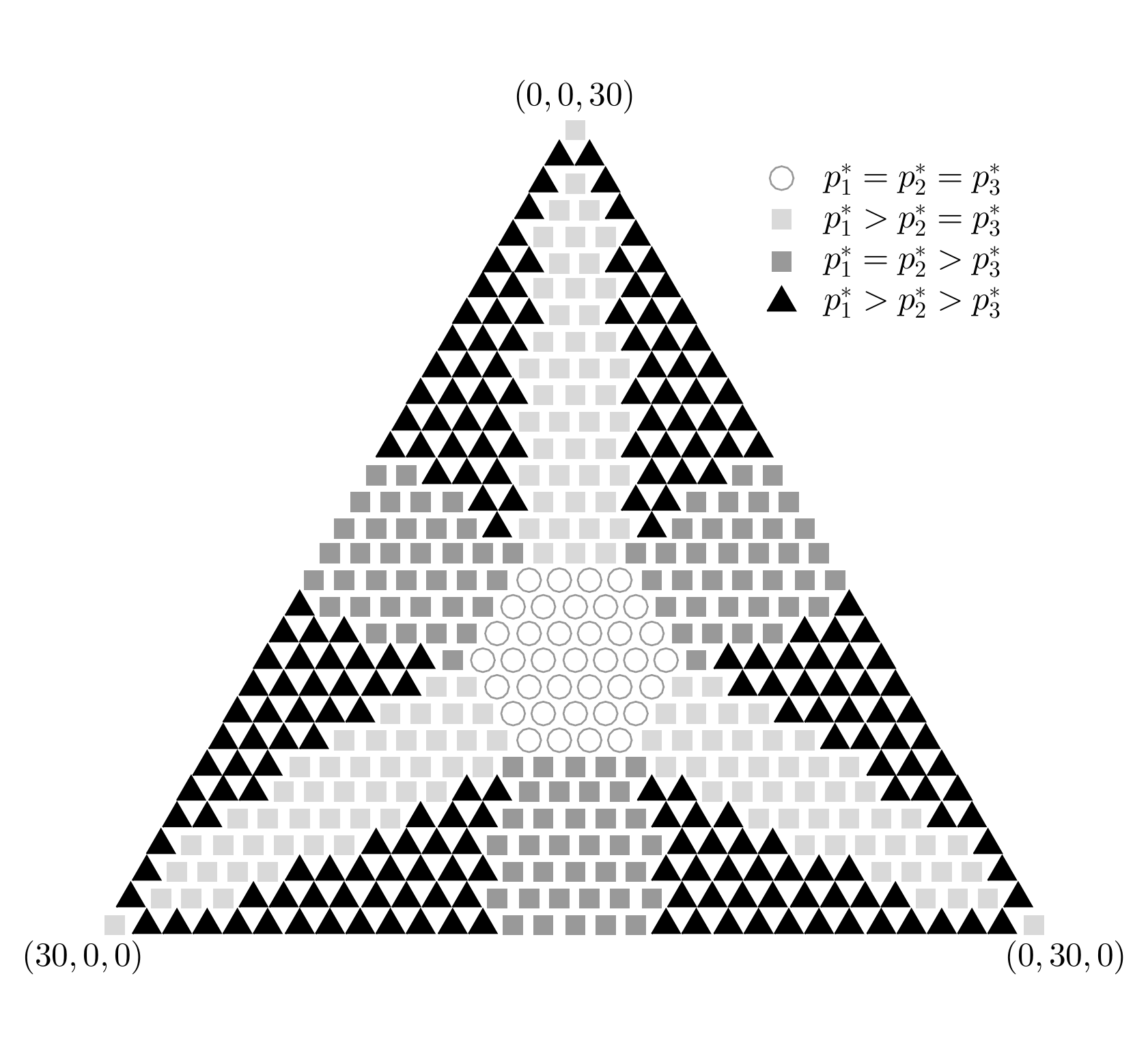}
    \caption{``Phase diagram'' for the PML distribution approximation 
        $p^*_\text{EM}$ for alphabet size $|\mc{X}| = 3$ bins and $n = 30$ 
        samples
        (analogous to Figure \ref{fig:UnlabeledML2bins}), computed by an 
        expectation-maximization (EM) algorithm as discussed in Section 
        \ref{sec:size_3_alphabet} and Appendix~\ref{app.emalgorithm}.  We plot 
        a marker for each possible empirical distribution -- a type -- on the 
        simplex.  Each type has three coordinates, which we then project onto 
        the 2-dimensional image shown (so, e.g., the uniform type $n \hat{p} = 
        (10,10,10)$ is in the center).  The shape of the marker at position $n 
        \hat{p}$ corresponds to the level set decomposition of the distribution 
        $p^*_\text{EM}$, with components sorted in non-increasing order.}
    \label{fig:UnlabeledML3bins}
    \end{center}
\end{figure}

\subsubsection{Size $3$ alphabet} \label{sec:size_3_alphabet}

Next let's consider the case of $|\mc{X}| = 3$ bins.  Given empirical 
distribution $\hat{p} = (\hat{p}_1,\hat{p}_2,\hat{p}_3)$, we use an 
expectation-maximization (EM) algorithm to numerically search for the PML 
distribution.  The EM algorithm starts from an initial point $p \leftarrow 
\hat{p}$ and converges to a local maximum of $\perm(Q)$ 
(\ref{eq:prob_unlabeled_empirical_distribution}).  We sort the components of 
this local maximum in non-increasing order and denote the result by 
$p^*_\text{EM}$.  Details of the EM algorithm can be found in 
Appendix~\ref{app.emalgorithm}.  We summarize our findings in Figure 
\ref{fig:UnlabeledML3bins}.  

\subsubsection{Exact solution for size $2$ alphabet with $D$ distributions}
Suppose we have $D$ distributions $((p_1^{(d)}, 1-p_1^{(d)}))_{d=1}^D$ on the 
same alphabet $\mc{X} = \{1,2\}$ and draw $(n_d)_{d=1}^D$ samples from each, 
obtaining empirical distributions
$((\hat{p}_1^{(d)}, 1-\hat{p}_1^{(d)}))_{d=1}^D$.  Then the $D$-dimensional PML 
distribution $(p^{*(d)})_{d=1}^D$ (defined for $D=2$ in Section 
\ref{sec.divergenceestimationpml} and for general $D$ in Appendix 
\ref{app:dd_PML}) is stated in Theorem \ref{thm:kd_PML_alphabet_size_2}.
\begin{theorem}\label{thm:kd_PML_alphabet_size_2} The $D$-tuple of PML 
    distributions $(p^{*(d)})_{d=1}^D$ on a binary alphabet satisfies:
\begin{equation}
    \sum_{d = 1}^D 4 n_d \left(\hat{p}_1^{(d)} - \frac{1}{2}\right)^2 > 1 \quad 
    \Rightarrow \quad (p^{*(d)})_{d=1}^D \neq \left(\left(\frac{1}{2}, 
    \frac{1}{2}\right)\right)_{d=1}^D 
    \label{eq:uniformity_D_distributions_alphabet_size_2}
\end{equation}
\end{theorem}
See Appendix \ref{app:condition_nonuniformity_kd_PML_binary_alphabet} for a 
proof.  We conjecture that the converse of statement 
(\ref{eq:uniformity_D_distributions_alphabet_size_2}) holds.

\begin{figure}[!h]
    \capstart
    \begin{center}
    \includegraphics[width=3in]{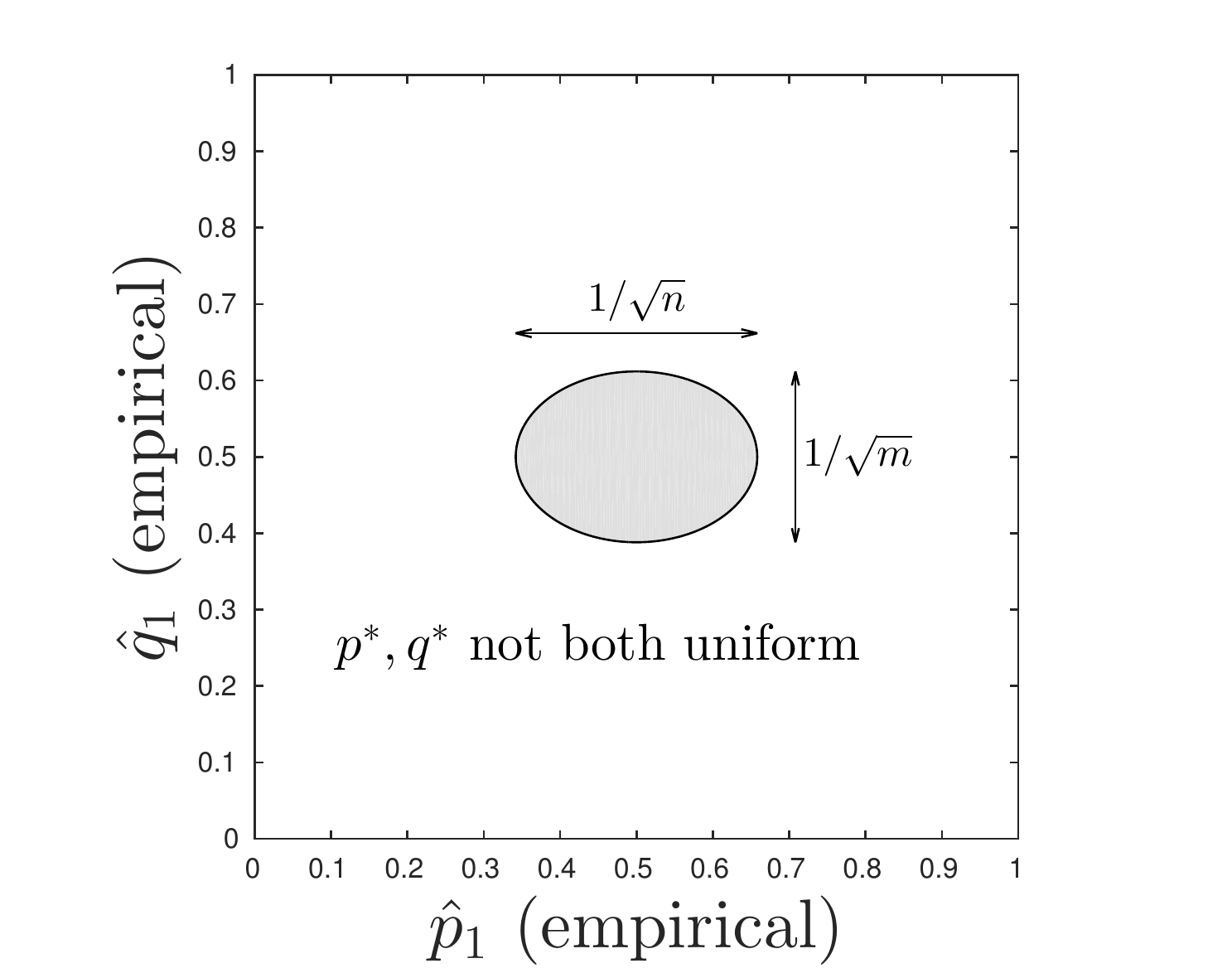}
        \caption{``Phase diagram'' for the 2-D PML pair of distributions $p^*, 
        q^*$ (\ref{eq:def_unlabeled_ML_alphabet_size_known_2D}) on binary 
        alphabet $\mc{X} = \{1,2\}$ and empirical distributions $(\hat{p}_1, 
        1-\hat{p}_1)$, $(\hat{q}_1, 1-\hat{q}_1)$ on $n = 10, m = 20$ samples.  
        $p^*, q^*$ are not both uniform whenever the empirical distribution 
        components $\hat{p}_1, \hat{q}_1$ both lie outside the shaded ellipse 
        $4 n \left(\hat{p}_1 - \frac{1}{2}\right)^2 + 
        4 m \left(\hat{q}_1 - \frac{1}{2}\right)^2 \leq 1$.
        }
    \label{fig:UnlabeledML2bins2distributions}
    \end{center}
\end{figure}

Theorem \ref{thm:kd_PML_alphabet_size_2}
extends the results of~\cite{orlitsky2004modeling} 
and~\cite{FernandesKashyap13} on the uniformity of the $D$-dimensional 
distribution beyond $D=1$.  The left side of 
(\ref{eq:uniformity_D_distributions_alphabet_size_2}) describes the interior of 
an ellipsoid in the space of distributions centered on $(\frac{1}{2})_{d=1}^D$, 
shown in Figure \ref{fig:UnlabeledML2bins2distributions} for the case $D=2$.
Note that (\ref{eq:uniformity_D_distributions_alphabet_size_2}) is not 
consistent with 
$D$ copies of condition (\ref{eq:uniformity_1_distribution_alphabet_size_2}) 
for $D \geq 2$; this reflects the fact that the probability of a 
$D$-dimensional fingerprint does not decompose into a product of $D$ terms (see 
(\ref{eq:prob_unlabeled_empirical_distribution_2_distributions}) for the case 
$D=2$ and Appendix \ref{app:dd_PML} for the general case).  

\subsubsection{Empirical observations of the exact solutions}
Figures~\ref{fig:UnlabeledML2bins}, ~\ref{fig:UnlabeledML3bins}, 
and~\ref{fig:UnlabeledML2bins2distributions} suggest that the PML distribution 
tends to ``cluster'' together similar entries of the empirical histogram 
$\hat{p}$, rather than smooth them out.
%
If $n |\hat{p}_x - \hat{p}_{x'}|$ is smaller than approximately $\sqrt{n} 
\hat{p}_x$, then the relative ranking of the true distribution's $x$ and $x'$ 
components is not well resolved statistically, and the permanent of $Q$ 
(\ref{eq:prob_unlabeled_empirical_distribution}) tends to be maximized with 
$p^*_x = p^*_{x'}$.  On the other hand if $n |\hat{p}_x - \hat{p}_{x'}|$ is 
larger than approximately $\sqrt{n} \hat{p}_x$, then it tends to happen that 
$p^*_x \neq p^*_{x'}$.  The similar phenomenon of a phase transition between a 
uniform and nonuniform PML distribution (for a different approximation of the 
PML \cite{Vontobel12}) has been reported by \cite{FernandesKashyap13, 
Chanetal15}.


These observations lead to the intuition that permutations $\sigma \in 
\mc{S}_\mc{X}$ that exchange sufficiently different components of $\hat{p}$ do 
not contribute much to the permanent $\perm(Q)$.
This intuition motivates our idea of simplifying the computation of the 
permanent by summing over only those permutations that contribute ``a lot'' to 
the sum.  Section \ref{sec:approx_PML_single_distribution} develops this idea.


\subsection{Approximate PML: the case of a single distribution} 
\label{sec:approx_PML_single_distribution}




In the sequel, denote by $\mc{A}(p) = \{\alpha\}$ the partition of $\mc{X}$ 
into the level sets of $p$:
\begin{equation}
    \mc{A}(p) \eqdef \left\{ \{x \in \mc{X}: p_x = u\} : u \in \{p_x : x \in \mc{X}\} \right\} \label{eq:def_partition_into_level_sets_of_p}
\end{equation}



In this Section we assume that the underlying distribution $p$ is supported on 
finite set $\mc{X}$ of known cardinality.  The case of unknown or infinite 
support set size is treated in Section 
\ref{sec:estimate_support_set_size_approximate_PML}.
If the support set size is known, then so is the number of ``unseen'' symbols 
$\mc{F}_0$, so from (\ref{eq:def_unlabeled_ML_alphabet_size_known}) we conclude 
that the PML distribution $p^*$ is a maximizer of the function:
\begin{align}
    V(p) &\eqdef \log(\perm(Q)) \label{eq:permanent_reexpressed}
\end{align}

We lower bound the log permanent $V(p)$ by summing over a subset of the terms 
in the summation -- a subgroup of the symmetric group $\mc{S}_\mc{X}$.  Denote 
the lower bound by $\bar{V}(p)$:
\beq
    \bar{V}(p) \eqdef \log\left(\sum_{\sigma \in \mc{S}_{\mc{X},p}} \prod_{x 
    \in \mc{X}} p_x^{n \hat{p}_{\sigma(x)}}\right) \leq \log\left(\sum_{\sigma 
    \in \mc{S}_\mc{X}} \prod_{x \in \mc{X}} p_x^{n \hat{p}_{\sigma(x)}}\right) 
    = V(p) \label{eq:permanent_lower_bound}
\eeq
where $\mc{S}_{\mc{X},p}$ is a subgroup of $\mc{S}_\mc{X}$ consisting of all 
permutations that exchange only those alphabet symbols $x, x' \in \mc{X}$ that 
are within the same level set of $p$.  Equivalently, $\mc{S}_{\mc{X},p}$ is the 
stabilizer of distribution $p$ under the action of relabeling its components: 
\begin{align}
    \mc{S}_{\mc{X},p} &\eqdef \{\sigma \in \mc{S}_\mc{X} : \sigma p = p\} \cong 
    \bigtimes_{\alpha \in \mc{A}(p)} \mc{S}_{\alpha} 
    \label{eq:def_subgroup_of_symmetric_group}
\end{align}
where $(\sigma p)_x \eqdef p_{\sigma(x)}$, $\cong$ denotes group isomorphism, 
$\times$ denotes the direct product of groups, and $\mc{S}_{\alpha}$ is the 
symmetric group acting on level set $\alpha \subset \mc{X}$.  Then 
$|\mc{S}_{\mc{X},p}| = \prod_{\alpha \in \mc{A}(p)} |\alpha|!$.

Our approximate profile maximum likelihood distribution $\bar{p}^*$ maximizes 
the lower bound $\bar{V}(p)$.  In other words, for the case of every 
distribution in collection $\mathcal{P}$ having the same support set size,
\beq
    \bar{p}^* \eqdef \argmax_{p \in \mc{P}} \bar{V}(p) 
    \label{eq:def_approximate_unlabeled_ML}
\eeq
The case of unknown support set size is treated in Section 
\ref{sec:estimate_support_set_size_approximate_PML}.

Some intuition for the computational difficulty of the proposed approximation:
the product structure (\ref{eq:def_subgroup_of_symmetric_group}) of 
$\mc{S}_{\mc{X},p}$ allows us to lower bound the log matrix permanent $V(p) = 
\log(\perm(Q)) \geq \log(\perm(\bar{Q})) = \bar{V}(p)$, where $\bar{Q}$ is the 
block-diagonal matrix $\bar{Q}_{x,x'} = Q_{x,x'} \mathbbm{1}(\hat{p}_x = 
\hat{p}_{x'})$
whose permanent is the product of the permanents in each block.  
Since $p$ is constant within each block, the permanent of each block is easy to 
evaluate, so evaluation of the lower bound $\bar{V}(p)$ is dramatically simpler 
than evaluation of $V(p)$, taking time $O(|\mc{X}|)$.  It moreover turns out to 
be possible to optimize $\bar{V}(p)$ computationally efficiently to find the 
approximate PML distribution $\bar{p}^*$; this is done below.


The restriction to summing over $\mc{S}_{\mc{X},p}$ rather than $\mc{S}_\mc{X}$ 
seems large, but the hope is that $\bar{p}^*$ clusters together only comparably 
frequent symbols
(that is, $\bar{p}^*_x = \bar{p}^*_{x'}$ implies $n |\hat{p}_x - \hat{p}_{x'}|$ 
is less than about $\sqrt{n} \hat{p}_x$), so we hope that $\bar{V}(\bar{p}^*) 
\approx V(\bar{p}^*)$ and that $\bar{p}^*$ is close to $p^*$, the exact profile 
maximum likelihood (\ref{eq:def_unlabeled_ML_alphabet_size_known}).
Figure \ref{fig:ML_unlabeled} shows that the approximate PML distribution 
$\bar{p}^*$ clustering together similar enough bins of $\hat{p}$.  We thus 
expect our approximate PML scheme to perform best when data is drawn from a 
distribution with a few well-separated level sets (like a uniform distribution 
or mixture of uniforms) and less well for more smoothly-varying distributions 
(like a Zipf distribution); this intuition is borne by the numerical 
experiments of Section \ref{sec.standardpmlplugin}.



We can rewrite our log permanent lower bound $\bar{V}(p)$ 
(\ref{eq:permanent_lower_bound}) as (see proof in Appendix~\ref{app:permanent_lower_bound_reexpressed}):
\begin{align}
    \bar{V}(p) &= \log\left(\sum_{\sigma \in \mc{S}_{\mc{X},p}} \prod_{x \in 
    \mc{X}} p_x^{n \hat{p}_{\sigma(x)}}\right) \\
    &= - n \big(D(\hat{p}||p) + H(\hat{p})\big) + \sum_{\alpha \in \mc{A}(p)} 
    \log(|\alpha|!) \label{eq:permanent_lower_bound_reexpressed}
\end{align}
where $\mc{A}(p)$ (\ref{eq:def_partition_into_level_sets_of_p}) is the 
partition of $\mc{X}$ into level sets of $p$, $D(\hat{p}||p)$ is the 
Kullback-Leibler divergence and $H(\hat{p})$ is the entropy of the empirical 
distribution.  

Expression (\ref{eq:permanent_lower_bound_reexpressed}) yields some intuition 
about the approximate PML distribution $\bar{p}^*$ -- a maximizer of 
$\bar{V}(p)$.
A distribution $p$ that ``clumps'' many symbols together (that is, has a few 
large level sets $\alpha$) boosts the second term (the summation over $\alpha 
\in \mc{A}(p)$) in (\ref{eq:permanent_lower_bound_reexpressed}), but is very 
different from $\hat{p}$, thus lowering the first term.  On the other hand, by 
setting $p = \hat{p}$, we maximize the first term, but reduce the second term.  
As the sample size $n \ra \infty$, the contribution of the second term in 
(\ref{eq:permanent_lower_bound_reexpressed}) vanishes relative to the first 
term, and we have $\bar{p}^* \rightarrow \hat{p}$, the ML distribution, in this 
limit.



Section \ref{sec:properties_of_approx_PML} establishes some properties of the 
approximate PML solution $\bar{p}^*$ (\ref{eq:def_approximate_unlabeled_ML}) 
that enable us to compute it efficiently via a dynamic programming algorithm in 
Section \ref{sec:dynamic_programming_ML_unlabeled_approximation}.

\subsection{Properties of the approximate PML distribution} 
\label{sec:properties_of_approx_PML}

Let $\alpha \subset \mc{X}$ and denote by $\hat{p}_\alpha$ the average value of 
the empirical distribution $\hat{p}$ over $\alpha$:
\begin{equation}
    \hat{p}_\alpha \eqdef \frac{1}{|\alpha|} \sum_{x \in \alpha} \hat{p}_x 
    \label{eq:def_averaging_distribution_over_set}
\end{equation}

We can show that the PML distribution approximation $\bar{p}^*$ 
(\ref{eq:def_approximate_unlabeled_ML}) satisfies for all $x \in \mc{X}$
\begin{equation}
    \bar{p}^*_x = \hat{p}_{\alpha(x)}
    \label{eq:averaging_condition_ML_unlabeled_approximation}
\end{equation}
where $\alpha(x) = \mc{X}_{\bar{p}^*_x}$ is the level set of $\bar{p}^*$ 
containing $x$.
That is, $\bar{p}^*_x$ 
(\ref{eq:averaging_condition_ML_unlabeled_approximation}) is equal to the 
average value of the empirical histogram $\hat{p}$ over the symbols clumped 
with $x$ into the same level set of $\bar{p}^*$. This follows from the fact 
that $D(\hat{p} \| p)$ is a Bregman divergence~\cite[Lemma 
4]{jiao2017relations}.

The \textit{averaging} property 
(\ref{eq:averaging_condition_ML_unlabeled_approximation}) determines our 
approximate PML distribution $\bar{p}^*$ in terms of its partition of $\mc{X}$ 
into level sets.  Therefore, instead of maximizing $\bar{V}(p)$, we maximize 
$\bar{V}(\mc{A})$, defined:
\begin{align}
    \bar{V}(\mc{A}) \eqdef \sum_{\alpha \in \mc{A}} \bar{V}(\alpha) 
    \label{eq:permanent_lower_bound_reexpressed2}
\end{align}
where for $\alpha \subset \mc{X}$
\begin{equation}
        \bar{V}(\alpha) \eqdef \log\left(|\alpha|!\right) + n |\alpha| 
        \hat{p}_\alpha \log(\hat{p}_\alpha) \label{eq:partition_element_value}
\end{equation}
We can check that $\bar{V}(\mc{A}) = \bar{V}(\hat{p}_\mc{A})$, where 
$\hat{p}_\mc{A}$ is the distribution obtained by averaging the empirical 
distribution $\hat{p}$ within each partition element $\alpha \in \mc{A}$.  Let 
$\bar{\mc{A}}^*$ denote the optimal partition:
\begin{equation}
        \bar{\mc{A}}^* \eqdef \argmax_{\mc{A}: \text{ partition of } \mc{X}} 
        \bar{V}(\mc{A}) \label{eq:def_optimal_partition_approx_PML}
\end{equation}
Now optimizing the lower bound to the permanent $\bar{V}(p)$ 
(\ref{eq:permanent_lower_bound}) is equivalent to optimizing $\bar{V}(\mc{A})$
(\ref{eq:permanent_lower_bound_reexpressed2}) over partitions of $\mc{X}$, 
since $\bar{V}(\bar{p}^*) = \bar{V}(\bar{\mc{A}}^*)$ and $\bar{p}^* = 
\hat{p}_{\bar{\mc{A}}^*}$.


We make two observations for the solution of the approximate PML problem 
(\ref{eq:def_optimal_partition_approx_PML}) that allow us to restrict the set 
of partitions over which we optimize.  Let $\mc{A}$ be a partition of $\mc{X}$.  
We say $\mc{A}$ has the \textit{iso-clumping} or \textit{convexity} properties 
defined below:
\begin{enumerate}
    \item \textit{Iso-clumping} property.  Symbols with the same empirical 
        probabilities are clumped together:  For all $x, x' \in \mc{X}$:
    \begin{equation}
        \hat{p}_x = \hat{p}_{x'} \Rightarrow \alpha(x) = \alpha(x')
        \label{eq:same_empirical_prob_symbols_clumped_together_property}
    \end{equation}
        Where $\alpha(x) \in \mc{A}$ is the partition element containing $x$.
    \item \textit{Convexity} property.  For all $x, x', x'' \in \mc{X}$:
    \begin{equation}
        \Big(\hat{p}_x < \hat{p}_{x'} < \hat{p}_{x''} \text{ and } 
        \alpha(x) = \alpha(x'')\Big) \Rightarrow \alpha(x) = \alpha(x') = 
        \alpha(x'') \label{eq:convexity_property}
    \end{equation}
\end{enumerate}

\begin{theorem} \label{thm:partition_properties}
    Let $\bar{\mc{A}}^*$ (\ref{eq:def_optimal_partition_approx_PML}) be the 
    partition of $\mc{X}$ into level sets of the approximate PML distribution 
    $\bar{p}^*$ (\ref{eq:def_approximate_unlabeled_ML}) and let $\hat{p}$ be 
    the empirical distribution.  Then $\bar{\mc{A}}^*$ has the iso-clumping and 
    convexity properties 
    (\ref{eq:same_empirical_prob_symbols_clumped_together_property}) and 
    (\ref{eq:convexity_property}).
\end{theorem}
See Appendix \ref{app:partition_properties_proof} for a proof.

\subsection{A dynamic programming computation of the approximate PML 
distribution} \label{sec:dynamic_programming_ML_unlabeled_approximation}





Theorem \ref{thm:partition_properties} lets us efficiently maximize 
$\bar{V}(\mc{A})$ (\ref{eq:permanent_lower_bound_reexpressed2}) by restricting 
to only those partitions of $\mc{X}$ that have the iso-clumping and convexity 
properties.  Once we find $\bar{\mc{A}}^*$ 
(\ref{eq:def_optimal_partition_approx_PML}), the averaging property 
(\ref{eq:averaging_condition_ML_unlabeled_approximation}) lets us compute the 
approximate PML distribution $\bar{p}^*$.


Let $\mc{F}_+ \eqdef (\mc{F}_i)_{i \geq 1}$.  Let 
\begin{equation}
    0 = m_0 < m_1 < m_2 < \ldots < m_{F_+} \leq n
\end{equation}
where 
\begin{equation}
    \supp(\mc{F}_+) = \{m \geq 1 : \mc{F}_m > 0\} = \{m_i\}_{i=1}^{F_+}
\end{equation}




Then any distribution whose level set partition satisfies the iso-clumping 
(\ref{eq:same_empirical_prob_symbols_clumped_together_property}) and convexity
(\ref{eq:convexity_property}) properties
has all level sets of the form $\mc{X}_{i:j}$ with $0 \leq i \leq j \leq F_+$:
\begin{align}
    \mc{X}_{i:j} &\eqdef \{x \in \mc{X}: m_i \leq n \hat{p}_x \leq m_j\} 
    \label{eq:composition_level_set_bijection}.
\end{align}
Thus we optimize $\bar{V}(\mc{A})$ 
(\ref{eq:permanent_lower_bound_reexpressed2}) over partitions of $\mc{X}$ into 
level sets of the form (\ref{eq:composition_level_set_bijection}).  Note that 
if $\mc{F}_0 > 0$, then the set $\mc{X}_{0:0} = \{x \in \mc{X}: \hat{p}_x = 
    0\}$ of unseen symbols can not appear in the optimal partition 
    $\bar{\mc{A}}^*$ because if it does, then its probability mass under 
    $p_{\bar{\mc{A}}^*}$ is 0, violating our assumption that $p$ has support 
    $\mc{X}$.  If $\mc{F}_0 = 0$, then $\mc{X}_{0:0} = \emptyset$.




This optimization can be done by a dynamic programming algorithm.  For integers 
$i \leq j$, let $[i,j] \eqdef \{k \in \mb{N}: i \leq k \leq j\}$.  Let 
$\bar{V}_i$ denote for $0 \leq i \leq F_+$:
\begin{align}
    \bar{V}_i &\eqdef \max_{j \in [i\vee 1,F_+]}(\bar{V}(\mc{X}_{i:j}) + 
    \bar{V}_{j+1}) \label{eq:dynamic_programming_update} \\
    &\overset{\text{(a)}}{=} \max_{\mc{A}:\text{ partition of } \mc{X}_{i:F_+}} 
    \bar{V}(\mc{A}) 
    \label{eq:dynamic_programming_update_recursive_partition_property}
\end{align}
with boundary condition $\bar{V}_{F_++1} \eqdef 0$, where $i \vee j$ denotes 
the greater of $i$ and $j$, and where $\bar{V}(\mc{X}_{i:j})$ is as in 
(\ref{eq:partition_element_value}).  (a) follows by induction on $i$ downwards 
from $F_+$.
We let $j \in [i \vee 1, F_+]$ in (\ref{eq:dynamic_programming_update}) (rather 
than $j \in [i, F_+]$)
in order to optimize over only those partitions of $\mc{X}$ that do not contain 
the set $\mc{X}_{0:0}$ of unseen symbols; this restriction forces the 
approximate PML distribution to have support set size $|\mc{X}|$, as we assumed 
in the beginning of Section \ref{sec:approx_PML_single_distribution}.

Then the PML distribution approximation $\bar{p}^* = \argmax_p \bar{V}(p)$ 
satisfies (setting $i = 0$ and using $\mc{X}_{0:F_+} = \mc{X}$ in 
(\ref{eq:dynamic_programming_update_recursive_partition_property})):
\beq
    \bar{V}(\bar{p}^*) = \max_{\mc{A}:\text{ partition of } \mc{X}}
    \bar{V}(\mc{A}) = \bar{V}_0
    \label{eq:dynamic_programming_value}
\eeq

We can compute the term $\bar{V}(\mc{X}_{i:j})$ in 
(\ref{eq:dynamic_programming_update}), corresponding to clustering all symbols 
of $\mc{X}_{i:j}$ into a level set of $p$, using 
(\ref{eq:partition_element_value}):
\begin{align}
    \bar{V}(\mc{X}_{i:j}) &= \log\left(|\mc{X}_{i:j}|!\right) + n 
    |\mc{X}_{i:j}| \hat{p}_{\mc{X}_{i:j}} 
    \log\left(\hat{p}_{\mc{X}_{i:j}}\right) \label{eq:V_ij} \\
    &= \log\left(\Big(\sum_{k=i}^j \mc{F}_{m_k}\Big)!\right) + \left(\sum_{k = 
    i}^j m_k \mc{F}_{m_k}\right) \log \left(\frac{\sum_{k = i}^j m_k 
    \mc{F}_{m_k}}{n \sum_{k=i}^j \mc{F}_{m_k}}\right) \label{eq:V_ij_value}
\end{align}
where we used $|\mc{X}_{i:j}| = \sum_{k=i}^j \mc{F}_{m_k}$ and 
$\hat{p}_{\mc{X}_{i:j}} = \frac{1}{n|\mc{X}_{i:j}|} \sum_{k=i}^j m_k 
\mc{F}_{m_k}$.


Relations (\ref{eq:dynamic_programming_update}) and (\ref{eq:V_ij_value}) allow 
us to compute the level set decomposition $\bar{p}^*$ by filling out a $(F_++1) 
\times F_+$ array $(\bar{V}(\mc{X}_{i:j}))_{i\in[0,F_+],j\in[1,F_+]}$ and 
keeping track of the maximizing index each time we compute $\bar{V}_i$.  Once 
we have the optimal level set decomposition $\bar{\mc{A}}^* = \{\mc{X}_{i:j}\}$ 
of $\bar{p}^*$, we set for all $x \in \mc{X}_{i:j}$, for all $i$, $j$
\begin{equation}
    \bar{p}^*_x = \hat{p}_{\mc{X}_{i,j}}
\end{equation}


For example, in Figure \ref{fig:ML_unlabeled}, we have $(m_i)_{i=0}^{F_+=12} = 
(0, 1, 2, 3, 4, 5, 7, 9, 12, 13, 15, 16, 20)$ and the approximate PML 
distribution $\bar{p}^*$ (red line) has level sets $\{\mc{X}_{0:6}, 
\mc{X}_{7:12}\}$.

The running time of this dynamic programming algorithm is 
$O(|\supp(\mc{F})|^2)$, where $|\supp(\mc{F})| \leq F_+ + 1$.
In terms of the sample size $n$, any empirical distribution $\hat{p}$ satisfies 
$|\supp(\mc{F}(\hat{p}))| \leq \frac{1}{2}\left(1+\sqrt{8n+1}\right)$, with 
equality achieved by the empirical distribution $\hat{p}_i = \frac{2 
i}{|\mc{X}|(|\mc{X}|+1)}$ for $i \in [0,|\mc{X}|]$ with $|\mc{X}| = 
|\supp(\mc{F})|$ and $n = \frac{1}{2} |\mc{X}| (|\mc{X}| + 1)$.  Thus the 
worst-case run time for the dynamic programming algorithm is $O(\sqrt{n}^2) = 
O(n)$.  We ``usually'' have $|\supp(\mc{F})|$ much smaller than $\sqrt{n}$, so 
this is a pessimistic estimate for a typical case.  The run time of our 
approximate PML scheme is usually dominated by computation of $\hat{p}$.
An implementation optimization is to pre-compute the sums $\sum_{k=i}^j 
\mc{F}_{m_k}$ and $\sum_{k=i}^j m_k \mc{F}_{m_k}$ for all $i \leq j$ in 
computing $\bar{V}(\mc{X}_{i,j})$.

\subsection{Unknown or infinite support set size} 
\label{sec:estimate_support_set_size_approximate_PML}

If the support set size $K = |\mc{X}|$ is unknown, then we can attempt to infer 
it.  
Our estimator $\bar{K}^*$ for the support set size is the same as 
(\ref{eq:def_unlabeled_ML_alphabet_size_unknown}), but replaces the permanent 
with our lower bound (\ref{eq:permanent_lower_bound}), $e^{\bar{V}(p)} \leq 
\perm(Q)$:
\begin{equation}
    \bar{K}^* \eqdef \argmax_K \left(\frac{1}{(K - \hat{K})!} \max_{p \in 
    \mc{P}_K} e^{\bar{V}(p)}\right) 
    \label{eq:def_unlabeled_ML_alphabet_size_unknown_approx_PML}
\end{equation}
whenever the max over $K$ exists, where $\mc{P}_K \eqdef \{p \in \mc{P}: 
|\supp(p)| = K\}$, where $\hat{K} = |\hat{\mc{X}}|$ is the support set size of 
the empirical distribution $\hat{p}$.  Then the PML distribution is $\bar{p}^*$ 
with support set $\bar{K}^*$.  The max over $p$ is obtained via our approximate 
PML algorithm in Section 
\ref{sec:dynamic_programming_ML_unlabeled_approximation}.  The maximizer 
$\bar{K}^*$ usually exists because a larger $K$ boosts the value of the log 
permanent lower bound $\bar{V}(p)$ (since there are more permutations to sum 
over), but reduces the first factor in 
(\ref{eq:def_unlabeled_ML_alphabet_size_unknown_approx_PML}).
It may happen that $\bar{K}^*$ does not exist, in which case we say our 
approximate PML distribution has a continuous part in the terminology of 
\cite{orlitsky2004modeling}.  
In general the function in the argmax in 
(\ref{eq:def_unlabeled_ML_alphabet_size_unknown_approx_PML}) is multimodal in 
$K$.

It turns out we can efficiently compute $\bar{K}^*$ if it exists and compute 
the corresponding approximate PML distribution; if $\bar{K}^*$ does not exist, 
then we can efficiently detect this case and compute the corresponding 
approximate PML distribution, which in this case has a continuous part.  


%
%
%

We state and derive our observations in Appendix 
\ref{app:estimate_support_set_size_approximate_PML}.
The idea is to optimize over all possible ways to clump the unseen symbols 
$\mc{X}_{0:0}$ with the other symbols.  Since the optimal clumping satisfies 
the iso-clumping and convexity properties, we show that there are only 
$|\supp(\mc{F}_+)| \leq \sqrt{2n}+1$ possibilities to check and that the value 
of each possibility can be checked quickly, reusing the work done in the 
dynamic programming approach described in the previous Section, so this can be 
done efficiently. 

\subsection{Approximate PML: the case of multiple distributions (see Appendix 
\ref{app:dd_PML})}\label{sec:dd_PML}

We generalize the PML distribution and approximate PML distribution to the case 
of $(n_d)_{d=1}^D$ samples drawn from distributions $(p^{(d)})_{d=1}^D$ on the 
same alphabet (see Section \ref{sec.divergenceestimationpml} for the case 
$D=2$, used in estimating functionals of pairs of distributions).  This 
requires some more notation that we delegate to Appendix \ref{app:dd_PML}.

For $D \geq 2$, we are unable to give an efficient algorithm to compute the 
$D$-dimensional approximate PML.  The difficulty is due to the lack of a 
natural ordering on $\mb{N}^D$ for $D \geq 2$, so the dynamic programming 
approach we use for $D=1$ does not work for $D \geq 2$.  We settle for 
approximately computing our the approximate PML distribution via a greedy 
heuristic: we iteratively enlarge level sets of a candidate solution by merging 
pairs of distinct level sets until the objective function $\bar{V}$ (suitably 
generalized to $D$ distributions) stops increasing.  See Appendix 
\ref{app:dd_PML} for details.

\section{Estimating symmetric functionals of a single discrete distribution} 
\label{sec.standardpmlplugin}

In the standard PML setting discussed in Section~\ref{sec.standardpml}, we 
argued the sufficiency of the fingerprint $\mc{F}$~(\ref{eq:def_fingerprint}) 
in estimating the sorted probability vector of a discrete distribution, which 
implies the sufficiency of the fingerprint for estimating any functional $F$ of 
the sorted probability vector of the form
\begin{align}
    F(p) = G\Big(\sum_{x\in \mathcal{X}} f(p_x)\Big), \label{eq:F_functional}
\end{align}
where we constrain $f(0) = 0$ to accommodate the unknown alphabet setting. Note that the fingerprint is also sufficient for functionals of type $\sum_{x,y \in \mathcal{X}} f(p_x,p_y)$, etc. 

In order to apply Theorem~\ref{theorem.ml}, one also needs to show the cardinality 
of the fingerprint $\mathcal{F}$ is small. The fingerprint $\mathcal{F} = 
(\mathcal{F}_i)_{i\geq 0}$ satisfies
\begin{align}
\sum_{i\geq 0} i \mathcal{F}_i = n,
\end{align}
and each fingerprint corresponds to an unordered integer partition of the integer $n$. We now recall the Hardy--Ramanujan result on integer partitions.
\begin{theorem}\label{theorem.hardyramanujan}\cite{hardy1918asymptotic}\cite[Theorem 
    2]{barany1992number} The cardinality of the set of
    fingerprints~(\ref{eq:def_fingerprint}) on $n$ samples is given by
\begin{align}
    e^{\pi \sqrt{\frac{2n}{3}}(1-o(1))} & \leq   |\{(\mathcal{F}_i)_{i\geq 
    1}\}| \leq e^{\pi \sqrt{\frac{2n}{3}}}.
\end{align}
\end{theorem}
The key observation is that the cardinality of the fingerprint is sub-exponential. Then, combining with Theorem~\ref{theorem.ml} and the fact that there exist estimators for various symmetric functionals with near exponential measure concentration, \cite{acharya2016unified} showed that plugging in the PML achieves the optimal sample complexity in estimating the Shannon entropy, the support size, the support coverage, and the $L_1$ distance to uniformity. 

In this section, we extensively test the performance of symmetric functional 
estimation via plugging in our approximate PML (APML) distribution into the 
functionals.  Recall that the PML distribution $p^*$ is defined as in 
(\ref{eq:def_unlabeled_ML_alphabet_size_known}) for empirical distribution 
$\hat{p}$ on $n$ samples with fingerprint $\mc{F}(\hat{p})$:
\begin{align}
    p^* & = \argmax_{p\in \mc{P}} \mathbb{P}_p \left( \mc{F} \right) = 
    \argmax_{p \in \mc{P}} \frac{\perm\left(\left(\begin{array}{c} p_x^{n 
    \hat{p}_{x'}} \end{array}\right)_{x,x'\in\mc{X}}\right)}{(K - \hat{K})!}.  
    \label{eq:def_PML_distribution_in_performance}
\end{align}
where the optimization is over distributions of possibly different support set 
size, $\mc{X}$ denotes the support set of distribution $p$, $K = |\mc{X}|$, and 
$\hat{K}$ denotes the support set size of the empirical distribution $\hat{p}$. 
$p^*$ is difficult to compute exactly, so we compute the APML distribution 
$\bar{p}^*$ (\ref{eq:def_approximate_unlabeled_ML}), 
(\ref{eq:def_unlabeled_ML_alphabet_size_unknown_approx_PML}):
\begin{align}
    \bar{p}^* &= \argmax_{p \in \mc{P}} \frac{e^{\bar{V}(p)}}{(K - \hat{K})!} 
    \label{eq:def_approximate_PML_distribution_in_performance}
\end{align}
which maximizes a lower bound to the probability $\mb{P}_{p}(\mc{F})$ since 
$\bar{V}(p)$ (\ref{eq:permanent_lower_bound}) lower bounds the log permanent in 
(\ref{eq:def_PML_distribution_in_performance}).  Then our estimator for 
function $F$ of the form (\ref{eq:F_functional}) is the plugin $F(\bar{p}^*)$.

Our approximate PML estimator performs comparably well to the competition 
\cite{Valiant--Valiant2011power} \cite{Valiant--Valiant2013estimating} 
\cite{Jiao--Venkat--Han--Weissman2015minimax}, \cite{Wu--Yang2014minimax},
\cite{Wu--Yang2015unseen} across different functions $F$ and distributions, and 
performs significantly better when the true distribution is uniform.  This good 
performance for the uniform case makes some intuitive sense: the approximate 
PML distribution $\bar{p}^*$ maximizes a lower bound 
(\ref{eq:permanent_lower_bound_preview}) to a matrix permanent obtained by 
summing over only those permutations that mix symbols within level sets of 
$\bar{p}^*$.  If there is only one level set -- that is, if $\bar{p}^*$ is 
uniform -- then we have exactly optimized the matrix permanent (or at least 
found a local maximum), rather than a lower bound to the permanent.

\subsection{Sorted $L_1$ loss}

Given $n$ samples from a distribution $p$, the usual $L_1$ loss in measuring the accuracy of estimating $p$ is given by
\begin{align}
\sum_{x\in \mathcal{X}}|p_x - q_x|.  \end{align}

The sorted $L_1$ loss measures the \emph{shape} difference between the 
reconstruction distribution $q$ and the true distribution $p$. In other words, 
we first sort the distributions $p, q$ into non-increasing probability vectors 
$(p_{(1)}, p_{(2)}, \ldots, p_{(K)}), (q_{(1)}, q_{(2)}, \ldots, q_{(K)})$, and 
then measure the loss by
\begin{align}
    \sum_{1\leq i\leq K} |p_{(i)} - q_{(i)}|.  \label{eq:def_sorted_L1_loss}
\end{align}

The sorted $L_1$ loss measures the error in reconstructing the sorted probability vector, for which the fingerprints are sufficient as shown by Theorem~\ref{theorem.groupaction}. The sorted $L_1$ loss can also be viewed as a Wasserstein distance~\cite{vallender1974calculation}, i.e., 
\begin{align}
\sum_{1\leq i\leq K} |p_{(i)} - q_{(i)}| & = K \sup_{f \in \mathsf{Lip}_1} \int 
    f(x) \left( \sum_{x} \frac{1}{K} \delta_{p_x}  - \sum_{x} \frac{1}{K} 
    \delta_{q_x} \right),
\end{align}
where the supremum is over all Lipschitz functions with Lipschitz constant one. 

We estimate the sorted $L_1$ loss using the plug-in $q = \bar{p}^*$, the APML 
distribution (\ref{eq:def_approximate_PML_distribution_in_performance}), 
computed over collection of distributions $\mc{P} = \Delta$, where
\begin{equation}
    \Delta \eqdef \{p : |\supp(p)| < \infty\} 
    \label{eq:def_distributions_with_finite_support_set_size}
\end{equation}
denotes the set of all discrete distributions with finite support set size, and 
collection of distributions $\mc{P} = \Delta_K$, where
\begin{equation}
    \Delta_K \eqdef \{p : |\supp(p)| = K\} 
    \label{eq:def_distributions_with_particular_support_set_size}
\end{equation}
denotes the set of all discrete distributions with support set size $K$.

\begin{figure}[H]
	\capstart
	\begin{center}
    \includegraphics[width=7in]{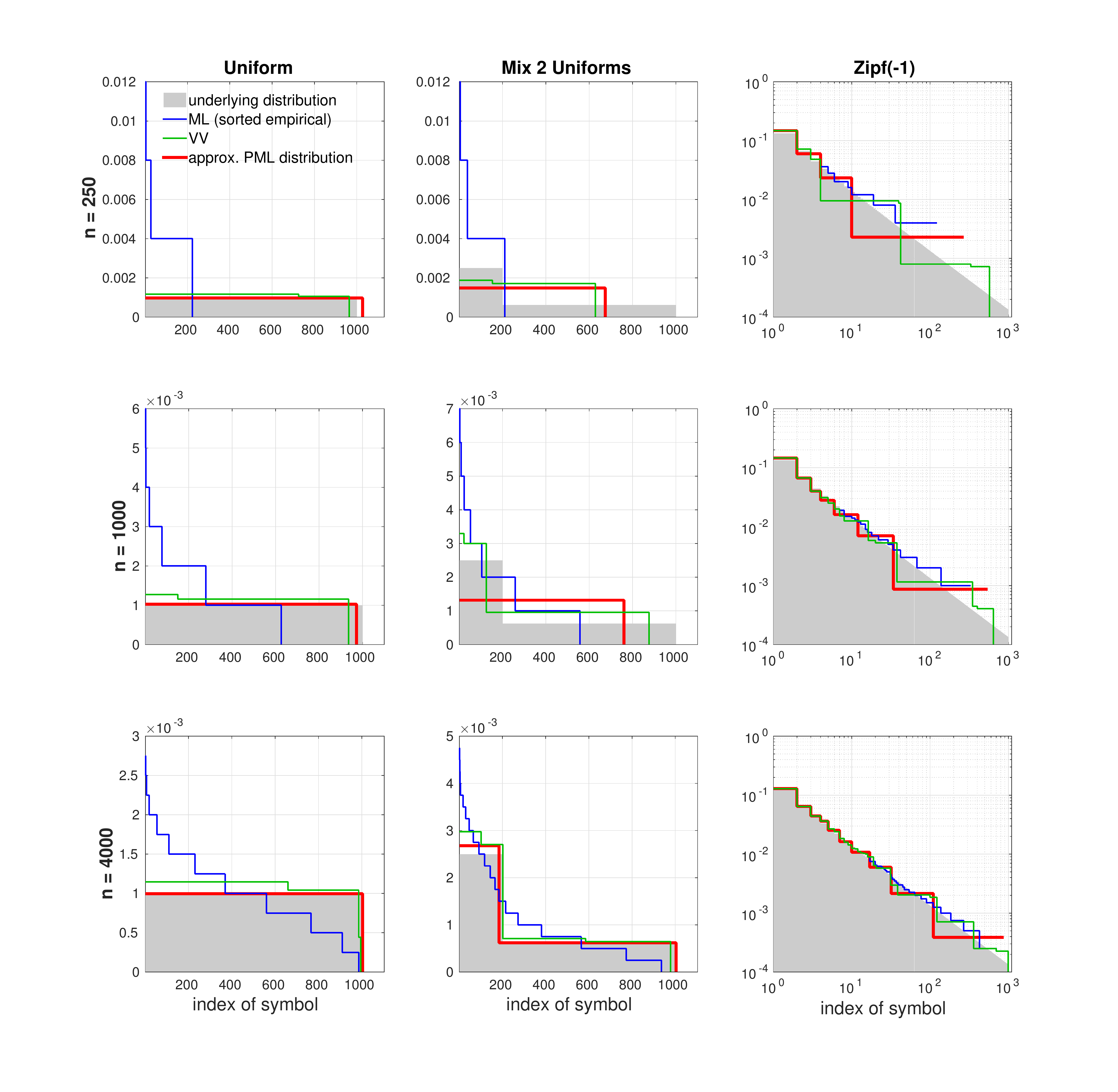}
        \caption{Inferring the sorted probability vector given samples from an 
        underlying distribution (plotted in gray).  The distribution is held 
        constant in each column and the sample size $n$ is held constant in 
        each row of the figure.  In all cases $K = |\mc{X}| = 1000$.  
        ``Uniform'' is uniform on $\mc{X}$, ``Mix 2 Uniforms'' is a mixture of 
        two uniform distributions, with half the probability mass on the first 
        $K/5$ symbols, and the other half on the remaining symbols, and 
        Zipf$(\alpha) \sim 1/i^\alpha$ with $i \in \{1,\ldots,K\}$.  ML denotes 
        the sorted empirical distribution.  VV is 
        \cite{Valiant--Valiant2013estimating}.}
    \label{fig:sorted_probability_vector}
	\end{center}
\end{figure}

Figure \ref{fig:sorted_probability_vector} shows the sorted probability vector 
inferred by our approximate PML distribution in $\Delta$ 
(\ref{eq:def_distributions_with_finite_support_set_size}) -- that is, the case 
of unknown support set size -- along with the sorted ML (empirical) 
distribution and the distribution of \cite{Valiant--Valiant2013estimating}.  We 
see that both the approximate PML distribution and the distribution of 
\cite{Valiant--Valiant2013estimating} perform much better than the sorted ML 
distribution, and that the PML distribution ``prefers'' to have fewer and 
larger level sets than the distribution of 
\cite{Valiant--Valiant2013estimating}.

\begin{figure}[H]
	\capstart
	\begin{center}
    \includegraphics[width=5in]{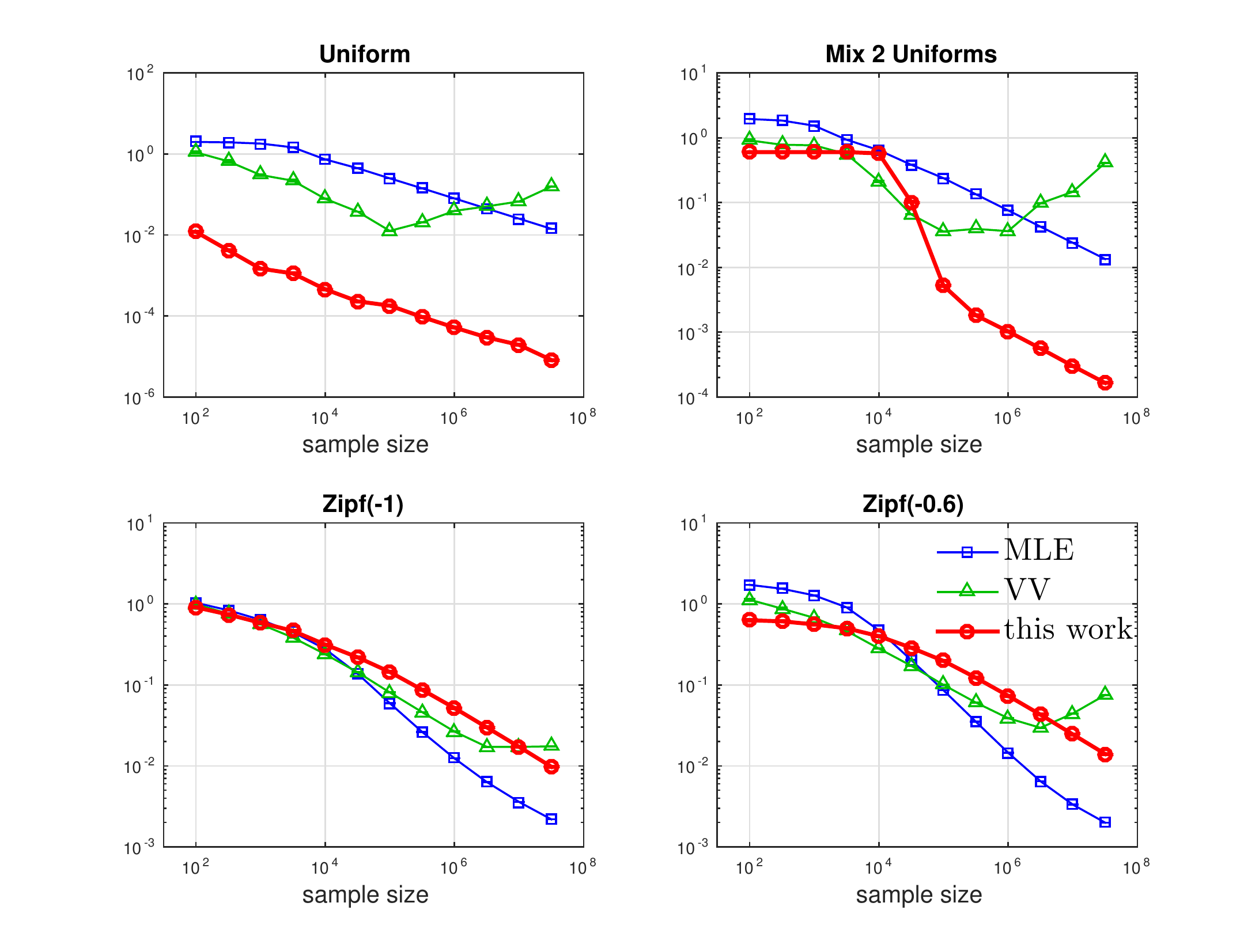}
        \caption{Sorted $L_1$ loss in estimating an unknown distribution $p$ 
        with known support set size $K = |\mc{X}|$.  In all cases $K = 10^4$.  
        ``Uniform'' is uniform on $\mc{X}$, ``Mix 2 Uniforms'' is a mixture of 
        two uniform distributions, with half the probability mass on the first 
        $K/5$ symbols, and the other half on the remaining symbols, and 
        Zipf$(\alpha) \sim 1/i^\alpha$ with $i \in \{1,\ldots,K\}$.  MLE 
        denotes the ML plugin (``naive'') approach of using the sorted 
        empirical distribution in (\ref{eq:def_sorted_L1_loss}).  VV is 
        \cite{Valiant--Valiant2013estimating}.  Each data point represents 100 
        random trials, with 2 standard error bars smaller than the plot marker 
        for most points.}
    \label{fig:sorted_L1_loss}
	\end{center}
\end{figure}

Figure \ref{fig:sorted_L1_loss} shows the performance of our approximate PML 
distribution in $\Delta_{K}$ 
(\ref{eq:def_distributions_with_particular_support_set_size}) -- that is, the 
case of known support set size -- as a plugin estimator for the sorted $L_1$ 
loss, along with the estimator of \cite{Valiant--Valiant2011power} and the 
empirical distribution plugin estimator (MLE) (see caption for performance test 
parameters).  We see that the approximate PML distribution performs well for 
the uniform and mixture-of-two-uniforms distributions, but more poorly for the 
Zipf distributions, consistent with the remarks in the previous Section.  Good 
performance on near-uniform distributions is observed for other performance 
tests as well (see below).

\subsection{Entropy and R\'enyi entropy estimation}

For entropy $H(p) = -\sum_{x \in \mc{X}} p_x \log(p_x)$ and R\'enyi entropy 
$H_\alpha(p) = \frac{1}{1-\alpha} \log\left(\sum_{x \in \mc{X}} 
p_x^\alpha\right)$ of distribution $p$, the corresponding approximate PML 
estimator is defined as the plugin estimator $H(\bar{p}^*)$, 
$H_\alpha(\bar{p}^*)$, where $\bar{p}^*$ is as in 
(\ref{eq:def_approximate_PML_distribution_in_performance}), optimized over the
collection of distributions $\mc{P} = \Delta$ 
(\ref{eq:def_distributions_with_finite_support_set_size}) for both the entropy 
and R\'enyi entropy.  Additionally, to enable direct comparison with the 
entropy estimator of \cite{Wu--Yang2014minimax}, which requires a support set 
size as input, we also find the approximate PML distribution optimized over 
$\mc{P} = \Delta_K$ 
(\ref{eq:def_distributions_with_particular_support_set_size}).



\begin{figure}[p]
	\capstart
    \includegraphics[width=7in]{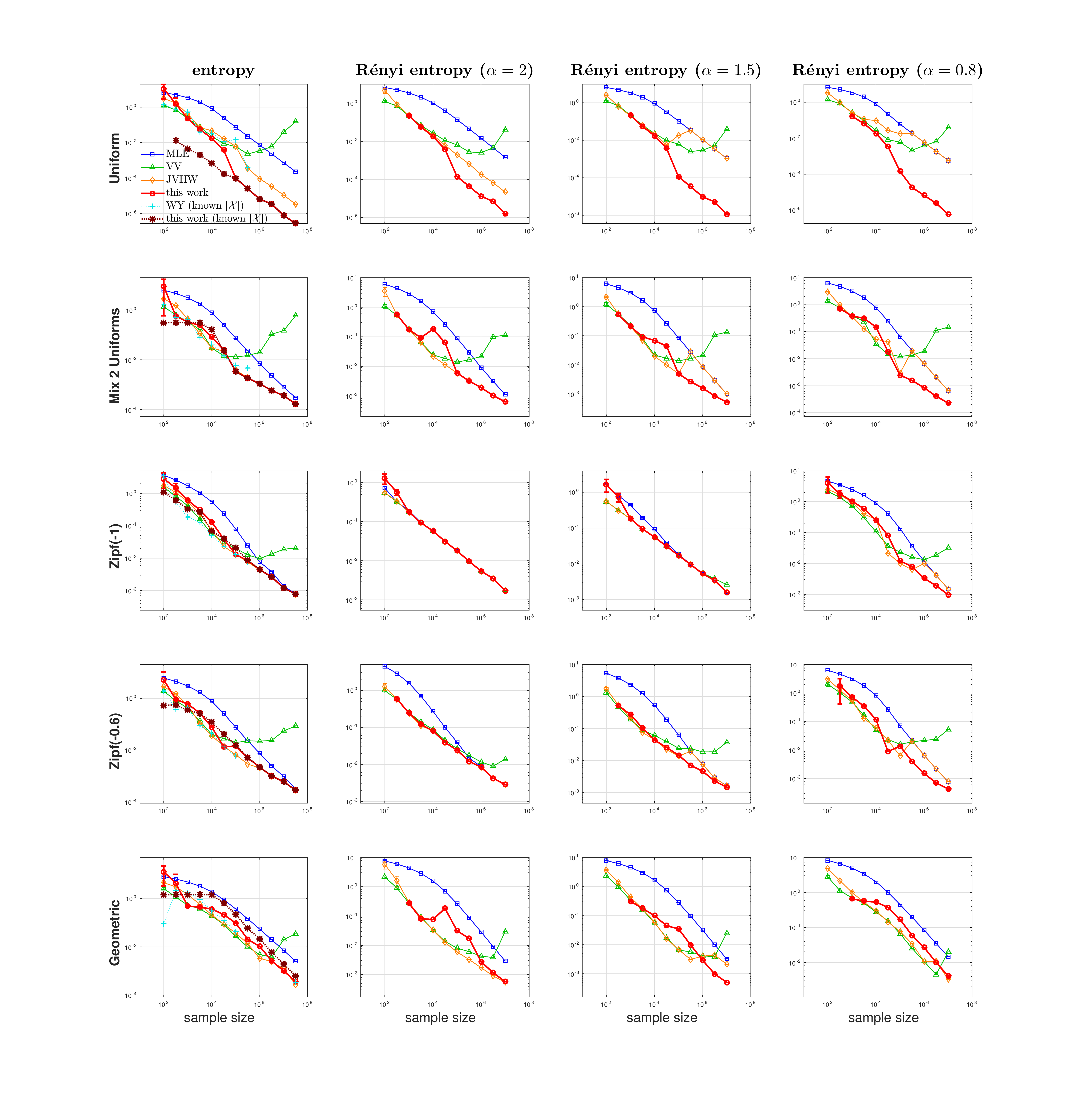}
        \caption{Root mean squared error of several estimators of entropy 
        (first column) and R\'enyi entropy (last three columns for $\alpha \in 
        \{2, 1.5, 0.8\}$).  In all cases except ``Geometric'' we set the 
        alphabet size $|\mc{X}| = 10^4$.  ``Uniform'' is uniform on $\mc{X}$, 
        Zipf$(\alpha) \sim 1/i^\alpha$ with $i \in \{1,\ldots,K\}$, 
        ``Geometric'' is the geometric distribution with infinite support and 
        mean $K$.  MLE denotes the ML plugin (``naive'') approach of computing 
        $H(\hat{p})$.  VV is \cite{Valiant--Valiant2013estimating}.  JVHW is 
        \cite{Jiao--Venkat--Han--Weissman2015minimax}.  WY is 
        \cite{Wu--Yang2014minimax} -- this estimator requires the support set 
        size $|\mc{X}|$ as an input.  Our estimator optionally accepts the 
        support set size $|\mc{X}|$ (shown in dark red).  Each data point 
        represents 100 random trials, with 2 standard error bars smaller than 
        the plot marker for most points.  Log base 2.}
	\label{fig:entro_Renyi_performance}
\end{figure}

Figure \ref{fig:entro_Renyi_performance} shows the performance of our 
approximate PML scheme for estimating the entropy and R\'enyi entropy with 
$\alpha \in \{2, 1.5, 0.8\}$.  Overall, our approximation looks competitive 
with the other approaches, and is particularly strong for the uniform 
distribution.  Adding in knowledge of the true support set size in estimating 
the entropy did not make much of a difference, except for improvement in the 
case of the uniform distribution.


\subsection{$L_1$ distance to uniformity}

For estimating the $L_1$ distance to uniformity, which is $\sum_{x\in 
\mathcal{X}} | p_x - \frac{1}{K}|$, the corresponding PML estimator 
is~\cite{acharya2016unified} the plugin of $p^*$ 
(\ref{eq:def_PML_distribution_in_performance}) optimized over the collection of 
distributions $\mc{P} = \Delta_K$ 
(\ref{eq:def_distributions_with_particular_support_set_size}) (the case of 
known support set size).  The APML estimator is the plugin of $\bar{p}^*$ 
(\ref{eq:def_approximate_PML_distribution_in_performance}) optimized over 
$\mc{P} = \Delta_K$.

\begin{figure}[H]
	\capstart
	\begin{center}
    \includegraphics[width=5in]{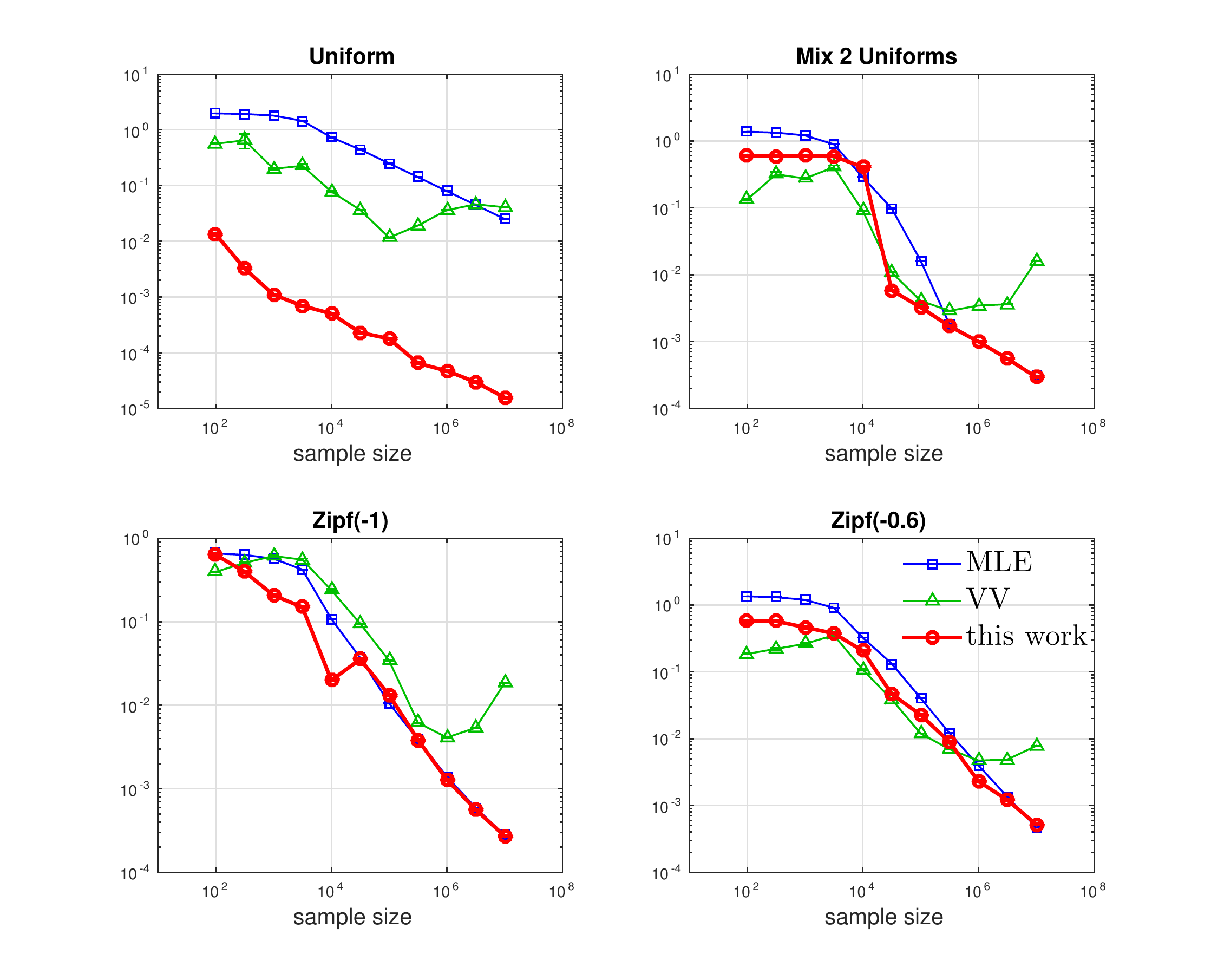}
        \caption{Root mean squared error for estimation of $L_1$ distance to 
        uniformity with known support set size $K = |\mc{X}|$.  In all cases $K 
        = 10^4$.  ``Uniform'' is uniform on $\mc{X}$, ``Mix 2 Uniforms'' is a 
        mixture of two uniform distributions, with half the probability mass on 
        the first $K/5$ symbols, and the other half on the remaining symbols, 
        and Zipf$(\alpha) \sim 1/i^\alpha$ with $i \in \{1,\ldots,K\}$.  MLE 
        denotes the ML plugin (``naive'') approach of computing $\sum_{x \in 
        \mc{X}}|\hat{p}_x - \frac{1}{K}|$.  VV is 
        \cite{Valiant--Valiant2013estimating}.  Each data point represents 100 
        random trials, with 2 standard error bars smaller than the plot marker 
        for most points.}
	\label{fig:L1_distance_to_uniformity}
	\end{center}
\end{figure}

Figure \ref{fig:L1_distance_to_uniformity} shows the performance of our 
approximate PML scheme for estimating the $L_1$ distance to a uniform 
distribution with known support set size.  Our approach looks competitive, and 
is by far the best if the true distribution is uniform.


\subsection{Support set size estimation}

For support size estimation, the corresponding PML estimator 
is~\cite{acharya2016unified} the plugin of $p^*$ 
(\ref{eq:def_PML_distribution_in_performance}) optimized over the collection of 
distributions $\mc{P} = \Delta_{\geq \frac{1}{K}}$, where
\begin{equation}
    \Delta_{\geq \frac{1}{K}} \eqdef \{p : p_x \geq \frac{1}{K}\ \forall x \in 
    \mc{X}\}
\end{equation}
denotes the set of all discrete distributions whose minimum nonzero probability 
for each symbol is at least $\frac{1}{K}$.  The approximate PML estimator is 
the plugin of $\bar{p}^*$ 
(\ref{eq:def_approximate_PML_distribution_in_performance}) optimized over 
$\mc{P} = \Delta_{\geq \frac{1}{K}}$.

\begin{figure}[H]
	\capstart
	\begin{center}
    \includegraphics[width=6in]{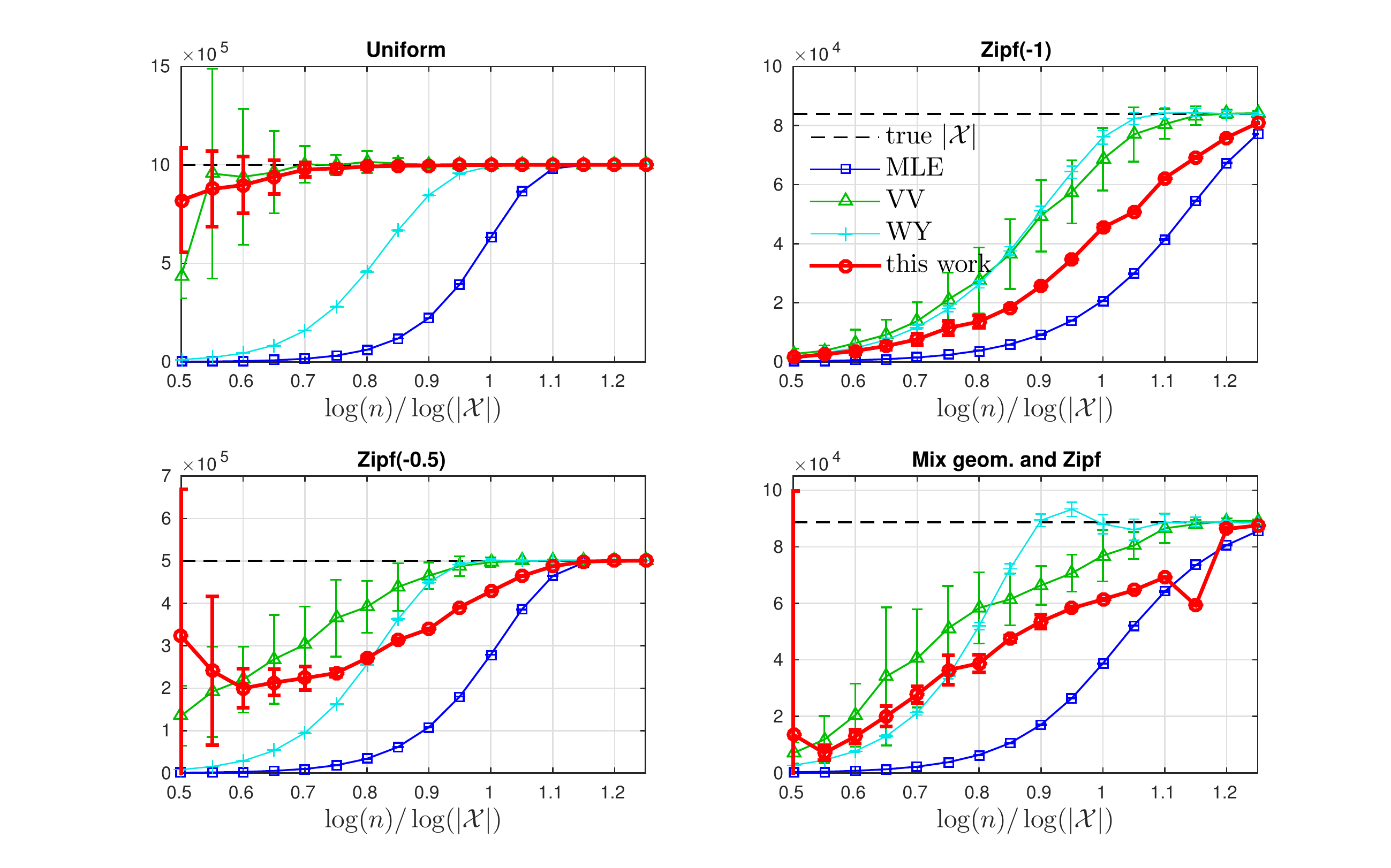}
        \caption{Support set size estimation for the same cases as 
        in~\cite{Wu--Yang2015unseen}, plotting the mean and 2 standard 
        deviations of the distribution of inferred support set sizes in 100 
        trials vs. sample size $n$.  In all cases the support set set size $K = 
        |\mc{X}|$ is chosen such that $\min_{i \in \{1,\ldots,K\}} p_i \approx 
        10^{-6}$.  ``Uniform'' is uniform on $\{1,\ldots,K\}$, Zipf$(\alpha) 
        \sim 1/i^\alpha$ with $i \in \{1,\ldots,K\}$, and ``Mix geom. and 
        Zipf'' is a mixture of $p_i \sim 1/i$ for $i \in \{1,\ldots,K/2\}$ and 
        $p_i \sim (1-2/K)^{i-K/2-1}$ for $i \in \{K/2+1,\ldots,K\}$ normalized 
        so that half the mass is on the first $K/2$ symbols.  MLE denotes the 
        ML plugin (``naive'') approach of using the empirical support set size.  
        VV is \cite{Valiant--Valiant2013estimating}.  WY 
        is~\cite{Wu--Yang2015unseen}.}
	\label{fig:support_set_size}
	\end{center}
\end{figure}

Figure \ref{fig:support_set_size} shows the performance of our approximate PML 
scheme for estimating the support set size.  Here we plot the mean and standard 
error of the support set size inferred by the different estimation schemes 
rather than a root mean squared error.  Overall, our approach is comparable to 
the others, performing worst on the Zipf distributions, and best on the uniform 
distribution.

\section{Estimating the divergence function between two distributions}\label{sec.divergencepmlplugin}

In the 2-D PML setting discussed in Section~\ref{sec.divergenceestimationpml}, 
we argued the sufficiency of the fingerprint~(\ref{eq:def_fingerprint2d}) in 
estimating functions of two discrete distributions of the form
\begin{align}
    F(p, q) = \sum_{x\in \mathcal{X}} f(p_x,q_x).  
    \label{eq:symmetric_functional_two_distributions}
\end{align}

In order to apply Theorem~\ref{theorem.ml}, one needs to show the cardinality of 
the sufficient statistics is not too big. Below we present an argument that 
applies to the $D$-dimensional PML.
Suppose we have $D$ discrete distributions on the same alphabet $\mathcal{X}$, 
denoted as $p^{(1)}, p^{(2)}, \ldots, p^{(D)}$. Suppose we obtain $n_d$ samples 
from each distribution $p^{(d)}$ with empirical distribution $\hat{p}^{(d)}$.  
Let the joint fingerprint be defined for every $(i_1,i_2,\ldots,i_D)\geq 0$,
\begin{align}
    \mathcal{F}_{i_1,i_2,\ldots,i_D} & = | \{ x\in \mathcal{X}: (n_d 
    \hat{p}^{(d)}_x)_{d=1}^D = (i_d)_{d=1}^D \} |.
\end{align}

We now argue that the joint fingerprint $(\mathcal{F}_{i_1,i_2,\ldots,i_D})_{(i_1,i_2,\ldots,i_D)\geq 0}$ corresponds to the partition of multipartite numbers. Indeed, the joint fingerprint satisfies the following equation:
\begin{align}
    \sum_{(i_1,i_2,\ldots,i_D)\geq 0} \mathcal{F}_{i_1,i_2,\ldots,i_D} 
    \begin{bmatrix} i_1 \\ i_2 \\ \vdots \\ i_D \end{bmatrix}  & = 
        \begin{bmatrix} n_1 \\ n_2 \\ \vdots \\ n_D \end{bmatrix}.
\end{align}

%
%
%

Theorem \ref{thm:cardinality_DD_fingerprint} quantifies the size of the 
sufficient statistic.  
\begin{theorem}\cite{auluck1953partitions,barany1992number, 
    acharya2011competitive} \label{thm:cardinality_DD_fingerprint}
Suppose $D = 2$. If $n_1 = n_2 = n$, then the cardinality of the set of 2-D 
    fingerprints on $n$ samples is given by
\begin{align}
    |\{(\mathcal{F}_{ij})_{(i,j)\neq (0,0)}\}| = e^{3 (\zeta(3))^{1/3} 
n^{2/3}(1+o(1))}.  \end{align} For general $D>2$, if $n_d \geq 2^{D+1},1\leq 
    d\leq D$, we have
\begin{align}
    |\{(\mathcal{F}_{i_1,i_2,\ldots,i_D})_{(i_1,i_2,\ldots,i_D)\neq \mbf{0}}\}| 
    & \leq \exp\left( 2 \left( 1 + \frac{1}{D} \right) \sum_{d =1}^D 
    n_d^{\frac{D}{D+1}} \right).
\end{align}
    where $\mbf{0} = (0,\ldots,0)$.  Moreover, if $n_d = n$ for all $d \in \{1, 
    \ldots, D\}$, we have
\begin{align}
    |\{(\mathcal{F}_{i_1,i_2,\ldots,i_D})_{(i_1,i_2,\ldots,i_D)\neq \mbf{0}}\}| 
    & = \exp\left( (D+1) (\zeta(D+1))^{1/(D+1)} n^{\frac{D}{D+1}} (1-o(1)) 
    \right).
\end{align}
Here $\zeta(D+1) = \sum_{k =1}^\infty k^{-(D+1)}$ is the Riemann zeta function. 
\end{theorem} 

Fortunately, the cardinality of the $D$-dimensional fingerprint is still 
sub-exponential in the sample size for all $D$.  Then, combining with 
Theorem~\ref{theorem.ml} and the fact that there exist estimators for the 
aforementioned divergence functions with near exponential measure concentration 
(which can be shown in a straightforward manner as 
in~\cite{acharya2016unified,Han--Jiao--Weissman2016minimaxdivergence,bu2016estimation,Valiant--Valiant2011power,Jiao--Han--Weissman2016l1distance}), 
we know that plugging in the 2-D PML achieves the optimal sample complexity in 
estimating the Kullback--Leibler divergence, $L_1$ distance, the squared 
Hellinger distance, and the $\chi^2$ divergence.  The proof of these results is 
to be reported elsewhere.

In this section, we extensively test the performance of divergence estimation 
via plugging in our approximate 2-D PML. We mainly compete with the released 
code of KL divergence estimator 
in~\cite{Han--Jiao--Weissman2016minimaxdivergence}.
The concrete divergence functional estimation algorithm is as follows. Suppose 
we observe $n$ samples from distribution $p$ with empirical distribution 
$\hat{p}$ and $m$ samples from distribution $q$ with empirical distribution 
$\hat{q}$.  Let the 2-D PML estimator be 
as~(\ref{eq:def_unlabeled_ML_alphabet_size_known_2D}):
\begin{align}
    (p^*, q^*) & = \argmax_{(p,q) \in \mathcal{P}} 
    \mathbb{P}_{p,q}(\mathcal{F}),
\end{align}
where $\mathcal{F}$ is the 2-D fingerprint (\ref{eq:def_fingerprint2d}) and 
$\mc{P}$ is a collection of pairs of distributions on the same alphabet 
$\mc{X}$.  The APML distributions maximize a lower bound to 
$\mb{P}_{p,q}(\mc{F})$:
\begin{equation}
    (\bar{p}^*, \bar{q}^*) = \argmax_{(p,q) \in \mathcal{P}} 
    \frac{e^{\bar{V}(p, q)}}{(K - \hat{K})!}
\end{equation}
where $\bar{V}$ is defined for the $D$-dimensional case in Appendix 
\ref{app:dd_PML}, $\mc{X} = \supp(p) \cup \supp(q)$, $K = |\mc{X}|$, and 
$\hat{K} = |\supp(\hat{p}) \cup \supp(q)|$.


As discussed in Section \ref{sec:dd_PML}, there is no natural ordering on the 
bins of the 2-D fingerprint, so we do not give a dynamic programming algorithm 
like the one in section 
\ref{sec:dynamic_programming_ML_unlabeled_approximation} to find the APML 
distributions.  Instead we use a greedy heuristic presented in Appendix 
\ref{app:dd_PML} to approximately maximize $\bar{V}(p, q)$ and approximate the 
already-approximate PML distributions:
\begin{equation}
    (\bar{\bar{p}}^*, \bar{\bar{q}}^*) \approx (\bar{p}^*, \bar{q}^*) 
    \label{eq:approx_approx_2D_PML_distributions_in_performance}
\end{equation}
Then our estimator for functional $F$ of the form 
(\ref{eq:symmetric_functional_two_distributions}) is $F(\bar{\bar{p}}^*, 
\bar{\bar{q}}^*)$.


Our approximate PML estimator performs overall best relative to the 
competition, which for the KL divergence consists of only 
\cite{Han--Jiao--Weissman2016minimaxdivergence} and the MLE plugin estimator, 
and for the $L_1$ distance consists of only the MLE plugin estimator.  
\cite{Valiant--Valiant2013estimating}, \cite{Valiant2008testing} generalize 
their approach to the 2-D fingerprint setting, but do not release code to 
repeat their experiments.

\subsection{KL divergence estimation}

For the KL divergence $D(p||q) = \sum_{x \in \mc{X}} p_x 
\log\left(\frac{p_x}{q_x}\right)$, the corresponding approximate PML estimator 
is the plugin $D(\bar{\bar{p}}^*||\bar{\bar{q}}^*)$, where $\bar{\bar{p}}^*, 
\bar{\bar{q}}^*$ are as in 
(\ref{eq:approx_approx_2D_PML_distributions_in_performance}), optimized over 
collection of distributions $\mc{P} = \Delta_\rho$, where
\begin{equation}
    \Delta_\rho = \{(p, q) : \sup_x (p_x / q_x) \leq \rho\}
\end{equation}
denotes the set of all pairs of discrete distributions with finite support and 
maximum ratio $\rho$.

\begin{figure}[p]
	\capstart
	\begin{center}
    \includegraphics[width=7.5in]{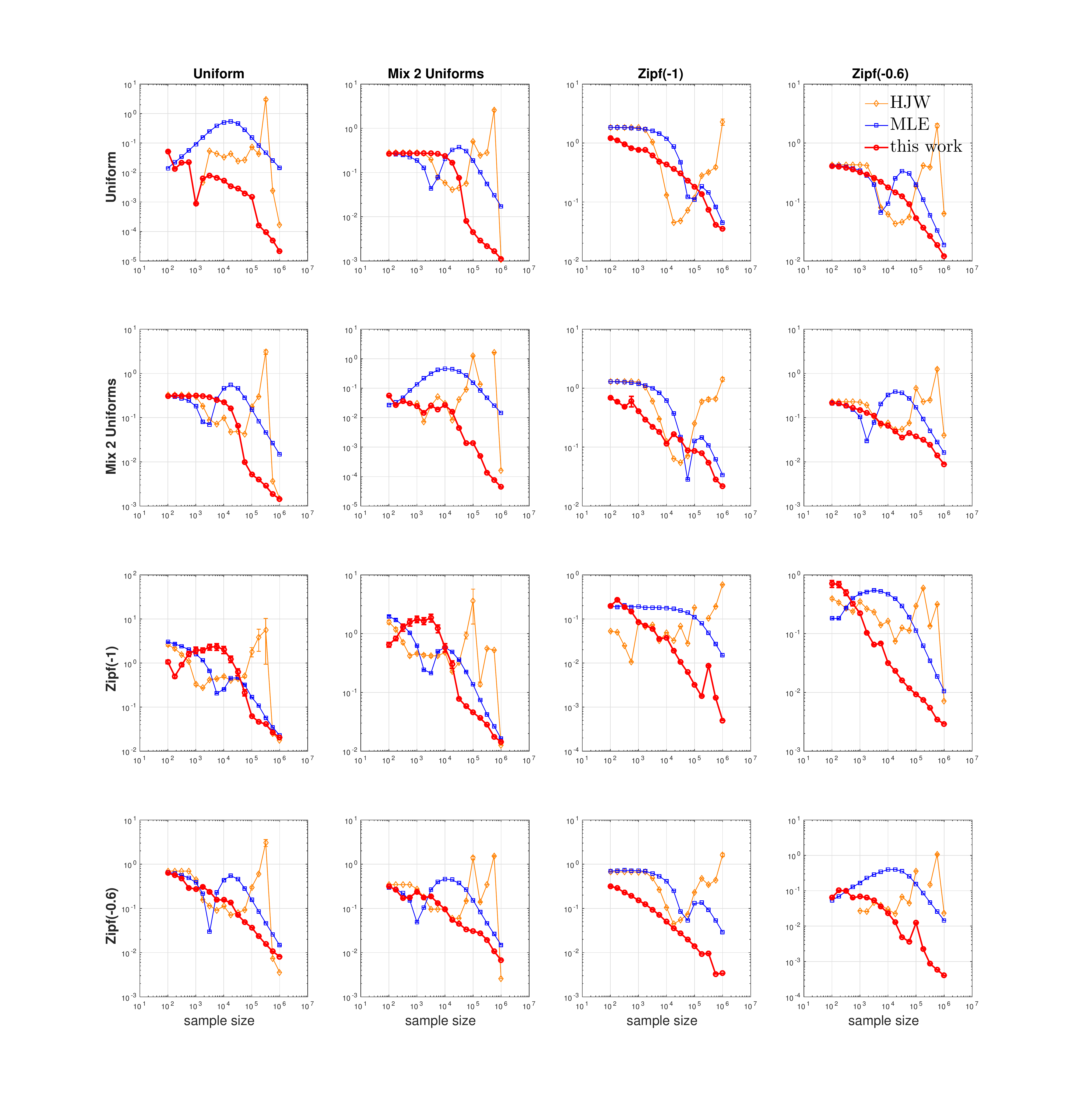}
        \caption{Root mean squared error for estimation of KL divergence 
        between distributions indicated in row and column names (that is, 
        $D(p_{\text{row}}||p_{\text{column}})$).  In all cases $K = |\mc{X}| = 
        10^4$.  ``Uniform'' is uniform on $\mc{X}$, ``Mix 2 Uniforms'' is a 
        mixture of two uniform distributions, with half the probability mass on 
        the first $K/5$ symbols, and the other half on the remaining symbols, 
        and Zipf$(\alpha) \sim 1/i^\alpha$ with $i \in \{1,\ldots,K\}$.  MLE 
        denotes the ML plugin (``naive'') approach of computing 
        $D(\hat{p}||\hat{q})$.  HJW is 
        \cite{Han--Jiao--Weissman2016minimaxdivergence}.  Each data point 
        represents 100 random trials, with 2 standard error bars smaller than 
        the plot marker for most points.  Log base 2.}
	\label{fig:KL_div_est_performance_fig}
	\end{center}
\end{figure}

Figure \ref{fig:KL_div_est_performance_fig} shows the performance of our 
approximate PML scheme for estimating the KL divergence.  The approximate PML 
estimator looks mostly better than 
\cite{Han--Jiao--Weissman2016minimaxdivergence} and the MLE.  All three 
estimators show non-monotonicity of their root mean squared performance in the 
sample size.  Qualitatively, the approximate PML estimator's performance looks 
best-behaved among the three in terms of being roughly monotone decaying in the 
sample size.


\subsection{$L_1$ distance}

For the $L_1$ distance $\|p-q\|_1 = \sum_{x \in \mc{X}} |p_x - q_x|$, the 
corresponding approximate PML estimator is the plugin $\|\bar{\bar{p}}^* - 
\bar{\bar{q}}^*\|_1$, where $\bar{\bar{p}}^*, \bar{\bar{q}}^*$ are as in 
(\ref{eq:approx_approx_2D_PML_distributions_in_performance}), optimized over 
collection of distributions $\mc{P} = \Delta$, where $\Delta$ denotes the set 
of all pairs of discrete distributions with finite support set size (possibly 
different support set sizes within the pair).

\begin{figure}[p]
	\capstart
	\begin{center}
    \includegraphics[width=7.5in]{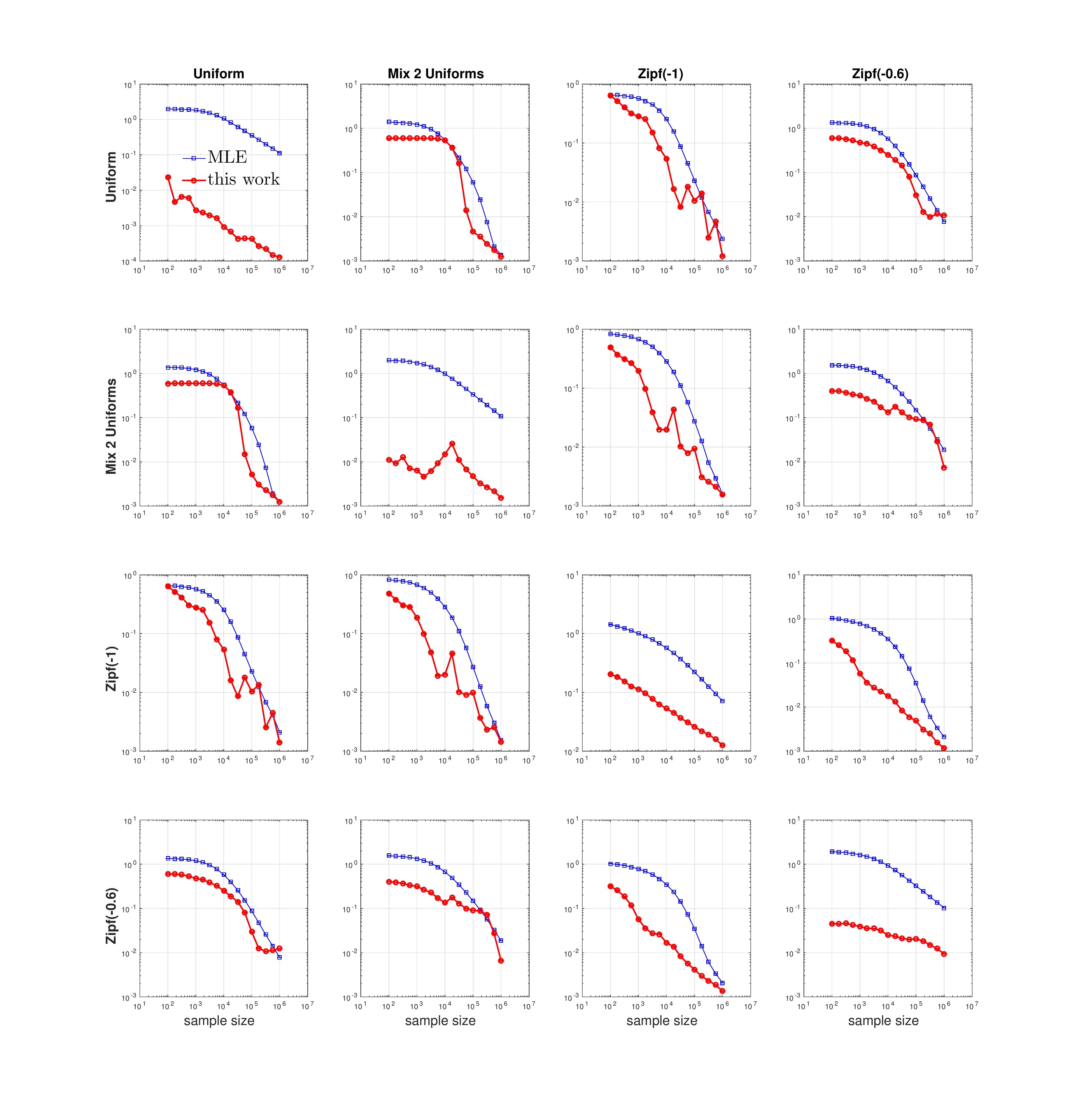}
        \caption{Root mean squared error for estimation of $L_1$ distance between distributions indicated in row and column names.  Same parameters as in Figure \ref{fig:KL_div_est_performance_fig}.  Since the $L_1$ distance is symmetric in its arguments, the performance matrix is symmetric.}
	\label{fig:L1_distance_est_performance_fig}
	\end{center}
\end{figure}

Figure \ref{fig:L1_distance_est_performance_fig} shows the performance of our 
approximate PML scheme for estimating the $L_1$ distance between two 
distributions.  The approximate PML estimator looks overall stronger than the 
MLE plugin estimator for all distributions and sample sizes, and seems to 
perform best when the two unknown distributions $p$ and $q$ are the same 
(corresponding to the diagonal of the matrix of plots in the figure).  Although 
\cite{Valiant--Valiant2013estimating} did not release their code for estimating 
the $L_1$ distance, their Figure 2, leftmost pane corresponds exactly in its 
parameters to the top-leftmost pane of our Figure 
\ref{fig:L1_distance_est_performance_fig}; from roughly checking a few sample 
sizes visually, the approximate PML seems to perform 
significantly better.


\section{Acknowledgement}

We are grateful to Huacheng Yu for providing the proof of Theorem~\ref{thm.huacheng}. We would like to thank Jayadev Acharya, Nick Loehr, Greta Panova, Richard Stanley, and Arthur Yang for helpful discussions on the size of the sufficient statistics. 

\appendices

\section{Proof of expression (\ref{eq:prob_unlabeled_empirical_distribution}) 
for the probability of a fingerprint} 
\label{app:prob_unlabeled_empirical_distribution}

We first state the probability of a specific fingerprint for a discrete, 
possibly infinite alphabet $\mc{X}$ and then specialize to the finite case.  
Let $\mc{S}_\mc{X}$ denote the symmetric group on $\mc{X}$ and let 
$\mc{S}_\mc{X} \hat{p}$ denote the orbit of empirical distribution $\hat{p}$ 
under the action of $\mc{S}_\mc{X}$ by relabeling the components of $\hat{p}$.  
That is, if $\sigma \in \mc{S}_\mc{X}$, then $(\sigma \hat{p})_x \eqdef 
\hat{p}_{\sigma(x)}$, and $\mc{S}_\mc{X}\hat{p} = \{\sigma \hat{p}\}_{\sigma 
\in \mc{S}_\mc{X}}$.  Each distribution in the orbit $\mc{S}_\mc{X} \hat{p}$ 
has the same fingerprint and distinct orbits have distinct fingerprints, so the 
set of distinct fingerprints $\mc{F}$ is in bijection with the set of distinct 
orbits $\mc{S}_\mc{X} \hat{p}$, so the probability of a specific fingerprint 
$\mc{F}$ is given by
\begin{align}
    \mb{P}_p(\mc{F}) = \mb{P}_p(\mc{S}_\mc{X} \hat{p}) & = \sum_{\hat{p}' \in 
    \mc{S}_\mc{X} \hat{p}} \mb{P}_p(\hat{p}') = \left(\begin{array}{c} n \\ n 
    \hat{p} \end{array}\right) \sum_{\hat{p}' \in \mc{S}_\mc{X} \hat{p}} 
    \prod_{x \in \mc{X}} p_x^{n \hat{p}_x'} 
    \label{eq:prob_unlabeled_empirical_distribution_discrete_alphabet_in_app}
\end{align}
where the last equality follows from (\ref{eq:prob_empirical_distribution}).

When the alphabet $\mc{X}$ is finite, then we can replace the summation over 
the orbit $\mc{S}_\mc{X} \hat{p}$ in 
(\ref{eq:prob_unlabeled_empirical_distribution_discrete_alphabet_in_app}) with 
a summation over the finite symmetric group $\mc{S}_\mc{X}$, taking care to 
avoid double-counting, since $|\mc{S}_\mc{X} \hat{p}| \leq |\mc{S}_\mc{X}|$ 
with strict inequality if there exist distinct $x,x'\in\mc{X}$ such that 
$\hat{p}_x = \hat{p}_{x'}$ (see Appendix~\ref{app.illustration} for an example 
of this).
The fingerprint notation lets us conveniently express $|\mc{S}_\mc{X} 
\hat{p}|$:
\beq
    |\mc{S}_\mc{X} \hat{p}| = \left(\begin{array}{c} |\mc{X}| \\ \mc{F} 
    \end{array}\right) = 
	|\mc{X}|! \prod_{i \geq 0} \frac{1}{\mc{F}_i!} .
	\label{eq:unlabeled_orbit_size_in_terms_of_fingerprint}
\eeq

The probability of a specific fingerprint $\mc{F}$ under distribution $p$ with 
finite support $\mc{X}$ is given by
\begin{align}
    \mb{P}_p(\mc{F}) = \mb{P}_p(\mc{S}_\mc{X} \hat{p}) & = \sum_{\hat{p}' \in 
    \mc{S}_\mc{X} \hat{p}} \mb{P}_p(\hat{p}') \\
    &\overset{\text{(a)}}{=} \frac{|\mc{S}_\mc{X} \hat{p}|}{|\mc{S}_\mc{X}|} 
    \sum_{\sigma \in \mc{S}_\mc{X}} \mb{P}_p(\sigma \hat{p}) \\
    &\overset{\text{(b)}}{=} \frac{1}{|\mc{X}|!} \left(\begin{array}{c} 
    |\mc{X}| \\ \mc{F} \end{array}\right) \sum_{\sigma \in \mc{S}_\mc{X}} 
    \mb{P}_p(\sigma \hat{p}) \\
    &\overset{\text{(c)}}{=} \left(\prod_{i \geq 0} \frac{1}{\mc{F}_i!}\right) 
    \left(\begin{array}{c} n \\ n \hat{p} \end{array}\right) \sum_{\sigma \in 
        \mc{S}_\mc{X}} \prod_{x \in \mc{X}} p_x^{n \hat{p}_{\sigma(x)}} \\
	&= \left(\prod_{i \geq 0} \frac{1}{\mc{F}_i!}\right) \left(\begin{array}{c} 
    n \\ n \hat{p} \end{array}\right) \perm\bigg(\underbrace{\left(\begin{array}{c} p_x^{n 
    \hat{p}_{x'}} \end{array}\right)_{x,x'\in\mc{X}}}_{Q}\bigg) 
    \label{eq:prob_unlabeled_empirical_distribution_in_app}
\end{align}
where in (a) the prefactor corrects for multiple-counting points in the orbit 
$\mc{S}_\mc{X} \hat{p}$ when we sum over $\sigma \in \mc{S}_\mc{X}$, (b) 
follows from (\ref{eq:unlabeled_orbit_size_in_terms_of_fingerprint}), and (c) 
follows from (\ref{eq:prob_empirical_distribution}).

\section{Illustration of computing the probability of a fingerprint}
\label{app.illustration}

To illustrate the computation of the probability of the fingerprint, suppose that we have an alphabet of $3$ symbols $\mc{X} = 
\{\text{a},\text{b},\text{c}\}$ and consider the sequence $x_1^7 = 
\left(\text{a,b,b,a,b,b,c}\right)$.  Then the empirical distribution is 
$\hat{p}(x_1^7) = \left(\hat{p}_\text{a},\hat{p}_\text{b}, 
\hat{p}_\text{c}\right) = \frac{1}{7}\left(2,4,1\right)$ and the fingerprint 
$\mc{F}(\hat{p})$ is $\mc{F}_i = 1$ if $i \in \{1, 2, 3\}$ and $\mc{F}_i = 0$ 
otherwise.  The probability under $p = (p_\text{a}, p_\text{b}, p_\text{c})$ of 
$\mc{F}(\hat{p})$ is indicated pictorially below:
\ytableausetup{smalltableaux}
\ytableausetup{centertableaux}
\begin{align}
    \mb{P}(\mc{F}(\hat{p})) = \mb{P}_p\left(\mc{S}_\mc{X} 
    \left(\begin{ytableau} \none & \ & \none \\ \none & \ & \none \\ \ & \ & 
    \none \\ \ & \ & \ \\ \none[\text{a}] & \none[\text{b}] & \none[\text{c}] 
    \end{ytableau}\right)\right) =&\ \ \   
	\mb{P}_p\left(\ydiagram{1,1,2,3}\right) + 
	\mb{P}_p\left(\ydiagram{1+1,1+1,2,3}\right) + 
	\mb{P}_p\left(\begin{ytableau} \none & \none & \\ \none & \none & \\ & 
	\none & \\ & & \end{ytableau}\right) \nonumber \\
	&+ \mb{P}_p\left(\begin{ytableau} \ & \none & \none \\ \ & \none & \none \\ 
	& \none & \\ & & \end{ytableau}\right) + 
	\mb{P}_p\left(\ydiagram{1+1,1+1,1+2,3}\right) + 
	\mb{P}_p\left(\ydiagram{2+1,2+1,1+2,3}\right) 
	\label{eq:pictorial_computation_prob_unlabeled_empirical_distribution_p1_1} 
	\\
        =& \frac{1}{1! 1! 1!} \ \frac{7!}{2! \ 4! \ 1!} \ \perm 
        \left(\begin{array}{ccc} p_\text{a}^2 & p_\text{a}^4 & p_\text{a}^1 \\ 
            p_\text{b}^2 & p_\text{b}^4 & p_\text{b}^1 \\ p_\text{c}^2 & 
            p_\text{c}^4 & p_\text{c}^1
	\end{array}\right) 
	\label{eq:pictorial_computation_prob_unlabeled_empirical_distribution_p1_2}
\end{align}
where each term on the right hand side of 
(\ref{eq:pictorial_computation_prob_unlabeled_empirical_distribution_p1_1}) is 
computed via (\ref{eq:prob_empirical_distribution}) and in 
    (\ref{eq:pictorial_computation_prob_unlabeled_empirical_distribution_p1_2}) 
    the two prefactors are the 
    multinomial coefficients in 
    (\ref{eq:prob_unlabeled_empirical_distribution_in_app}).

Suppose now that we happen to observe only the symbol `b' in $x_1^4 = 
\left(\text{b,b,b,b}\right)$, but know that $\mc{X}$ consists of $3$ symbols as 
    in the previous example.  Then $\hat{p}(x_1^4) = 
    \frac{1}{4}\left(0,1,0\right)$ and the fingerprint is $\mc{F}_0 = 2$, 
    $\mc{F}_1 = 1$, and $\mc{F}_i = 0$ for $i \geq 2$.
    Our pictorial computation of the probability under $p = (p_\text{a}, 
    p_\text{b}, p_\text{c})$ of $\mc{F}(\hat{p})$ is shown in 
    (\ref{eq:pictorial_computation_prob_unlabeled_empirical_distribution_p2_1}):
\begin{align}
    \mb{P}_p(\mc{F}(\hat{p})) = \mb{P}_p\left(\mc{S}_\mc{X} 
    \left(\begin{ytableau} \none & \ & \none \\ \none & \ & \none \\ \none & \ 
    & \none \\ \none & \ & \none \\ \none[\text{a}] & \none[\text{b}] & 
    \none[\text{c}] \end{ytableau}\right) \right) &= 
    \mb{P}_p\left(\begin{ytableau} \ & \none & \none \\ \ & \none & \none \\  \ 
    & \none & \none \\  \ & \none & \none  \end{ytableau} \right) +
    \mb{P}_p\left(\begin{ytableau} \none & \ & \none \\ \none & \ & \none \\  
	\none & \ & \none \\  \none & \ & \none  \end{ytableau} \right) +
	\mb{P}_p\left(\ydiagram{2+1,2+1,2+1,2+1}\right) 
	\label{eq:pictorial_computation_prob_unlabeled_empirical_distribution_p2_1} 
	\\
    &= \frac{1}{2! 1!} \ \frac{4!}{0! 4! 0!} \ \perm\left(\begin{array}{ccc} 1 
        & p_\text{a}^4 & 1 \\ 1 & p_\text{b}^4 & 1 \\ 1 & p_\text{c}^4 & 1
	\end{array}\right) = p_\text{a}^4 + p_\text{b}^4 + p_\text{c}^4
\end{align}
    Note that there are now fewer than $|\mc{S}_\mc{X}| = |\mc{X}|!$
distinct permutations to sum over, reflected in the prefactor $1/\mc{F}_0!$ 
with $\mc{F}_0 = 2$ since there are two ``unseen'' symbols.

\section{Proof of Theorem~\ref{thm.huacheng}}

Let $Z$ be the number in Theorem~\ref{thm.huacheng}. We first prove the lower bound. 

Fix a (constant) parameter $\alpha\in (0, 1)$.
Consider all \emph{0-1} assignments to an $\alpha n\times \alpha n$ array such that the total number of $1$'s is exactly $n$.
The number of such assignments is
\[
	\binom{\alpha^2 n^2}{n}\geq \left(\alpha^2 n\right)^n.
\]

On the other hand, each equivalent class generates at most
\[
	((\alpha n)!)^2\leq (\alpha n)^{2\alpha n}
\]
such assignments. Therefore, the number of equivalent classes $Z$ is at least
\[
	Z\geq \left(\alpha^2 n\right)^n/(\alpha n)^{2\alpha n}=n^{(1-2\alpha)n}\cdot \alpha^{2n(1-\alpha)}.
\]

The above inequality holds for any $\alpha>0$.
In particular, for any constant $\alpha$, we have
$Z\geq n^{(1-2\alpha-o(1))n}$, which implies the claimed bound.

Now we prove the upper bound. Consider the following procedure for generating an assignment to a $n\times n$ array.
\begin{enumerate}
	\item
		Choose an integer $1\leq m\leq n$ arbitrarily;
	\item
		Choose $m$ entries from the $n^2$ entries arbitrarily;
	\item
		For each of the $m$ chosen entries from last step, fill in a positive integer such that the $m$ entries sum up to $n$.
\end{enumerate}

It is easy to verify that for every equivalent class, some assignment in the class can be generated by the above procedure.
On the other hand, the number of different assignments that can be generated is at most
\[
	\sum_{m=1}^n \binom{n^2}{m}\cdot \binom{n}{m}\leq 2^n\cdot n\cdot \max_{1\leq m\leq n}\binom{n^2}{m}\leq 2^n\cdot n\cdot (e\cdot n)^n=n^{(1+o(1))n}.
\]
This proves the upper bound.

\section{The EM algorithm used to numerically solve the PML for small alphabets in Section~\ref{sec.approximatepml}} \label{app.emalgorithm}

\subsection{1D case}
We treat $p$ as the unknown parameter to be inferred and assume that $p$ is 
sorted.  Recall that for a fixed alphabet size, the PML distribution is:
\begin{equation}
    p^* = \argmax_p \sum_{\sigma \in \mc{S}_\mc{X}} \prod_{x \in \mc{X}} p_x^{n 
    \hat{p}_{\sigma(x)}}
\end{equation}

We treat permutation $\sigma$ as the hidden data drawn uniformly on 
$\mc{S}_\mc{X}$, and the observed data $\hat{p}$ drawn from the permuted 
distribution $\sigma p$.
Then if $p_{(t)}$ is our estimate for $p$ after $t$ iterations of the EM 
algorithm, the next value is:
\beq
    p_{(t+1)} \sim \sum_{\sigma \in \mc{S}_\mc{X}} (\sigma \hat{p}) \prod_{x 
    \in \mc{X}} p_{(t),x}^{n \hat{p}_{\sigma(x)}}
\eeq
where $\sigma \hat{p}$ denotes the permuted empirical histogram and $\sim$ 
denotes equality up to normalization.  For the starting point we choose 
$p_{(0)} = \hat{p}$.
%

\subsection{2-D case}
Analogously to the 1D case above, let $p_{(t)}$, $q_{(t)}$ be our estimates for 
$p$, $q$ after $t$ iterations of the EM algorithm.  Then the next values are:
\begin{align}
    p_{(t+1)} &\sim \sum_{\sigma \in \mc{S}_\mc{X}} (\sigma \hat{p}) w_\sigma 
    \\
    q_{(t+1)} &\sim \sum_{\sigma \in \mc{S}_\mc{X}} (\sigma \hat{q}) w_\sigma 
\end{align}
where
\beq
    w_\sigma \eqdef \prod_{x \in \mc{X}} p_{(t),x}^{n \hat{p}_{\sigma(x)}} 
    q_{(t),x}^{m \hat{q}_{\sigma(x)}}
\eeq
and $\sim$ denotes equality up to normalization.  For the starting point we 
choose $p_{(0)} = \hat{p}$, $q_{(0)} = \hat{q}$.


\section{Proof of Theorem \ref{thm:kd_PML_alphabet_size_2} ($D$-dimensional PML 
distribution on a binary alphabet)} 
\label{app:condition_nonuniformity_kd_PML_binary_alphabet}

The $D$-dimensional PML distributions $\mbf{p}^* \eqdef (p^{(d)*})_{d=1}^D$ are 
(see Appendix \ref{app:dd_PML} for a derivation):
\begin{align}
    \mbf{p}^* &\eqdef \argmax_{\mbf{p} \in \mc{P}} 
    \perm\left(\left(\prod_{d=1}^D (p^{(d)}_x)^{n_d 
    \hat{p}^{(d)}_{x'}}\right)_{x, x' \in \{1, 2\}}\right) \\
    &= \argmax_{\mbf{p} \in \mc{P}} \left(\prod_{d=1}^D (p_1^{(d)})^{n_d 
    \hat{p}_1^{(d)}} (1-p_1^{(d)})^{n_d (1-\hat{p}_1^{(d)})} + \prod_{d=1}^D 
    (p_1^{(d)})^{n_d (1-\hat{p}_1^{(d)})} (1-p_1^{(d)})^{n_d 
    \hat{p}_1^{(d)}}\right)
\end{align}
where $\mc{P}$ is a collection of $D$-tuples of distributions on the binary 
alphabet $\mc{X} = \{1, 2\}$.
To lighten notation, let
\begin{equation}
    r_d \eqdef p_1^{(d)}
\end{equation}
and
\begin{equation}
    \Delta_d \eqdef n_d \left(\hat{p}_1^{(d)} - \frac{1}{2}\right)
\end{equation}
Then $\mbf{p}^* = ((r_d^*, 1-r_d^*))_{d=1}^D$, where
\begin{align}
    (r_d^*)_{d=1}^D &= \argmax_{(r_d)_{d=1}^D \in [0,1]^D} 
    \Bigg(\underbrace{\prod_{d=1}^D r_d^{n_d/2 + \Delta_d} (1-r_d)^{n_d/2 - 
    \Delta_d} + \prod_{d=1}^D r_d^{n_d/2 - \Delta_d} (1-r_d)^{n_d/2 + 
    \Delta_d}}_{-U(r_1,\ldots,r_d)}\Bigg) \\
    &= \argmin_{(r_d)_{d=1}^D \in [0,1]^D} U(r_1, \ldots, r_d)
\end{align}
If $r_d^* = 1/2$ for all $d$, then $U(r_1,\ldots,r_d)$ must have positive 
definite Hessian $H_U$ at $(r_d)_{d=1}^D = (1/2)_{d=1}^D$.  Let's compute the 
Hessian at $r_d = 1/2$ for all $d$, letting $[D] \eqdef \{1, \ldots, D\}$:
\begin{align}
    H_U^* \eqdef \left.H_U\right|_{(r_d)_{d=1}^D = (1/2)_{d=1}^D} &= 
    \left.\left(\frac{\partial^2 U}{\partial r_d \partial r_{d'}}\right)_{d, d' 
    \in [D]}\right|_{(r_d)_{d=1}^D = (1/2)_{d=1}^D} \\
    &= c \left(\left(\Delta_d \Delta_{d'}\right)_{d, d' \in [D]} - 
    \text{diag}\left(\left(\frac{n_d}{4}\right)_{d\in[D]}\right)\right)
    \label{eq:hessian}
\end{align}
where $c = 2^{5 - \sum_{d=1}^D n_d}$ and in the second term in 
(\ref{eq:hessian}) $\text{diag}(\cdot)$ denotes the diagonal matrix with entry 
$n_d$ in the $d$-th row and $d$-th column.

The matrix $H_U^*$ is positive definite when $\Delta_d = 0$ for all $d$ with 
eigenvalues $(-c n_d/4)_{d=1}^D$.  Since the eigenvalues of $H_U^*$ vary 
continuously in $(\Delta_d)_{d=1}^D$, we conclude that $H_U^*$ is positive 
definite when $(\Delta_d)_{d=1}^D$ is in the neighborhood of $(0)_{d=1}^D$ 
determined by:
\begin{align}
    \det(H) = c'\left(-1 + \sum_{d=1}^D \frac{4 \Delta_d^2}{n_d}\right) &> 0 \\
    &\Updownarrow \nonumber \\
    \sum_{d=1}^D 4 n_d \left(\hat{p}_1^{(d)} - \frac{1}{2}\right)^2 &> 1 
    \label{eq:uniformity_D_distributions_alphabet_size_2_in_app}
\end{align}
where $c' = c \  2^{2D} \prod_{d=1}^D n_d$. Thus $H_U^*$ is not positive 
definite at $(r_d)_{d=1}^D = (1/2)_{d=1}^D$ when the empirical distribution 
components $(\hat{p}_1^{(d)})_{d=1}^D$ lie outside the ellipse determined by 
(\ref{eq:uniformity_D_distributions_alphabet_size_2_in_app}), so the $D$ PML 
distributions $(p^{(d)*})_{d=1}^D$ are not uniform in this case.  \qed

%

\section{Proof of permanent lower bound identity 
(\ref{eq:permanent_lower_bound_reexpressed})} 
\label{app:permanent_lower_bound_reexpressed}

Given distribution $p$ supported on set $\mc{X}$, denote by $\mc{X}_u(p)$ a 
level set of $p$:
\beq
    \mc{X}_u(p) \eqdef \{x \in \mc{X}: p_x = u\} 
    \label{eq:def_equivalence_class_by_p}
\eeq
Denote by $\mc{U}(p)$ the set of unique entries of $p$:
\beq
    \mc{U}(p) \eqdef \{p_x : x \in \mc{X}\} \label{eq:def_unique_entries_of_p}
\eeq
and denote by $\mc{A}(p)$ the partition of $\mc{X}$ into level sets of $p$ 
(equivalent to the definition (\ref{eq:def_partition_into_level_sets_of_p})):
\begin{equation}
    \mc{A}(p) \eqdef \{\mc{X}_u(p): u \in \mc{U}(p)\} 
    \label{eq:def_partition_into_level_sets_of_p_in_proof}
\end{equation}
It is convenient to express the group $\mc{S}_{\mc{X},p}$ 
(\ref{eq:def_subgroup_of_symmetric_group}) as isomorphic to:
\begin{equation}
   \mc{S}_{\mc{X},p} \cong \bigtimes_{\alpha \in \mc{A}(p)} \mc{S}_{\alpha} = 
    \bigtimes_{u \in \mc{U}(p)} \mc{S}_{\mc{X}_u}
    \label{eq:subgroup_of_symmetric_group_expression_in_proof}
\end{equation}
We write
\begin{align}
    V(p) = \log(\perm(Q)) &\geq \bar{V}(p) \\
    &= \log\left(\sum_{\sigma \in \mc{S}_{\mc{X},p}} \prod_{x \in \mc{X}} 
    p_x^{n \hat{p}_{\sigma(x)}}\right) \\
    &\overset{\text{(a)}}{=} \log\left(\sum_{\sigma \in \mc{S}_{\mc{X},p}} 
    \prod_{x \in \mc{X}} p_{\sigma(x)}^{n \hat{p}_x}\right) \\
    &\overset{\text{(b)}}{=} \log\left(\sum_{\sigma \in \bigtimes_{u \in 
    \mc{U}(p)} \mc{S}_{\mc{X}_u}} \ \prod_{x \in \mc{X}} p_{\sigma(x)}^{n 
    \hat{p}_x}\right) \\
    &\overset{\text{(c)}}{=} \log\left(\prod_{u \in \mc{U}(p)} \sum_{\sigma \in 
    \mc{S}_{\mc{X}_u}} \prod_{x \in \mc{X}_u} u^{n \hat{p}_x}\right) \\
    &\overset{\text{(d)}}{=} \log\left(\prod_{u \in \mc{U}(p)} 
    |\mc{S}_{\mc{X}_u}|\  u^{n \sum_{x \in \mc{X}_u}\hat{p}_x}\right) \\
    &= \sum_{u \in \mc{U}(p)} \Big(\log(|\mc{S}_{\mc{X}_u}|) + n \Big(\sum_{x 
    \in \mc{X}_u} \hat{p}_x\Big) \log(u)\Big) 
    \label{eq:permanent_lower_bound_intermediate_expression} \\
    &= \sum_{u \in \mc{U}(p)} \log(|\mc{S}_{\mc{X}_u}|!) + n \sum_{u \in 
    \mc{U}(p)} \sum_{x \in \mc{X}_u} \hat{p}_x \log\left(u\right) \\
    &\overset{\text{(e)}}{=} \sum_{u \in \mc{U}(p)} \log(|\mc{X}_u|!) + n 
    \sum_{x \in \mc{X}} \hat{p}_x \log\left(p_x\right) \\
    &= \sum_{u \in \mc{U}(p)} \log(|\mc{X}_u|!) + n \sum_{x \in \mc{X}} 
    \hat{p}_x \left(\log\left(\frac{p_x}{\hat{p}_x}\right) + 
    \log\left(\hat{p}_x\right)\right) \\
    &\overset{\text{(f)}}{=} \sum_{\alpha \in \mc{A}(p)} \log(|\alpha|!) - n 
    (D(\hat{p}||p) + H(\hat{p}))
\end{align}
where in (a) we changed order of summation over the group $\mc{S}_\mc{X}$, in 
(b) we summed over the isomorphic group 
(\ref{eq:subgroup_of_symmetric_group_expression_in_proof}), in (c) we used the 
product structure (\ref{eq:subgroup_of_symmetric_group_expression_in_proof}) of 
$\mc{S}_{\mc{X},p}$ to rewrite the sum as a product of sums and used the fact 
that $p_x = u$ for all $x \in \mc{X}_u$ (\ref{eq:def_equivalence_class_by_p}).  
In (d) we used the fact that the summands are independent of $\sigma$ and moved 
the sum into the exponent.  In (e) we used $|\mc{S}_{\alpha}| = |\alpha|!$ and 
used the fact that $u = p_x$ for all $x \in \mc{X}_u$.  In (f) we used $\alpha 
\in \mc{A}(p) \Leftrightarrow \exists u \in \mc{U}(p): \mc{X}_u = \alpha$.

\section{Proof of Theorem \ref{thm:partition_properties} (properties of the 
approximate PML distribution's level set partition $\bar{\mc{A}}^*$)} 
\label{app:partition_properties_proof}

\subsection{Proof of iso-clumping property 
(\ref{eq:same_empirical_prob_symbols_clumped_together_property})} 
\label{app:same_empirical_prob_symbols_clumped_together_property_proof}

Suppose $\bar{\mc{A}}^*$ does not satisfy property 
(\ref{eq:same_empirical_prob_symbols_clumped_together_property}).  Then there 
exist $x, x' \in \mc{X}$ such that $\hat{p}_x = \hat{p}_{x'}$, but $\alpha(x) 
\neq \alpha(x')$.  We show that the partition obtained by either 
``reassigning'' $x'$ to the same level set as $x$ or reassigning $x$ to the 
same level set as $x'$ achieves a higher value of the log permanent lower bound 
$\bar{V}(\cdot)$ (\ref{eq:permanent_lower_bound_reexpressed2}), contradicting 
the optimality of $\bar{\mc{A}}^*$ (\ref{eq:def_optimal_partition_approx_PML}).

Let $m \eqdef n \hat{p}_x = n \hat{p}_{x'}$, let $N_{\alpha(x)} \eqdef \sum_{x 
\in \alpha(x)} n \hat{p}_x$.  Let $V_\delta$ denote for $\delta \in [-1,1]$:
\begin{align}
    V_\delta \eqdef& \log((|\alpha(x)| + \delta)!) + \left(N_{\alpha(x)} + m 
    \delta\right) \log\left(\frac{N_{\alpha(x)} + m \delta}{n(|\alpha(x)| + 
    \delta)}\right) \nonumber \\
    & + \log((|\alpha(x')| - \delta)!) + \left(N_{\alpha(x')} - m \delta\right) 
    \log\left(\frac{N_{\alpha(x')} - m \delta}{n(|\alpha(x')| - \delta)}\right)
\end{align}

Then $V_0 = \bar{V}(\alpha(x)) + \bar{V}(\alpha(x'))$ 
(\ref{eq:partition_element_value}) corresponds to two terms in 
$\bar{V}(\bar{\mc{A}}^*)$ (\ref{eq:permanent_lower_bound_reexpressed2}), and 
$V_1$ (resp. $V_{-1}$) corresponds to two terms in the value of the partition 
of $\mc{X}$ obtained by reassigning symbol $x'$ to the same level set as $x$ 
(resp.  reassigning symbol $x$ to the same level set as $x'$).  We can check 
that $V_\delta$ is strictly convex in $\delta \in [-1,1]$.  A strictly convex 
function attains its maximum on the boundaries of a convex set, so either 
$V_{-1} > V_0$ or $V_1 > V_0$, contradicting the optimality of partition 
$\bar{\mc{A}}^*$, thus establishing the iso-clumping property.  \qed

\subsection{Proof of convexity property (\ref{eq:convexity_property})}

Suppose $\bar{\mc{A}}^*$ does not satisfy property 
(\ref{eq:convexity_property}).  Then there exist $x_1, x_2, x_3 \in \mc{X}$ 
such that $\hat{p}_{x_1} < \hat{p}_{x_2} < \hat{p}_{x_3}$, but $\alpha(x_1) = 
\alpha(x_3) \neq \alpha(x_2)$.  For all $x \in \mc{X}$, let $N_{\alpha(x)} 
\eqdef \sum_{x \in \alpha(x)} n \hat{p}_x$.  If 
$\frac{N_{\alpha(x_1)}}{|\alpha(x_1)|} \geq 
\frac{N_{\alpha(x_2)}}{|\alpha(x_2)|}$, then let $x = x_2$ and $x' = x_1$; 
otherwise let $x = x_3$, $x' = x_2$.  Let $\Delta \eqdef n (\hat{p}_x - 
\hat{p}_{x'})$.  Then $\Delta > 0$ and \begin{equation}
    \frac{N_{\alpha(x')}}{|\alpha(x')|} \geq \frac{N_{\alpha(x)}}{|\alpha(x)|}.  
    \label{eq:average_comparison_in_proof}
\end{equation}

Let $\mc{A}'$ denote the partition of $\mc{X}$ obtained by swapping the set 
memberships of $x$ and $x'$ in $\bar{\mc{A}}^*$.  We show that 
$\bar{V}(\mc{A}') > \bar{V}(\bar{\mc{A}}^*)$, contradicting the optimality of 
$\bar{\mc{A}}^*$ ($\mc{A}'$ might not satisfy the iso-clumping property 
(\ref{eq:same_empirical_prob_symbols_clumped_together_property})).

Since $\mc{A}'$ is obtained from $\bar{\mc{A}}^*$ by swapping the level set 
memberships of $x$ and $x'$, the elements of $\mc{A}'$ and $\bar{\mc{A}}^*$ 
have the same sizes, so $\sum_{\alpha \in \mc{A}'} \log(|\alpha|!) = 
\sum_{\alpha \in \bar{\mc{A}}^*} \log(|\alpha|!)$, so using 
(\ref{eq:permanent_lower_bound_reexpressed2}) and 
(\ref{eq:partition_element_value}) we write:
\begin{align}
    \bar{V}(\mc{A}') - \bar{V}(\bar{\mc{A}}^*) =& (N_{\alpha(x)} - \Delta) 
    \log\left(\frac{N_{\alpha(x)} - \Delta}{n |\alpha(x)|}\right) + 
    (N_{\alpha(x')} + \Delta) \log\left(\frac{N_{\alpha(x')} + \Delta}{n 
    |\alpha(x)|}\right) \nonumber \\
    &- N_{\alpha(x)} \log\left(\frac{N_{\alpha_x}}{n |\alpha(x)|}\right) - 
    N_{\alpha(x')} \log\left(\frac{N_{\alpha(x')}}{n |\alpha(x')|}\right) 
    \label{eq:diff_V_in_proof} \\
    =& \Delta 
    \log\left(\frac{N_{\alpha(x')}}{|\alpha(x')|}\frac{|\alpha(x)|}{N_{\alpha(x)}}\right) 
    + f(N_{\alpha(x)}, \Delta) + f(N_{\alpha(x')}, -\Delta) 
    \label{eq:diff_V_in_proof_2} \\
    \overset{\text{(a)}}{>}& 0
\end{align}
where $f(N,\Delta) \eqdef \Delta + (N - \Delta) \log\left(\frac{N - 
\Delta}{N}\right)$ with $f(N,N) \eqdef \lim_{\Delta \ra N} f(N, \Delta) = N$.
(a) follows because we can show that $f(N,\Delta) \geq 0$ for $\Delta \leq N$ 
with strict inequality iff $\Delta \neq 0$, so the last two terms in 
(\ref{eq:diff_V_in_proof_2}) are positive since $\Delta > 0$.  The first term 
in (\ref{eq:diff_V_in_proof_2}) is non-negative since $\Delta > 0$ and 
(\ref{eq:average_comparison_in_proof}).  Thus $\bar{V}(\mc{A}') > 
\bar{V}(\bar{\mc{A}}^*)$, contradicting the optimality of partition 
$\bar{\mc{A}}^*$, thus establishing the convexity property.  \qed

\section{Our approximate PML algorithm: the case of multiple 
distributions}\label{app:dd_PML}

%
%

We generalize the discussion of Section~\ref{sec.divergenceestimationpml} to 
$D$-dimensional fingerprints.  In this case we are not able to optimize a 
$D$-dimensional analogue of our objective function $\bar{V}$ 
(\ref{eq:permanent_lower_bound}) and will instead settle for an approximation 
of our already-approximate PML distribution.  We use a greedy heuristic to 
approximate the optimal partition $\bar{\mc{A}}^*$ 
(\ref{eq:def_optimal_partition_approx_PML}) by iteratively, greedily merging 
partition elements (enlarging the second term in a $D$-dimensional analogue of
(\ref{eq:permanent_lower_bound_reexpressed}) until the first term becomes too 
large) (see Section \ref{sec:dd_PML_approximation}).
Even this doubly-approximate algorithm turns out to be too slow to run on 
examples of practical interest, so we introduce a further heuristic to speed 
things up (only trying to merge partition elements that are close in Euclidean 
distance), which comes at a cost of adding some tunable knobs to the algorithm.

%


\subsection{Some notation} 

Suppose we have $D$ distributions $(p^{(d)})_{d=1}^D$ on the same alphabet 
$\mc{X}$ and for each $d$ draw $n_d$ samples i.i.d. from $p^{(d)}$ with 
empirical distribution $\hat{p}^{(d)}$.  Let bold symbols denote $D$-component 
collections:
\begin{equation}
    \begin{array}{cccc}\mathbf{p} \eqdef (p^{(d)})_{d=1}^D & \mathbf{n} \eqdef (n_d)_{d=1}^D & \hat{\mathbf{p}} \eqdef (\hat{p}^{(d)})_{d=1}^D & \mbf{n} \hat{\mbf{p}} \eqdef (n_d p^{(d)})_{d=1}^D\end{array}
\end{equation}

We assume that the $D$ distributions $\mbf{p}$ satisfy:
\begin{equation}
    \sum_{d=1}^D p^{(d)}_x > 0 \ \forall x \in \mc{X}
\end{equation}
This condition ensures that there are no symbols $x$ such that $p^{(d)}_x = 0$ 
for all $d$, which simplifies the expressions below.  Denote by $\hat{\mc{X}}$ 
the empirical support set:
\begin{equation}
    \hat{\mc{X}} \eqdef \bigcup_{d=1}^D \supp(\hat{p}^{(d)}) = \{x \in \mc{X}: 
    \sum_{d=1}^D \hat{p}^{(d)}_x > 0\}
\end{equation}
Finally let $K \eqdef |\mc{X}|$ and $\hat{K} \eqdef |\hat{\mc{X}}|$.


The probability to draw empirical distributions $\hat{\mbf{p}}$ under 
distributions $\mbf{p}$ is (using the independence of the $D$ empirical 
distributions and (\ref{eq:prob_empirical_distribution})):
\begin{equation}
    \mb{P}_{\mbf{p}}(\hat{\mbf{p}}) = \left(\prod_{d=1}^D 
    \left(\begin{array}{c} n_d \\ n_d \hat{p}^{(d)}\end{array}\right)\right) 
        \prod_{x \in \mc{X}} \prod_{d=1}^D (p_x^{(d)})^{n_d \hat{p}^{(d)}_{x}}
\end{equation}
where the first product is over multinomial coefficients in the notation of 
(\ref{eq:prob_empirical_distribution}).

Hereon denote by $\mc{F} = \mc{F}(\hat{p}^{(1)},\ldots,\hat{p}^{(D)})$ the 
$D$-dimensional fingerprint \cite{Raghunathan--Valiant--Zou2017estimating} 
indexed by vector $\mathbf{i} = (i_d)_{d=1}^D \in \mb{N}^D$ (which we write as 
satisfying $\mathbf{i} \geq \mbf{0} \Leftrightarrow i_d \geq 0 \ \forall d$, 
where $\mbf{0} \eqdef (0)_{d=1}^D$):
\begin{equation}
    \mc{F} = \mc{F}(\hat{\mathbf{p}}) = (\mc{F}_{\mathbf{i}})_{\mathbf{i} \geq 
    \mbf{0}} \eqdef \left(|\{x \in \mc{X}: (n_d \hat{p}^{(d)}_x)_{d=1}^D = 
    \mbf{i}\}|\right)_{\mathbf{i} \geq \mbf{0}}
\end{equation}
Note that $\mc{F}_\mbf{0} = |\mc{X} \setminus \hat{\mc{X}}| = K - \hat{K}$ 
counts the number of symbols ``unseen'' by any of the $D$ samples, and is thus 
unknown if the support set size $K$ is unknown.
Then the probability to draw the $D$-dimensional fingerprint $\mc{F}$ under 
$\mathbf{p}$ is (obtained analogously to the $D=1$ case in Appendix 
\ref{app:prob_unlabeled_empirical_distribution}):
\begin{align}
    \mb{P}_{\mathbf{p}}(\mc{F}) &= \left(\prod_{\mathbf{i} \geq \mbf{0}} 
    \frac{1}{\mc{F}_{\mathbf{i}}!}\right) \left(\prod_{d=1}^D 
    \left(\begin{array}{c} n_d \\ n_d \hat{p}^{(d)}\end{array}\right)\right) 
        \sum_{\sigma \in \mc{S}_\mc{X}} \prod_{x \in \mc{X}} \prod_{d=1}^D 
        (p_x^{(d)})^{n_d \hat{p}^{(d)}_{\sigma(x)}} 
        \label{eq:prob_unlabeled_empirical_distribution_DD_in_terms_of_support_set} 
        \\
        &= \frac{1}{(K - \hat{K})!} \left(\prod_{\mathbf{i} \geq \mbf{0}, 
        \mbf{i} \neq \mbf{0}} \frac{1}{\mc{F}_{\mathbf{i}}!}\right) 
        \left(\prod_{d=1}^D \left(\begin{array}{c} n_d \\ n_d 
        \hat{p}^{(d)}\end{array}\right)\right) 
        \perm\Bigg(\underbrace{\left(\prod_{d=1}^D (p^{(d)}_x)^{n_d 
        \hat{p}^{(d)}_{x'}}\right)_{x, x' \in 
        \mc{X}}}_{Q}\Bigg) 
        \label{eq:prob_unlabeled_empirical_distribution_DD_in_terms_of_support_set_sizes}
\end{align}
where in 
(\ref{eq:prob_unlabeled_empirical_distribution_DD_in_terms_of_support_set_sizes}) 
the first product is over all $D$-dimensional indices $\mbf{i}$ such that 
$\min_d i_d \geq 0$ and $\max_d i_d > 0$.


\subsection{The PML distributions in $D$ dimensions}

For a given collection $\mc{P}$ of $D$-tuples of distributions, the 
$D$-dimensional PML distributions $\mathbf{p}^*$ are:
\begin{align}
    \mathbf{p}^* \eqdef (p^{(d)*})_{d=1}^D &\eqdef \argmax_{\mathbf{p} \in 
    \mc{P}} \mb{P}_\mathbf{p}(\mc{F}(\hat{\mathbf{p}})) \\
    &= \argmax_{\mathbf{p} \in \mc{P}} 
    \frac{\perm(Q)}{(K - \hat{K})!} 
    \label{eq:def_unlabeled_ML_DD_alphabet_size_known}
\end{align}
where in (\ref{eq:def_unlabeled_ML_DD_alphabet_size_known}) we discarded all 
$\mbf{p}$-independent factors of $\mb{P}_{\mbf{p}}(\mc{F})$ 
(\ref{eq:prob_unlabeled_empirical_distribution_DD_in_terms_of_support_set_sizes}).  
$\mc{F}_\mbf{0} = K - \hat{K} =
|\mc{X}\setminus\hat{\mc{X}}|$ depends on $p$ through its support set size.  
Note that $\mb{P}_\mbf{p}(\mc{F})$ is invariant under relabeling of the 
components of $\mbf{p}$, so we can choose $\mbf{p}^*$ to be non-increasing in 
the same ordering as we choose for the support set $\mc{X}$.
Note that the set $\mathcal{P}$ is not necessarily the same as the set of all 
discrete distributions.

Note that finding the $D$-dimensional PML distributions 
(\ref{eq:def_unlabeled_ML_DD_alphabet_size_known}) is not equivalent to finding 
$D$ $1$-dimensional PML distributions since $\mb{P}_\mbf{p}(\mc{F})$ 
(\ref{eq:prob_unlabeled_empirical_distribution_DD_in_terms_of_support_set_sizes}) 
does not factor into $D$ pieces.

If the collection of $D$-tuples of distributions $\mc{P}$ includes 
distributions with different support set sizes, then we can estimate the 
support set size by breaking up the optimization in 
(\ref{eq:def_unlabeled_ML_DD_alphabet_size_known}) into two steps:
\begin{equation}
    K^* \eqdef \argmax_K \left( \frac{1}{(K - \hat{K})!} \max_{\mbf{p} \in 
    \mc{P}_K} \perm\left(Q\right)\right)
    \label{eq:def_unlabeled_ML_DD_alphabet_size_unknown}
\end{equation}
whenever the max over $K$ exists, where $\mc{P}_K \eqdef \{\mbf{p} \in \mc{P}: 
|\bigcup_{d=1}^D \supp(p^{(d)})| = K\}$.

The case $D \geq 2$ is more complicated than the case $D = 1$ because the $D$ 
support sets $(\supp(p^{(d)}))_{d=1}^D$ can overlap partially.  If $\mc{X}$ is 
the joint support set -- that is, $\mc{X} = \bigcup_{d=1}^D \supp(p^{(d)})$ -- 
then we can associate with each symbol $x \in \mc{X}$ an indicator function 
$\phi_d(x) = \mathbbm{1}(x \in \supp(p^{(d)}))$, so there are $2^D - 1$ 
possible values $(\phi_d(x))_{d=1}^D$ for each $x \in \mc{X}$~\footnote{$1$ 
less than $2^D$ since $\phi_d(x) = 0 \ \forall d$ implies $x \notin \mc{X}$.}.  
We could then be more ambitious and estimate not just the joint support set 
size $K = |\mc{X}|$, but also the optimal value $(\phi_d(x))_{d=1}^D$ for all 
symbols.  Our proposed approximate PML approximation scheme for $D \geq 2$ is 
less ambitious and only supports estimation of $|\mc{X}|$, but not of the finer 
structure of the joint support set $\mc{X}$ in terms of $D$ partially 
overlapping support sets $(\supp(p^{(d)}))_{d=1}^D$.

%
%
%
%

\subsection{Approximate PML in $D$ dimensions: the case of known support set 
size} \label{app:approx_PML_DD_known_support_set_size}

In this Section we assume the support set size $K = |\mc{X}|$ is known.
Analogously to the notation in the $D=1$ case of Section 
\ref{sec:approx_PML_single_distribution} and the proof of the permanent lower 
bound for $D=1$ in Appendix \ref{app:permanent_lower_bound_reexpressed}, define 
the level sets $\mc{X}_{\mbf{u}}(\mbf{p})$ of $\mbf{p}$ indexed by 
$D$-dimensional real vector $\mbf{u} = (u_d)_{d=1}^D$:
\begin{equation}
    \mc{X}_{\mbf{u}}(\mbf{p}) \eqdef \{x \in \mc{X}: (p^{(d)}_x)_{d=1}^D = 
    \mbf{u}\}
\end{equation}
the unique $D$-tuple entries of $\mbf{p}$ by $\mc{U}(\mbf{p})$:
\begin{equation}
    \mc{U}(\mbf{p}) \eqdef \{(p^{(d)}_x)_{d=1}^D : x \in \mc{X}\}
\end{equation}
and the partition $\mc{A}(\mbf{p})$ of $\mc{X}$ into level sets of $\mbf{p}$:
\begin{equation}
    \mc{A}(\mbf{p}) \eqdef \{\mc{X}_{\mbf{u}}(\mbf{p}): \mbf{u} \in 
    \mc{U}(\mbf{p})\} 
    \label{eq:def_partition_into_level_sets_of_p_multiple_distributions}
\end{equation}


We assume $K = |\mc{X}|$ is known, so $\mbf{p}^*$ 
(\ref{eq:def_unlabeled_ML_DD_alphabet_size_known}) is a maximizer of
\begin{equation}
    V(\mbf{p}) \eqdef \log(\perm(Q))
\end{equation}

As for the $D=1$ case, we do not know how to computationally efficiently 
maximize $V(\mathbf{p})$ over $\mathbf{p}$, so we settle for maximizing a lower 
bound $\bar{V}(\mathbf{p})$, analogous to (\ref{eq:permanent_lower_bound}) and 
motivated by our empirical observations for the case $D=2$ in Section 
\ref{sec:solving_PML_exactly}:
\begin{align}
    V(\mbf{p}) \geq \bar{V}(\mathbf{p}) &\eqdef \log\left(\sum_{\sigma \in 
    \mc{S}_{\mc{X},\mbf{p}}} \prod_{x \in \mc{X}} \prod_{d=1}^D 
    (p_x^{(d)})^{n_d \hat{p}^{(d)}_{\sigma(x)}}\right) 
    \label{eq:permanent_lower_bound_multiple_distributions} \\
    &= -\sum_{d=1}^D n_d \left(D(\hat{p}^{(d)}||p^{(d)}) + 
    H(\hat{p}^{(d)})\right) + \sum_{\alpha \in \mc{A}(\mbf{p})} \log(|\alpha|!)
    \label{eq:permanent_lower_bound_reexpressed_multiple_distributions}
\end{align}
where $\mc{S}_{\mc{X},\mbf{p}}$ is analogous to 
(\ref{eq:def_subgroup_of_symmetric_group}) -- a subgroup of $\mc{S}_\mc{X}$ 
consisting of all permutations that exchange only those alphabet symbols that 
are in the same level set of $\mbf{p}$:
\begin{equation}
    \mc{S}_{\mc{X},\mbf{p}} \eqdef \{\sigma \in \mc{S}_{\mc{X}} : \sigma 
    \mbf{p} = (\sigma p^{(d)})_{d=1}^D = \mbf{p}\}
\end{equation}
(\ref{eq:permanent_lower_bound_reexpressed_multiple_distributions}) follows 
from (\ref{eq:permanent_lower_bound_multiple_distributions}) by an analogous 
argument to the one given in in Appendix 
\ref{app:permanent_lower_bound_reexpressed} for the case $D=1$.

As in the $D=1$ case, the intuition is that we aproximate the permanent by 
summing over only those permutations that contribute ``a lot'' to the value, 
and then maximize this lower bound $\bar{V}(\mbf{p})$ over $\mbf{p}$.
Our approximate PML distributions are the $D$-tuple:
\begin{equation}
    \bar{\mathbf{p}}^* = (\bar{p}^{(d)*})_{d=1}^D \eqdef \argmax_{\mathbf{p} 
    \in \mc{P}} \bar{V}(\mathbf{p})
    \label{eq:def_approximate_unlabeled_ML_multiple_distributions}
\end{equation}
As in the 1-D case, the first term in 
(\ref{eq:permanent_lower_bound_reexpressed_multiple_distributions}) encourages 
$\bar{\mbf{p}}^*$ to clump many symbols together, while the second term 
encourages $\bar{\mbf{p}}^*$ to be similar to $\hat{\mbf{p}}$, dominating as $n 
\ra \infty$.


Note that optimizing $\bar{V}(\mbf{p})$ over $\mbf{p}$ is not equivalent to 
solving $D$ independent optimization problems because the $D$ distributions 
``interact'' through the second summation in expression 
(\ref{eq:permanent_lower_bound_reexpressed_multiple_distributions}) (the 
summation over $\mc{A}(\mbf{p})$).

As in the 1-D case, we can show that for all $d$ the approximate PML 
distribution $\bar{p}^{(d)*}$ 
(\ref{eq:def_approximate_unlabeled_ML_multiple_distributions}) satisfies the 
averaging property (\ref{eq:averaging_condition_ML_unlabeled_approximation}).  
Therefore as in the 1-D case, $\bar{\mbf{p}}^*$ is determined by its partition 
of $\mc{X}$ into level sets.  Thus we seek to maximize the quantity
\begin{equation}
    \bar{V}(\mc{A}) \eqdef \sum_{\alpha \in \mc{A}} \bar{V}(\alpha)
\end{equation}
where for $\alpha \subset \mc{X}$
\begin{equation}
    \bar{V}(\alpha) \eqdef \log(|\alpha|!) + |\alpha| \sum_{d=1}^D n_d 
    \hat{p}^{(d)}_\alpha \log(\hat{p}^{(d)}_\alpha) 
    \label{eq:def_objective_function_on_partition_element_multiple_distributions}
\end{equation}
The optimal partition is
\begin{equation}
        \bar{\mc{A}}^* \eqdef \argmax_{\mc{A}: \text{ partition of } \mc{X}} 
        \bar{V}(\mc{A}) \label{eq:def_optimal_partition_approx_PML_DD}
\end{equation}
Finally $\bar{\mbf{p}}^*$ is obtained by averaging the empirical distributions 
$\hat{\mbf{p}}^*$ within each partition element $\alpha \in \bar{\mc{A}}^*$.

As in the 1-dimensional case, the approximate PML partition $\bar{\mc{A}}^*$ 
has the iso-clumping property (generalizing 
(\ref{eq:same_empirical_prob_symbols_clumped_together_property}) to $D$ 
dimensions):
\begin{equation}
    (\hat{p}^{(d)}_x)_{d=1}^D = (\hat{p}^{(d)}_{x'})_{d=1}^D \Rightarrow 
    \alpha(x) = \alpha(x')
\end{equation}
for all $x, x' \in \mc{X}$, where $\alpha(x), \alpha(x') \in \bar{\mc{A}}^*$ 
are the partition elements containing $x$, $x'$, respectively.  This can be 
shown by adapting the proof for the 1-D case in Appendix 
\ref{app:same_empirical_prob_symbols_clumped_together_property_proof}.


We do not know if $\bar{\mc{A}}^*$ satisfies a convexity property analogous to 
the 1-D case (\ref{eq:convexity_property}).

\subsection{A greedy heuristic to approximately maximize the permanent for $D 
\geq 2$ distributions} \label{sec:dd_PML_approximation}

Unlike the $D = 1$ case, we do not solve the optimization problem 
(\ref{eq:def_approximate_unlabeled_ML_multiple_distributions}) exactly for $D 
\geq 2$.
The difficulty is that there is no natural ordering on $\mb{N}^D$ for $D \geq 
2$, so we can no longer define a structured collection of subsets of $\mc{X}$ 
from which to build the approximate PML partition $\bar{\mc{A}}^*$ 
(\ref{eq:def_optimal_partition_approx_PML_DD}) $\mc{X}$, so we do not offer a 
dynamic programming algorithm like the one described in Section 
\ref{sec:approx_PML_single_distribution}.
Our search for an optimal partition of $\mc{X}$ is similar in flavor to a 
clustering problem, so the added difficulty in $D \geq 2$ dimensions makes some 
sense.  For example, k-means clustering in 1 dimension is exactly solvable by a 
dynamic programming algorithm \cite{WangSong2011kmeans} similar to the one we 
present, but no such algorithm is available for dimension greater than 1 
(indeed k-means is NP-hard for dimension greater than 1).

We settle for a greedy heuristic to approximately maximize $\bar{V}(\mc{A})$, 
and call the resulting approximate optimizer $\bar{\bar{\mc{A}}}^*$:
\begin{equation}
    \bar{\bar{\mc{A}}}^* \approx \argmax_{\mc{A}: \text{ partition of } \mc{X}} 
    \bar{V}(\mc{A}) \label{eq:def_approx_approx_PML_multiple_distributions}
\end{equation}
the double bar 
signaling 
the two levels of approximation (an approximation to the approximate PML 
partition $\bar{\mc{A}}^*$ 
(\ref{eq:def_approximate_unlabeled_ML_multiple_distributions})).  The 
doubly-approximate PML distribution $\bar{\bar{\mbf{p}}}^*$ is obtained by 
averaging the empirical distributions $\hat{\mbf{p}}^*$ within each partition 
element $\alpha \in \bar{\bar{\mc{A}}}^*$.



The greedy heuristic used to compute $\bar{\bar{\mc{A}}}^*$ is given in 
algorithm (\ref{alg:approx_approx_DD_PML_partition}).  In words, we start with 
initial partition $\mc{A}(\hat{\mbf{p}})$ of $\mc{X}$ into the level sets of 
the empirical distributions $\hat{\mbf{p}}$ (we clump equally empirically 
frequent symbols into the same level sets since we know that $\bar{\mc{A}}^*$ 
satisfies the iso-clumping property; see Section 
\ref{app:approx_PML_DD_known_support_set_size})
and then iteratively merge partition elements until it is no longer profitable 
to do so in terms of the objective function $\bar{V}$ increasing.  At each step 
we merge two partition elements that most boost the value of the objective 
function.


\begin{algorithm}
\caption{Approximation to level set partition of the approximate $D$-dimensional PML distribution} \label{alg:approx_approx_DD_PML_partition}
\begin{algorithmic}[1]
    \Function{$\bar{\bar{\mc{A}}}^*$}{$\hat{\mbf{p}},\mbf{n},|\mc{X}|$} \Comment 
    Input: $D$ empirical distributions, $D$ sample sizes, and assumed support set size $|\mc{X}|$
    \State $\mc{A} \gets \mc{A}(\hat{\mbf{p}})$ \Comment Initial partition into 
    level sets of empirical distribution $\hat{\mbf{p}}$ 
    \label{eq:initial_partition}
    \While{$\max_{\alpha_1, \alpha_2 \in \mc{A} : \alpha_1 \neq \alpha_2} 
    \left(\bar{V}(\alpha_1 \cup \alpha_2) - (\bar{V}(\alpha_1) + 
    \bar{V}(\alpha_2))\right) > 0$} \label{eq:max_over_pairs}
    \State $(\alpha_1^*, \alpha_2^*) \gets \argmax_{\alpha_1, \alpha_2 \in 
    \mc{A} : \alpha_1 \neq \alpha_2} \left(\bar{V}(\alpha_1 \cup \alpha_2) - 
    (\bar{V}(\alpha_1) + \bar{V}(\alpha_2))\right)$
    \State $\mc{A} \gets (\mc{A} \cup \{\alpha_1^* \cup \alpha_2^*\}) \setminus \{\alpha_1^*, \alpha_2^*\}$ \Comment Merge two level sets into one 
    \EndWhile
    \State \Return $\mc{A}$ \Comment Output: 
    $\bar{\bar{\mc{A}}}^*(\hat{\mbf{p}})$
\EndFunction
\end{algorithmic}
\end{algorithm}
The inputs $\hat{\mbf{p}}$ and $\mbf{n}$ in algorithm 
(\ref{alg:approx_approx_DD_PML_partition}) are used when evaluating 
$\bar{V}(\alpha)$ 
(\ref{eq:def_objective_function_on_partition_element_multiple_distributions}) 
for $\alpha \subset \mc{X}$.  The input of the support set size $|\mc{X}|$ is 
used in line (\ref{eq:initial_partition}) to initially assign the $|\mc{X}| - 
|\mc{X}(\hat{\mbf{p}})|$ unobserved symbols to a partition element.  The loop 
in algorithm (\ref{alg:approx_approx_DD_PML_partition}) always terminates since 
we only merge distinct partition elements of the finite support set $\mc{X}$.
We can implement algorithm (\ref{alg:approx_approx_DD_PML_partition}) 
efficiently by observing that for $\alpha_1, \alpha_2 \subset \mc{X}$ $\alpha_1 
\neq \alpha_2$:
\begin{equation}
    \hat{p}_{\alpha_1 \cup \alpha_2} = \frac{|\alpha_1| \hat{p}_{\alpha_1} + 
    |\alpha_2| \hat{p}_{\alpha_2}}{|\alpha_1| + |\alpha_2|} 
    \label{eq:averaging_union_disjoint}
\end{equation}
Relation (\ref{eq:averaging_union_disjoint}) lets us evaluate $\bar{V}(\alpha_1 
\cup \alpha_2)$ without summing over all points in $\alpha_1 \cup \alpha$, so 
we can reuse the work we did when we previously computed $\bar{V}(\alpha_1)$, 
$\bar{V}(\alpha_2)$.

%

We obtain the doubly-approximate set of $D$ distributions 
$\bar{\bar{\mbf{p}}}^*$ by setting for all $x \in \mc{X}$, $d \in \{1, \ldots, 
D\}$
\begin{equation}
    \bar{\bar{p}}^{(d)*}_x = \hat{p}^{(d)}_{\alpha(x)}
\end{equation}
where $\alpha(x) \in \bar{\bar{\mc{A}}}^*$ is the partition element containing 
$x$, and $\hat{p}^{(d)}_{\alpha}$ 
(\ref{eq:def_averaging_distribution_over_set}) is the average of 
$\hat{p}^{(d)}$ over $\alpha \subset \mc{X}$.

We know that the partition $\bar{\bar{\mc{A}}}^*$ computed by algorithm 
$(\ref{alg:approx_approx_DD_PML_partition})$ is suboptimal (that is, does not 
maximize $\bar{V}$ (\ref{eq:def_optimal_partition_approx_PML_DD})) because it 
only tries to merge pairs of existing partition elements rather than triplets 
or larger collections; it is possible to construct an example where the optimal 
partition consists of a single set (all of $\mc{X}$) but no pair of partition 
elements of $\mc{A}(\hat{\mbf{p}})$ is ``worth'' merging in the sense of 
satisfying the loop condition in line (\ref{eq:max_over_pairs}).

The running time of the heuristic (\ref{alg:approx_approx_DD_PML_partition}) is 
$O(|\supp(\mc{F})|^2) = O(n)$, where $n = \sum_{d=1}^D n_d$: In line 
$(\ref{eq:initial_partition})$, we have $|\mc{A}(\hat{\mbf{p}})| = 
|\supp(\mc{F})| = O(\sqrt{n})$.  To evaluate the max over $\alpha_1, \alpha_2 
\in \mc{A}$ for the first time in line (\ref{eq:max_over_pairs}) we consider 
$O(|\mc{A}(\hat{\mbf{p}})|^2) = O(n)$ pairs of partition elements.  On every 
subsequent iteration of the loop, we add one new partition element ($\alpha_1^* 
\cup \alpha_2^*$) to $\mc{A}$ and remove two partition elements ($\alpha_1^*$ 
and $\alpha_2^*$) from $\mc{A}$, so there are $O(|\mc{A}|) = O(\sqrt{n})$ new 
pairs of distinct partition elements to check in computing the max in line 
(\ref{eq:max_over_pairs}).  The size of $|\mc{A}|$ decrements by one each time 
the loop is run, so the loop is run at most $O(|\mc{A}(\hat{\mbf{p}})|) = 
O(\sqrt{n})$ times, so the overall complexity of algorithm 
$\ref{alg:approx_approx_DD_PML_partition}$ is $O(n) + O(\sqrt{n}) O(\sqrt{n}) = 
O(n)$.

If the support set size $|\mc{X}|$ is unknown, then we vary the assumed support 
set size $|\mc{X}|$ to optimize $\bar{V}(\bar{\bar{\mc{A}}}^*(\hat{\mbf{p}}, 
\mbf{n}, |\mc{X}|))$ using a bisection search on the range 
$[|\mc{X}(\hat{\mbf{p}})|, \max_{d \in [1,D]} n_d^2]$, increasing the overall 
running time of the algorithm (optimizing over both the level set partition 
$\mc{A}$ and the support set size) to $O(n \log(n))$.  Numerical experiments 
show that $\bar{V}(\bar{\bar{\mc{A}}}^*(\hat{\mbf{p}}, \mbf{n}, |\mc{X}|))$ is 
not unimodal in $|\mc{X}|$, so this heuristic might choose a suboptimal support 
set size.

The $O(n)$ worst-case running time of algorithm 
(\ref{alg:approx_approx_DD_PML_partition}) turns out to be slow for examples of 
practical interest (in making the performance plots of Section 
\ref{sec.divergencepmlplugin}), so we introduce another heuristic to speed 
things up at the cost of adding some tunable knobs to the algorithm.  The 
heuristic is to avoid checking all pairs of partition elements $\alpha_1, 
\alpha_2 \in \mc{A}$, instead checking only the ``promising'' pairs.  A pair 
$\alpha_1, \alpha_2 \in \mc{A}$ is called \emph{promising} if the point
$(\hat{p}^{(d)}_{\alpha_1})_{d=1}^D \in \mb{R}^D$ is one of the $k$ nearest 
neighbors of the point $(\hat{p}^{(d)}_{\alpha_2})_{d=1}^D \in \mb{R}^D$ in 
Euclidean distance\footnote{The nearest neighbor relation is not symmetric, so 
$\alpha_1, \alpha_2$ being promising does not imply that $\alpha_2, \alpha_1$ 
is a promising pair.}.  This heuristic is motivated by the observation that the 
level sets that get merged in the loop of algorithm 
(\ref{alg:approx_approx_DD_PML_partition}) tend to correspond to nearby points 
in Euclidean distance.  Then we modify algorithm 
(\ref{alg:approx_approx_DD_PML_partition}) to first compute the set of $k$ 
nearest neighbors (doable in time $O(|\supp(\mc{F})| \log(|\supp(\mc{F})|)) = 
O(\sqrt{n} \log (\sqrt{n})) = O(\sqrt{n} \log (n))$) and then to update the set 
of nearest neighbors after each loop execution.  This updating can be done 
heuristically by taking the $k$ nearest neighbors of $\hat{p}^{(d)}_{\alpha_1 
\cup \alpha_2}$ to be the $k$ nearest points chosen from among only the union 
of the $k$ nearest neighbors of $(\hat{p}^{(d)}_{\alpha_1})_{d=1}^D$ and the 
$k$ nearest neighbors of $(\hat{p}^{(d)}_{\alpha_2})_{d=1}^D$ (rather than the 
$k$ nearest points chosen from among $(\hat{p}^{(d)}_{\alpha})_{d=1}^D$ for all 
$\alpha \in \mc{A}$).  The loop is executed $O(\sqrt{n})$ times, so the overall 
running time of the algorithm is $O(\sqrt{n} \log(n) + k \sqrt{n})$.  The time 
to compute the empirical histograms $\hat{\mbf{p}}$ is $O(n)$, asymptotically 
in $n$ dominating the runtime of our algorithm for large $n$, but tends to take 
less time in our use cases than running algorithm 
\ref{alg:approx_approx_DD_PML_partition} modified with our $k$ nearest neighbor 
heuristic.
For the performance plots of Section \ref{sec.divergencepmlplugin} we used $k = 
5$.

\begin{figure}[!h]
    \capstart
    \begin{center}
    \includegraphics[width=7in]{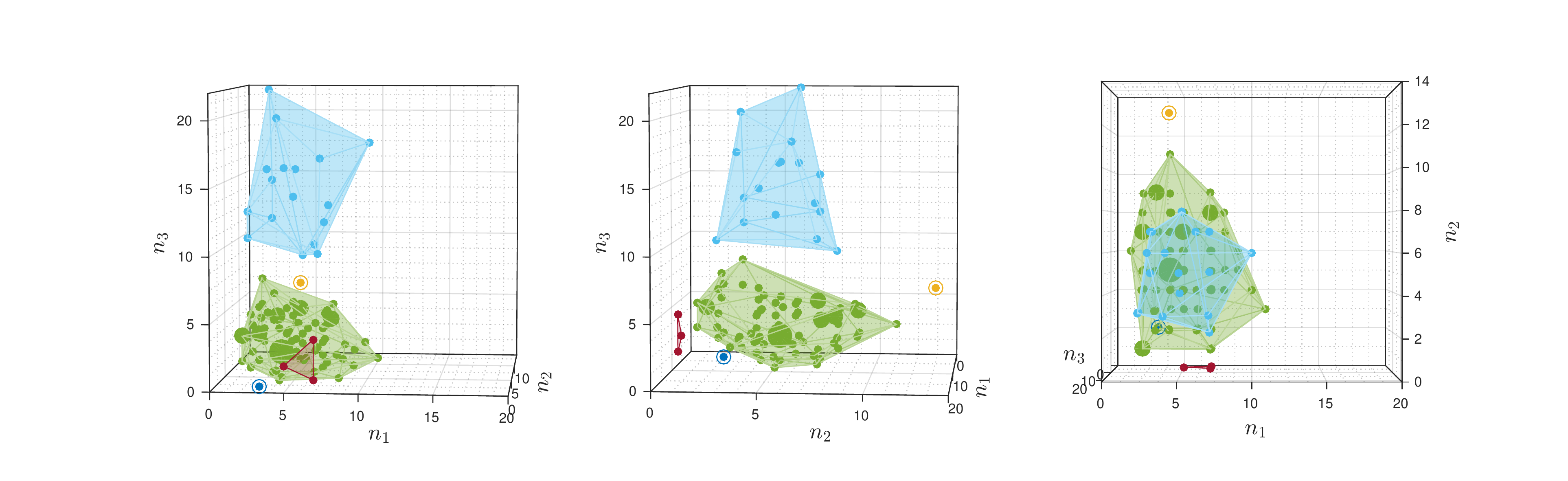}
        \caption{Computing the doubly-approximate PML distribution level set 
        partition $\bar{\bar{\mc{A}}}^*$ 
        (\ref{eq:def_approx_approx_PML_multiple_distributions}) by clumping 
        entries of a 3-D fingerprint $\mc{F}_{i,j,k}$ using algorithm 
        $\ref{alg:approx_approx_DD_PML_partition}$ modified with the 5-nearest 
        neighbors heuristic as described in Section 
        \ref{sec:dd_PML_approximation}, shown in three projections.  The 
        underlying distributions are $p^{(1)} = p^{(2)}$ uniform on 
        $\{1,\ldots,100\}$ and $p^{(3)}$ is a mixture of two uniforms, with 
        half its mass uniform on $\{1,\ldots,20\}$ and the other half uniform 
        on $\{21,\ldots,100\}$.  The sample size is $500$ for all three 
        empirical distributions $(\hat{p}^{(d)})_{d=1}^3$.  Colored convex 
        hulls correspond to symbols assigned to the same clump (partition 
        element).  The marker size of point $(i,j,k)$ is roughly proportional 
        to $\mathcal{F}_{i,j,k}$.
        We can see two large clumps, corresponding to symbols $\{1,\ldots,20\}$ 
        and $\{21,\ldots,100\}$.}
    \label{fig:fingerprint3d}
    \end{center}
\end{figure}

Figures \ref{fig:fingerprint2d} and \ref{fig:fingerprint3d} show the 
doubly-approximate PML level set partition $\bar{\bar{\mc{A}}}^*$ 
(\ref{eq:def_approx_approx_PML_multiple_distributions}) for $D = 2$ and $D = 
3$, respectively.

\section{Mutual and Lautum information} \label{sec.mutuallautum}


For the case of estimating the mutual information $I(X;Y)$, defined as
\begin{align}\label{eqn.mutual}
I(X;Y) = \sum_{x\in \mathcal{X},y\in \mathcal{Y}}  p_{x,y} \log 
    \left(\frac{p_{x,y}}{p_x p_y}\right),
\end{align}
or lautum information~\cite{palomar2008lautum}, defined as
\begin{align}\label{eqn.lautum}
L(X;Y) & = \sum_{x\in \mathcal{X},y\in \mathcal{Y}} p_x p_y \log 
    \left(\frac{p_x p_y}{p_{x,y}}\right),
\end{align} 
we have $n$ samples from a joint distribution $p$ on $\mc{X} \times \mc{Y}$ with 
joint histogram $\hat{p}$. Here $p_x = \sum_{y} p_{x,y}, p_y = \sum_x p_{x,y}$ are the marginal distributions of $X$ and $Y$, respectively. 

\subsection{Product of permutation groups: one candidate for PML} 
\label{sec.mipmlformulation1}

It is clear that the mutual information and the lautum information are 
invariant to the product of two permutation groups $\mc{S}_{\mathcal{X}} \times 
\mc{S}_{\mathcal{Y}}$, acting on $p$ as:
\begin{equation}
    ((\sigma_\mc{X}, \sigma_\mc{Y}) p)_{x,y} \eqdef p_{\sigma_\mc{X}(x), 
    \sigma_\mc{Y}(y)}
\end{equation}
for all $(\sigma_\mc{X}, \sigma_\mc{Y}) \in \mc{S}_{\mathcal{X}} \times 
\mc{S}_{\mathcal{Y}}$.  Let $(\mc{S}_{\mathcal{X}} \times \mc{S}_{\mathcal{Y}}) 
p$ denote the orbit of $p$ under the group action.  Following the general 
profile maximum likelihood methodology for the product of these two permutation 
groups, the PML distribution $p^*$ maximizes the following quantity over a set 
of distribution $\mathcal{P}$ on $\mathcal{X}\times \mathcal{Y}$:
\begin{align}
    \mb{P}_p((\mc{S}_\mc{X} \times \mc{S}_\mc{Y})\hat{p}) &= \sum_{\hat{p}' \in 
    (\mc{S}_\mc{X} \times \mc{S}_\mc{Y}) \hat{p}} \mb{P}_p(\hat{p}') \\
    &\overset{\text{(a)}}{=} \frac{|(\mc{S}_\mc{X} \times 
    \mc{S}_\mc{Y})\hat{p}|}{|\mc{S}_\mc{X} \times \mc{S}_\mc{Y}|} 
    \sum_{(\sigma_\mc{X}, \sigma_\mc{Y}) \in \mc{S}_\mc{X} \times 
    \mc{S}_\mc{Y}} \mb{P}_p((\sigma_\mc{X}, \sigma_\mc{Y}) \hat{p}) \\
    &= \left(\prod_{i \geq 0} \frac{1}{\mc{F}_i!}\right) \left(\begin{array}{c} 
    n \\ n \hat{p}\end{array}\right) \sum_{(\sigma_\mc{X}, \sigma_\mc{Y}) \in 
    \mc{S}_\mc{X} \times \mc{S}_\mc{Y}} \prod_{(x,y) \in \mc{X} \times \mc{Y}} 
    p_{x,y}^{n \hat{p}_{\sigma_\mc{(\sigma_\mc{X}, \sigma_\mc{Y})X}(x), 
    \sigma_\mc{Y}(y)}} \\
    &= \left(\prod_{i \geq 0} \frac{1}{\mc{F}_i!}\right) \left(\begin{array}{c} 
    n \\ n \hat{p}\end{array}\right) \perm\left(\left(p_{x,y}^{n 
    \hat{p}_{x',y'}}\right)_{(x,y),(x',y') \in \mc{X} \times \mc{Y}}\right),
\end{align}
where in (a) the prefactor corrects for multiple-counting points of the orbit 
$(\mc{S}_\mc{X} \times \mc{S}_\mc{Y}) \hat{p}$ when we sum over 
$(\sigma_\mc{X}, \sigma_\mc{Y}) \in (\mc{S}_\mc{X} \times \mc{S}_\mc{Y})$, and
where $(\mathcal{F}_i)_{i\geq 0}=(|\{(x,y) \in  \mc{X} \times \mc{Y} : n 
\hat{p}_{x,y} = i\}|)_{i \geq 0}$ is the fingerprint of the distribution when 
one treats $\mathcal{X} \times \mathcal{Y}$ as a single alphabet.
Note that in this setting $(\mathcal{F}_i)_{i\geq 0}$ is not sufficient for 
estimating the mutual (lautum) information: the mutual (lautum) information 
cannot be written as a functional of the sorted probabilities of $p_{x,y}$. 

In this setting, it follows from Theorem~\ref{theorem.groupaction} that the sufficient statistic could be defined via the following equivalence class representation: for any empirical distribution $\hat{p}^{(1)} = (\hat{p}^{(1)}_{x,y})$,$\hat{p}^{(2)} = (\hat{p}^{(2)}_{x,y})$, we say $\hat{p}^{(1)}$ is equivalent to $\hat{p}^{(2)}$ if there exists an element $g\in \mathcal{S}_{\mathcal{X}} \times \mathcal{S}_{\mathcal{Y}}$ such that $g \hat{p}^{(1)} = \hat{p}^{(2)}$. The sufficient statistic is the corresponding equivalent classes.  


In this PML formulation, we analyze the cardinality of the corresponding sufficient statistic below. This number appeared in~\cite{oeisa007716}, and its precise asymptotics is obtained in the following theorem. 
\begin{theorem}\label{thm.huacheng}
For any two empirical distributions $\hat{p}^{(1)} = 
(\hat{p}^{(1)}_{x,y})$,$\hat{p}^{(2)} = (\hat{p}^{(2)}_{x,y})$ with identical 
sample size $n$, we say $\hat{p}^{(1)}$ is equivalent to $\hat{p}^{(2)}$ if 
    there exists an element $(\sigma_\mc{X}, \sigma_\mc{Y}) \in 
    \mathcal{S}_{\mathcal{X}} \times \mathcal{S}_{\mathcal{Y}}$ such that 
    $(\sigma_\mc{X}, \sigma_\mc{Y}) \hat{p}^{(1)} = \hat{p}^{(2)}$, where 
    $\mathcal{S}_{\mathcal{X}}, \mathcal{S}_{\mathcal{Y}}$ are the permutation 
    groups on $\mathcal{X},\mathcal{Y}$, respectively. Then, the number of 
equivalence classes is given by $e^{(1+o(1))n\log n }$.  \end{theorem}

Theorem~\ref{thm.huacheng} shows that the cardinality of the sufficient statistic is no longer sub-exponential. This negative result immediately renders Theorem~\ref{theorem.ml} not applicable, which leaves us with no guarantee on the performance of this type of PML on estimating mutual information or lautum information. 

\subsection{Three permutation groups: the other candidate for the PML} 
\label{sec.mipmlformulation2}

The mutual information $I(X;Y)$ can be rewritten as 
\begin{align}
I(X;Y) & = H(P_X) + H(P_Y) - H(P_{XY}) \\
    & = \sum_{x \in \mathcal{X}} p_x \log \left(\frac{1}{p_x}\right) + 
    \sum_{y\in \mathcal{Y}} p_y \log \left(\frac{1}{p_y}\right) - \sum_{x\in 
    \mathcal{X}, y\in \mathcal{Y}} p_{x,y}\log \left(\frac{1}{p_{x,y}}\right),
\end{align}
where $H(P_X) = \sum_{x} p_x \log \frac{1}{p_x}$ denotes the Shannon entropy of 
discrete distribution $P_X$. 

Following the discussion in Section~\ref{sec.standardpml}, if we introduce
\begin{align}
(\mathcal{F}_i^{XY})_{i\geq 0} & = \left( | \left \{ (x,y) \in \mathcal{X}\times \mathcal{Y}: n\hat{p}_{x,y} = i \right \} | \right )_{i\geq 0} \\
(\mathcal{F}_i^{X})_{i\geq 0} & = \left( | \left \{ x \in \mathcal{X}: n\hat{p}_{x} = i \right \} | \right )_{i\geq 0} \\
(\mathcal{F}_i^{Y})_{i\geq 0} & = \left( | \left \{ y \in \mathcal{Y}: n\hat{p}_{y} = i \right \} | \right )_{i\geq 0},
\end{align}
it is clear that $\left( \mathcal{F}_i^{XY},\mathcal{F}_i^{X},\mathcal{F}_i^{Y} \right)_{i\geq 0}$ is sufficient for estimating $I(X;Y)$. 

This motivates the following PML definition:
\begin{align}
    p^* \triangleq \argmax_{p\in \mathcal{P}} \mathbb{P}_{p} \left( 
    \mathcal{F}^{XY}, \mathcal{F}^X, \mathcal{F}^Y \right) 
    \label{eq:def_three_fingerprints_PML}.
\end{align}
In other words, the PML distribution $(p_{x,y}^*)$ on $\mathcal{X}\times 
\mathcal{Y}$ maximizes the probability of observing the three joint 
fingerprints $\left( \mathcal{F}^{XY}, \mathcal{F}^X, \mathcal{F}^Y \right)$.  
This PML formulation, which only applies to mutual information but not lautum 
information, does not suffer from the super-exponential cardinality problem. 
Applying Theorem~\ref{theorem.hardyramanujan}, it follows that the cardinality of 
the sufficient statistic is at most \begin{align}
\left( e^{\pi \sqrt{\frac{2n}{3}}} \right)^3 = e^{\pi \sqrt{6n}},
\end{align}
which is still sub-exponential. However, it seems to be a non-trivial task to 
even approximately solve the PML formulation 
(\ref{eq:def_three_fingerprints_PML}). Moreover, this PML formulation does not 
apply to lautum information.


\section{Estimating the support set size of the approximate PML distribution} 
\label{app:estimate_support_set_size_approximate_PML}

We first state our results in the notation introduced in Sections 
\ref{sec:dynamic_programming_ML_unlabeled_approximation} and 
\ref{sec:estimate_support_set_size_approximate_PML}.

\begin{theorem} \label{thm:optimizing_support_set_size}
    Let $\bar{p}^*$ be the approximate PML distribution.  Then  $\mc{X}_{0:i^*} 
    \in \mc{A}(\bar{p}^*)$ is a level set of $\bar{p}^*$, where
    \begin{equation}
        i^* \eqdef \argmax_{i \in [1,F_+]} \left( \sup_{\mc{F}_0 \in \mb{N}} 
        \left(- \log(\mc{F}_0!) + \bar{V}(\mc{X}_{0:i})\right) + 
        \bar{V}_{i+1}\right) 
        \label{eq:optimal_grouping_zeros_optimize_over_support_set_size}
    \end{equation}
    For each value of $i \in [1,F_+]$ the supremum over $\mc{F}_0$ in 
    (\ref{eq:optimal_grouping_zeros_optimize_over_support_set_size}) is finite 
    and we compute it below.
    
    Let $K_i \eqdef |\mc{X}_{1:i}|$ and $N_i \eqdef \sum_{x \in \mc{X}_{1:i}} n 
    \hat{p}_x$.
    
    If $i = 1$ and $\mc{F}_1 \geq 1$, then $N_1 = K_1 = \mc{F}_1 \geq 1$ is the 
    number of symbols seen exactly once in the sample, and the supremum in 
    (\ref{eq:optimal_grouping_zeros_optimize_over_support_set_size}) is equal 
    to $N_1 \log\left(\frac{N_1}{n}\right)$.  If $\mc{F}_1 > 1$, then the 
    supremum is approached as $\mc{F}_0 \ra \infty$ and is not achieved by any 
    $\mc{F}_0 \in \mb{N}$.  If $\mc{F}_1 = 1$, then the supremum is achieved by 
    all $\mc{F}_0 \in \mb{N}$, so we arbitrarily choose $\mc{F}_0^{(1)} = 0$ as 
    an achiever.

    If $i > 1$ or $\mc{F}_1 = 0$, then $N_i > K_i$ and the supremum in 
    (\ref{eq:optimal_grouping_zeros_optimize_over_support_set_size}) is 
    achieved by $\mc{F}_0^{(i)} \in \mb{N}$ where
    \begin{equation}
        \mc{F}_0^{(i)} \eqdef \argmax_{\mc{F}_0 \in \mb{N}} 
        \left(\log\left(\frac{(K_i + \mc{F}_0)!}{\mc{F}_0!}\right) - N_i 
        \log\left(K_i + \mc{F}_0\right)\right)
        \ \begin{cases}
            = 0 &: N_i \geq K_i H_{K_i} \\
            \in [0, \frac{K_i^2 - N_i}{N_i - K_i} + 1] &: K_i H_{K_i} > N_i > K_i
        \end{cases} \label{eq:optimal_number_unseen_symbols}
    \end{equation}
    where $H_k \eqdef \sum_{j=1}^k \frac{1}{j}$ is the $k$-th harmonic number.  


\end{theorem}

If the supremum in 
(\ref{eq:optimal_grouping_zeros_optimize_over_support_set_size}) is achieved by 
finite $\mc{F}_0 = \mc{F}_0^{(i^*)}$ (this is the case unless $i^* = 1$ and 
$\mc{F}_1 > 1$)
then the approximate PML distribution $\bar{p}^*$ has finite support set 
size $\bar{K}^* = \mc{F}_0^{(i^*)} + \hat{K}$, where $\hat{K}$ is the 
support set size of the empirical distribution $\hat{p}$.  Then we compute 
$\bar{p}^*$ as in Section 
\ref{sec:dynamic_programming_ML_unlabeled_approximation} with known support set 
size $\bar{K}^*$.  If the supremum is not achieved by finite $\mc{F}_0$ (this 
is the case if $i^* = 1$ and $\mc{F}_1 > 1$), then we say $\bar{p}^*$ has a 
continuous part of mass $N_1 / n$ and discrete part $\bar{p}^*_\text{d}$ of 
mass $1 - N_1/n$ supported on $\mc{X}_{2:F_+} = \mc{X}\setminus\mc{X}_{0:1}$.  
Then we compute $\bar{p}^*_\text{d}$ as in Section 
\ref{sec:dynamic_programming_ML_unlabeled_approximation} with known support set 
size $|\mc{X}_{2:F_+}|$.

Since the function in the argmax in (\ref{eq:optimal_number_unseen_symbols}) is 
unimodal in $\mc{F}_0$ and $0 \leq \mc{F}_0^{(i)} \leq \frac{K_i^2 - N_i}{N_i - 
K_i} + 1 \leq N_i^2 + 2$ (where the last inequality follows from  $N_i \geq K_i 
+ 1$), we can compute the optimizing integer $\mc{F}_0^{(i)}$ in $O(\log(N_i))$ 
iterations of a bisection search.  We must do this bisection search $F_+ = 
|\supp(\mc{F}_+)|$ times.  As remarked earlier, since $|\supp(\mc{F}_+)| \leq 
\sqrt{2n}+1$ and $N_i \leq n$ for all $i$, then the total run time of our 
support set size optimization scheme is $O(\sqrt{n} \log(n))$.

Thus to estimate the support set size of the approximate PML distribution we 
solve $F_+$ optimization problems, corresponding to estimation of the support 
set size of a uniform distribution (since all symbols in $\mc{X}_{0:i}$ are 
assigned to a single level set) given a sample of size $N_i$ with $K_i$ 
distinct symbols.  

\begin{proof}

Optimizing over the support set size $K$, the approximate PML 
distribution $\bar{p}^*$ satisfies:
\begin{align}
    \log(\mb{P}_{\bar{p}^*}(\mc{F})) &= \sup_{K} \Big( - \log((K - \hat{K})!)+ 
    \max_{p \in \mc{P}_{K}} \left(\bar{V}(p)\right)\Big) \\
    &= \sup_{\mc{F}_0} \Big( - \log(\mc{F}_0!)+ \max_{p \in \mc{P}_{\hat{K} + 
    \mc{F}_0}} \left(\bar{V}(p)\right)\Big) \\
    &\overset{\text{(a)}}{=} \sup_{\mc{F}_0} \Big( - \log(\mc{F}_0!) + 
    \bar{V}_0\Big) \\
    &\overset{\text{(b)}}{=} \sup_{\mc{F}_0} \Big( - \log(\mc{F}_0!) + \max_{i 
    \in [1,F_+]}\left(\bar{V}(\mc{X}_{0,i}) + \bar{V}_{i+1}\right)\Big) \\
    &\overset{\text{(c)}}{=} \max_{i \in [1,F_+]}\left(\bar{V}_{i+1} + 
    \sup_{\mc{F}_0} \left(\bar{V}(\mc{X}_{0,i}) - \log(\mc{F}_0!)\right)
    \right) \\
    &\overset{\text{(d)}}{=} \max_{i \in [1,F_+]}\bigg(\bar{V}_{i+1} + N_i 
    \log\left(\frac{N_i}{n}\right) + \sup_{\mc{F}_0} 
    \bigg(\underbrace{\log\left(\frac{(K_i + \mc{F}_0)!}{\mc{F}_0!}\right) - 
    N_i \log\left(K_i + \mc{F}_0\right)}_{f(\mc{F}_0, K_i, N_i)}\bigg)
    \bigg) \label{eq:optimize_approx_PML_over_support_set_size}
\end{align}
where (a) follows from (\ref{eq:dynamic_programming_value}), (b) follows from 
    (\ref{eq:dynamic_programming_update}), (c) follows since $\bar{V}_i$ is 
    independent of $\mc{F}_0$ (and hence $K$) for $i \geq 1$, and (d) from 
    (\ref{eq:partition_element_value}) and (\ref{eq:V_ij_value}).  We use 
    $\sup_{\mc{F}_0}$ (rather than $\max_{\mc{F}_0}$) since a maximizing value 
    of $\mc{F}_0$ might not exist.  We later show that the supremum in 
    (\ref{eq:optimize_approx_PML_over_support_set_size}) is finite, so we 
    interchange the order of the max and supremum.  The function $f(\mc{F}_0, 
    K, N)$ is defined for non-negative arguments as in 
    (\ref{eq:optimize_approx_PML_over_support_set_size}).
Note that we do not allow $i = 0$ in the optimization 
(\ref{eq:optimize_approx_PML_over_support_set_size}) in order to exclude 
$\mc{X}_{0,0}$ (the set of unseen symbols) as a level set of $p$; otherwise, 
due to the averaging-over-clumps property 
(\ref{eq:averaging_condition_ML_unlabeled_approximation}), we would have $p$ 
    vanish on $\mc{X}_{0,0}$, violating the assumption that the support set 
    size of $p$ is $K$.  It is possible to have $\bar{K}^* = \hat{K} = 
    |\supp(\hat{p})|$, corresponding to the case $\mc{F}_0 = 0$, so that 
    $\mc{X}_{0,i} = \mc{X}_{1,i}$ for all $i$.

    Let's compute the supremization in 
    (\ref{eq:optimize_approx_PML_over_support_set_size}).  We can check that 
    $f(\mc{F}_0, N, N) < 0$ for all $N > 1$ and $\mc{F}_0 \geq 0$, and
    \begin{equation}
        \lim_{\mc{F}_0 \ra \infty} f(\mc{F}_0, N, N) = 0
    \end{equation}
    so if $i = 1$ and $\mc{F}_1 > 1$, then $N_1 = K_1 = \mc{F}_1 > 1$ and the 
    supremum in (\ref{eq:optimize_approx_PML_over_support_set_size}) is equal 
    to $0$, is approached as $\mc{F}_0 \ra \infty$, and is not achieved by any 
    $\mc{F}_0 \in \mb{N}$.
    
    We can check that $f(\mc{F}_0, 1, 1) = 0$ for all $\mc{F}_0$, so if $i = 1$ 
    and $\mc{F}_1 = 1$, then $N_1 = K_1 = \mc{F}_1 = 1$ and the supremum in 
    (\ref{eq:optimize_approx_PML_over_support_set_size}) is equal to $0$ and is 
    achieved by all $\mc{F}_0 \in \mb{N}$, so we arbitrarily choose 
    $\mc{F}_0^{(1)} = 0$ as an achiever.

    If $i > 1$ or $\mc{F}_1 = 0$, then $N_i > K_i$.  When $N > K$ the function 
    $f(\mc{F}_0, K, N)$ is continuous in $\mc{F}_0$~\footnote{Replacing $x!$  
    with $\Gamma{(x+1)}$, where $\Gamma$ is the gamma function.} and unimodal 
    with a global maximum.  Solving
    \begin{equation}
        \frac{\partial}{\partial \mc{F}_0} f(\mc{F}_0, K, N^*) = 0
    \end{equation}
    for $N^*$ we find $N^* = K H_K$, where $H_K$ is the $K$-th harmonic number.  
    Thus if $N_i \geq K_i H_{K_i}$, then the supremum in 
    (\ref{eq:optimize_approx_PML_over_support_set_size}) is achieved by 
    $\mc{F}_0 = 0$.  If $N_i < K_i H_{K_i}$, then we can upper bound the 
    supremizing value $\mc{F}_0$ by finding the unique inflection point of $f$ 
    in $\mc{F}_0$ beyond the global maximum.  Setting the ``discrete'' second 
    derivative to $0$ (we take the discrete derivative because the function 
    $\frac{\partial f}{\partial \mc{F}_0}$ is easier to work with for integer 
    $\mc{F}_0$):
    \begin{equation}
        \left.\frac{\partial}{\partial \mc{F}_0}\left(f(\mc{F}_0+1,K,N) - 
        f(\mc{F}_0,K,N)\right)\right|_{\mc{F}_0 = \mc{F}_0^*} = 0
    \end{equation}
    we find $\mc{F}_0^* = \frac{K^2 - N}{N - K}$.  If $K < N < K H_K$ then $K > 
    1$ (since $1 = H_1$) and $N < K^2$, so $\mc{F}_0^* > 0$.  Thus we can upper 
    bound the inflection point of $f$ by $\frac{K^2 - N}{N - K} + 1$, so if 
    $K_i < N_i < K_i H_{K_i}$, then $\mc{F}_0^{(i)} \in [0, \frac{K_i^2 - 
    N_i}{N_i - K_i} + 1]$.

    
\end{proof}

\bibliographystyle{IEEEtran}
\bibliography{di}

\end{document}